\documentclass[10pt]{article}
\usepackage[margin=1in]{geometry}

\usepackage{amsmath,amssymb,amsthm,mathtools,bm}
\usepackage{algorithm}
\usepackage{algpseudocode}
\usepackage{booktabs}
\usepackage{tabularx}
\usepackage{array}
\usepackage{makecell}
\usepackage{enumitem}
\usepackage{graphicx}
\usepackage{subcaption}
\usepackage{titletoc}
\usepackage{placeins}
\usepackage{xcolor}
\usepackage{url}
\usepackage{xurl}
\usepackage{float}
\IfFormatAtLeastTF{2026-06-01}{}{\usepackage{aliascnt}}
\usepackage{microtype}
\usepackage{tcolorbox} 
\usepackage{listings}
\usepackage{enumitem}
\usepackage{tikz}
\usetikzlibrary{arrows.meta,positioning,calc,fit,backgrounds}

\definecolor{todored}{RGB}{190,30,45}

\usepackage{natbib}
\setcitestyle{authoryear,round,semicolon}

\usepackage[
    colorlinks=true,
    linkcolor=blue,
    citecolor=blue,
    urlcolor=blue
]{hyperref}

\usepackage[
    capitalise,
]{cleveref}
\usepackage{mathtools}
\mathtoolsset{showonlyrefs}
\theoremstyle{plain}
\newtheorem{theorem}{Theorem}[section]
\IfFormatAtLeastTF{2026-06-01}{%
\newtheorem{lemma}[theorem]{Lemma}
\newtheorem{proposition}[theorem]{Proposition}
\newtheorem{corollary}[theorem]{Corollary}
}{%
\newaliascnt{lemma}{theorem}
\newtheorem{lemma}[lemma]{Lemma}
\aliascntresetthe{lemma}
\newaliascnt{proposition}{theorem}
\newtheorem{proposition}[proposition]{Proposition}
\aliascntresetthe{proposition}
\newaliascnt{corollary}{theorem}
\newtheorem{corollary}[corollary]{Corollary}
\aliascntresetthe{corollary}
}

\theoremstyle{definition}
\newtheorem{definition}{Definition}[section]

\theoremstyle{remark}
\newtheorem{remark}{Remark}[section]

\newcommand{\R}{\mathbb{R}}
\newcommand{\E}{\mathbb{E}}
\renewcommand{\P}{\mathbb{P}}
\newcommand{\1}{\mathbf{1}}
\newcommand{\calP}{\mathcal{P}}
\newcommand{\calE}{\mathcal{E}}

\newcommand{\kl}{\operatorname{kl}}
\newcommand{\KL}{\operatorname{KL}}
\newcommand{\Borda}{\mathrm{B}}
\newcommand{\VBEGE}{\textnormal{\textsc{VB-EGE}}}
\newcommand{\thr}{\mathrm{thr}}

\newcommand{\experimenttablesize}{\footnotesize}

\newcommand{\slcomment}[1]{%
  \textcolor{red}{\textbf{[SL: #1]}}%
}

\setlist[itemize]{leftmargin=*,topsep=3pt,itemsep=2pt}
\setlist[enumerate]{leftmargin=*,topsep=3pt,itemsep=2pt}
\graphicspath{{figures/}}
 
\title{Beyond Scalar Leaderboards: Adaptive Sampling for \\ 
Pareto Frontier Identification in Multi-Objective LLM Arenas
}

\author{
Liang Twist Shan
\textsuperscript{1,}\thanks{Correspondence to Liang Twist Shan (\href{mailto:lshan9@wisc.edu}{\texttt{lshan9@wisc.edu}}) and Zhimei Ren (\href{mailto:zren@wharton.upenn.edu}{\texttt{zren@wharton.upenn.edu}}).}
\qquad
Jiahao Ai
\textsuperscript{2}
\qquad
Zhimei Ren
\textsuperscript{2,}\footnotemark[2]
\\[0.75em]
{\normalsize
\textsuperscript{1}
Department of Statistics, University of Wisconsin-Madison
}
\\
{\normalsize
\textsuperscript{2}
Department of Statistics and Data Science, University of Pennsylvania
}
}
\date{}

\crefname{section}{Section}{Sections}
\Crefname{section}{Section}{Sections}
\crefname{subsection}{Section}{Sections}
\Crefname{subsection}{Section}{Sections}
\crefname{subsubsection}{Section}{Sections}
\Crefname{subsubsection}{Section}{Sections}
\crefname{appendix}{Appendix}{Appendices}
\Crefname{appendix}{Appendix}{Appendices}
\crefname{figure}{Figure}{Figures}
\Crefname{figure}{Figure}{Figures}
\crefname{subfigure}{Figure}{Figures}
\Crefname{subfigure}{Figure}{Figures}
\crefname{table}{Table}{Tables}
\Crefname{table}{Table}{Tables}
\crefname{theorem}{Theorem}{Theorems}
\Crefname{theorem}{Theorem}{Theorems}
\crefname{lemma}{Lemma}{Lemmas}
\Crefname{lemma}{Lemma}{Lemmas}
\crefname{proposition}{Proposition}{Propositions}
\Crefname{proposition}{Proposition}{Propositions}
\crefname{corollary}{Corollary}{Corollaries}
\Crefname{corollary}{Corollary}{Corollaries}
\crefname{definition}{Definition}{Definitions}
\Crefname{definition}{Definition}{Definitions}
\crefname{assumption}{Assumption}{Assumptions}
\Crefname{assumption}{Assumption}{Assumptions}
\crefname{algorithm}{Algorithm}{Algorithms}
\Crefname{algorithm}{Algorithm}{Algorithms}
\crefname{equation}{Equation}{Equations}
\Crefname{equation}{Equation}{Equations}
\crefname{remark}{Remark}{Remarks}
\Crefname{remark}{Remark}{Remarks}
\crefname{example}{Example}{Examples}
\Crefname{example}{Example}{Examples}

\begin{document}

\setcounter{footnote}{1}
\maketitle
\begin{abstract}
The evaluation of large language models (LLMs) is inherently multi-objective, as different models may offer different tradeoffs across performance dimensions such as math, code, and writing. Because such evaluations are often based on costly and time-consuming human feedback, it is important to draw reliable conclusions using as few comparisons as possible. In this paper, we present {\em Vector-Borda Empirical-Gap Elimination (VB-EGE)}, a fixed-confidence adaptive sampling algorithm for efficiently identifying models on the Pareto frontier in arena-style evaluation, where each comparison evaluates a pair of models on a prompt with respect to a selected performance dimension. VB-EGE progressively eliminates models from an active set once sufficient statistical evidence has been accumulated to certify their Pareto status. Under a multi-objective Bradley--Terry model, we show that VB-EGE returns the correct Pareto frontier with high probability and derive an instance-dependent upper bound on its sample complexity. We further establish a worst-case lower bound showing that VB-EGE is minimax-optimal up to logarithmic factors. Simulations and experiments on three real-world Arena datasets (Arena, SearchArena, and VisionArena) demonstrate VB-EGE's empirical effectiveness. 
\end{abstract}

\section{Introduction}
\label{sec:introduction}
Pairwise human-preference judgments have become a standard tool to evaluate large language models (LLMs). 
In arena-style evaluation, a user submits a prompt, views two 
responses with the model identities hidden, 
and selects the preferred response~\citep{zheng2023judging, chiang2024chatbot}.
Given these pairwise comparisons, a common approach is to fit a statistical model 
to estimate a latent score for each model, and use the estimated scores 
to rank the models. But model quality is not one-dimensional: 
a model preferred for code generation need not be preferred for 
creative writing or vision~\citep{liang2023holistic,frick2025prompt}. 
How should models be compared without collapsing 
these tradeoffs into a scalar score?

\begin{figure}[!t]
    \centering
    \includegraphics[trim=0 6bp 0 10bp,clip,width=\linewidth,keepaspectratio]{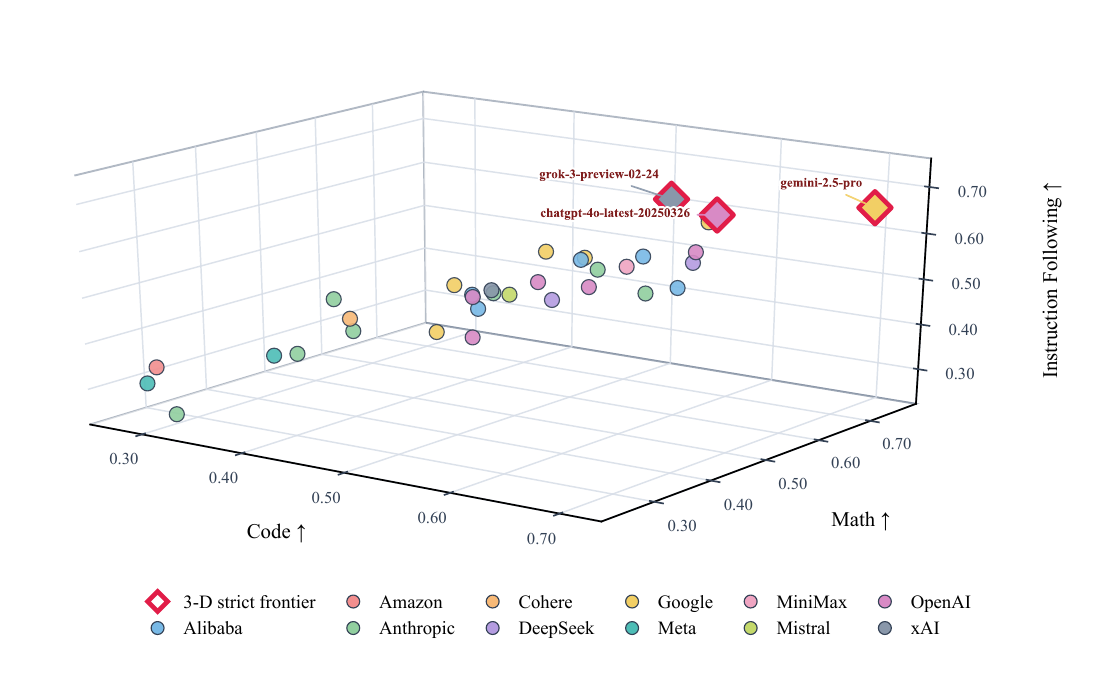}
    \caption{Three-dimensional comparison along the Code, Math, and Instruction-Following objectives from the Arena-Human-Preference-140K dataset~\citep{chiang2024chatbot}. Diamonds denote the strict Pareto frontier.}
    \label{fig:arena-pareto-frontiers}
\end{figure}

A natural target is the strict Pareto frontier:
a model lies outside the frontier only if another model 
has a strictly higher latent score on every objective. 
This criterion excludes uniformly inferior models, while retaining models 
with different cross-objective strengths.
Identifying this frontier reliably, however,
requires costly human-preference comparisons 
across model pairs and objectives.
How should the comparisons be allocated to 
identify the frontier efficiently? 
We develop an adaptive sampling  procedure that 
identifies the strict Pareto frontier at a prescribed confidence level 
while reducing the number of comparisons required.
As a preview, Figure~\ref{fig:arena-pareto-frontiers}
presents an application of our method to comparisons from  
Arena-Human-Preference-140K~\citep{chiang2024chatbot},
collected through July 2025. Treating 
code, math, and Instruction-Following as three objectives,
the method identifies three models on the strict Pareto frontier.

\subsection{Problem formulation}
\label{sec:problem}
We formalize this task as a fixed-confidence 
sequential identification problem. 
Consider \(K\ge 2\) candidate models evaluated on \(d\) objectives. 
Suppose that each model \(i\in [K]:= \{1,\dots,K\}\) 
is characterized by an unknown 
latent score vector
\(\theta_i=(\theta_{i,1},\ldots,\theta_{i,d})\), 
with larger values indicating better performance along 
every coordinate.

Let \(\mathcal{F}_{t-1}\) denote the observations and algorithmic
randomization revealed before round \(t\).
At round \(t\), the learner uses a possibly randomized policy based on
\(\mathcal{F}_{t-1}\) to select a query \(Q_t=(I_t,J_t,R_t)\),
where \(I_t,J_t\in[K]\) are distinct models to be compared and \(R_t\in[d]\)
indexes the evaluation metric.
For each possible query $(i,j,r)$, let $Y_t^{(r)}(i,j)\in\{0,1\}$ denote the potential comparison outcome at round $t$, with value $1$ indicating that $i$ is preferred over $j$ on objective $r$. Only the outcome of the selected query is observed: $Y_t:=Y_t^{(R_t)}(I_t,J_t)$.
Conditional on the past history and the chosen query, the comparison outcome follows the Bradley--Terry (BT) model~\citep{bradley1952rank}:
\begin{equation}
    \mathbb{P}_{\theta}
    \left(
        Y_t=1
        \,\middle|\,
        \mathcal{F}_{t-1},
        Q_t=(i,j,r)
    \right)
    =
    p_{ij}^{(r)}(\theta)
    :=
    \sigma(\theta_{i,r}-\theta_{j,r}),
    \qquad \text{where }
    \sigma(x):=\frac{1}{1+e^{-x}}.
    \label{eq:bt-model}
\end{equation}
Consequently, \(p_{ji}^{(r)}=1-p_{ij}^{(r)}\); 
for convenience, we set \(p_{ii}^{(r)}=1/2\). 
Each query yields a single objective-specific binary label 
and incurs unit cost. Following~\citet[Definition~1]{kone2024bandit},
we define the {\em strict Pareto frontier}.
\begin{definition}[Strict Pareto frontier]
\label{def:strict-pareto-frontier}
For a collection of vectors \(x=(x_1,\ldots,x_K)\in(\mathbb R^d)^K\), 
we say that $j$ strictly dominates $i$ and 
write \(i\prec_x j\), if \(x_{i,r}<x_{j,r}\) for all \(r\in [d]\).
The strict Pareto frontier of $x$ is
\[
    \mathcal P(x)
    :=
    \{i\in[K]:\nexists j\in[K]\text{ such that }i\prec_x j\}.
\]
\end{definition}
\begin{remark}
 Our strict-dominance convention differs slightly from the standard Pareto convention, under which a model dominates another 
if it is {\em at least as good} in every coordinate and strictly better in 
at least one coordinate~\citep{auer2016pareto}.
This strict convention also has a natural decision-making interpretation: 
if a user cares only about an objective on which two models are tied, neither model should be discarded on performance grounds; token cost, availability, or other application-specific considerations may instead determine the final choice.
From a technical perspective, non-strict dominance requires additional assumptions to handle ties on individual coordinates. 
Our finite-sample termination guarantee  
requires all classification gaps defined in \Cref{def:classification-gap} 
to be positive, which may fail at coordinate-wise ties.
\end{remark}
Accordingly, our goal is to design an algorithm $\mathsf A$ 
that adaptively chooses the query $(I_t,J_t,R_t)$, 
stops at a data-dependent time $\tau_{\mathsf A}$ 
(with $\tau_{\mathsf A} = \infty$ if it never stops), 
and outputs an estimator $\hat{\mathcal{P}}$ of 
the strict Pareto frontier 
\(\mathcal P^\star:=\mathcal P(\theta)\).
For a prescribed error tolerance $\delta \in (0,1)$, 
we require the algorithm to be $\delta$-correct, i.e., 
to terminate and return the true frontier with probability at least $1-\delta$, as formalized below.
\begin{definition}[\(\delta\)-correctness]
\label{def:delta-correctness}
Let \( \Theta \subseteq (\mathbb{R}^{d})^{K}\) be an instance class. For \(\delta\in(0,1)\), an algorithm $\mathsf A$ is \(\delta\)-correct over \(\Theta\) if, for every \(\theta\in\Theta\),
\begin{equation*}
    \mathbb{P}_{\theta}
    \big(
        \tau_{\mathsf A}<\infty,
        \;
        \widehat{\mathcal{P}}
        =
        \mathcal{P}(\theta)
    \big)
    \geq
    1-\delta.
\end{equation*}
\end{definition}
Among $\delta$-correct algorithms, 
we seek to minimize the number of comparisons needed.
Since the stopping time is random, we characterize 
an algorithm's sample complexity through its upper quantile.
Specifically, for $\alpha \in (0,1)$, define 
    \[
    Q_{\theta,\alpha}(\mathsf A)
    :=
    \inf\left\{
        T\in\mathbb N_0:
        \mathbb P_\theta(\tau_{\mathsf A}\leq T)\geq1-\alpha
    \right\},
\]
with the convention \(\inf\varnothing:=\infty\). 

\subsection{Our contributions}
Our contributions are threefold.
First, we formulate fixed-confidence Pareto-set identification
under objective-specific Bradley--Terry feedback and introduce
{\em Vector-Borda Empirical-Gap Elimination (VB-EGE)}, 
an adaptive sampling procedure for this problem.
The key technical insight is 
an order-preserving reduction from the latent score vectors 
to vector-Borda scores. Each Borda coordinate is the average 
win probability of a model against all opponents on a given objective, 
and the transformation preserves the coordinate-wise ordering of the 
latent scores---and therefore the strict Pareto set.
Moreover, each Borda coordinate can be estimated directly from  
Bernoulli outcomes by comparing a model with a uniformly sampled 
opponent. This problem therefore reduces to 
Pareto-set identification over $Kd$-dimensional vectors of Bernoulli means.
VB-EGE builds on this reduction to adaptively concentrates comparison on
models whose Pareto status remains difficult to certify.

Second, we prove that VB-EGE is $\delta$-correct and 
establish an instance-dependent upper bound on its sample  
complexity. We further establish a worst-case lower bound over a
gap-separated instance class, showing that VB-EGE is minimax-optimal
with respect to the number of models, the number of objectives, and the
inverse squared classification gap, up to logarithmic factors.



Finally, we evaluate VB-EGE on synthetic Bradley--Terry instances 
and logged comparisons from three real-world Arena datasets: Arena, 
SearchArena, and VisionArena.
The synthetic experiments confirm the predicted scaling behavior 
and demonstrate the sample efficiency of VB-EGE.
Arena experiments also demonstrate substantial gains in 
sample efficiency over uniform baselines in realistic settings.




\FloatBarrier

\subsection{Related work}
\label{sec:related-work}

We review three related literatures, arena-based LLM evaluation, adaptive pairwise comparison, and Pareto set identification, 
and provide additional discussion on related work in Appendix~\ref{app:additional_related_work}.

\paragraph{Arena-based LLM evaluation.}
Chatbot Arena aggregates human pairwise preferences into scalar model
leaderboards \citep{zheng2023judging,chiang2024chatbot}. Subsequent work has improved their statistical stability by modeling ties and score dependence
\citep{ameli2025statistical}, replacing iterative Elo updates with
maximum-likelihood estimation and accounting for annotator reliability
\citep{liu2025elo}, and developing nonparametric ranking and uncertainty
quantification \citep{frauen2026nonparametric}. Adaptive model-pair selection improves comparison efficiency \citep{liu2026fewer}, while prompt- or task-specific rankings address heterogeneity \citep{frick2025prompt,li2026low}. These methods still produce scalar rankings, either overall or conditional on prompts or tasks. We instead identify the set of models that remain nondominated across multiple objectives.

\paragraph{Pure exploration from pairwise comparisons.}
Pairwise-comparison methods study winner and ranking identification
without directly observing item utilities.
Random-opponent comparisons lead to Borda scores and have been used for
Borda-winner identification and active ranking
\citep{jamieson2015sparse,heckel2019active};
simple Borda-type procedures can also be robust beyond parametric
Bradley--Terry models \citep{shah2018simple}.
More recently, \citet{goldberger2026optimal}
developed an asymptotically optimal fixed-confidence method for top-$k$
identification under a latent-utility comparison model.

Beyond these scalar targets, comparison-based methods have also been used
for Pareto or skyline recovery.
\citet{asudeh2015crowdsourcing} adaptively select
criterion-specific crowd comparisons to identify Pareto-optimal objects,
\citet{groz2020skyline} study skyline recovery under noisy
coordinate comparisons, and \citet{audiffren2017bandits} identify maximal arms from dueling feedback
over a partial order.
Our setting instead considers objective-specific Bradley--Terry
comparisons generated by latent utilities and uses an order-preserving
vector-Borda reduction for Pareto-set identification.

\paragraph{Pareto set identification.}
Pareto-set identification has primarily been studied in multi-objective
bandits where pulling an arm directly yields a vector-valued observation.
Existing work considers approximate and relaxed fixed-confidence recovery
\citep{auer2016pareto,kone2023adaptive}, fixed-budget identification through
Empirical-Gap Elimination and its cost-aware extension \citep{kone2024bandit,xue2026costaware}, and exact fixed-confidence identification with asymptotically optimal allocation \citep{garivier2024sequential}. Structured extensions include preference-cone and multi-output linear models \citep{karagozlu2024learning,kone2025bandit,
shukla2024preference,das2026frappe}.
Our elimination rule is closest to \citep{kone2024bandit}, but our feedback is objective-specific and binary rather than a directly observed vector.

\FloatBarrier

\section{Vector-Borda Empirical-Gap Elimination (VB-EGE)}
\label{sec:method}

In this section, we present our proposed algorithm, VB-EGE, 
which adaptively allocates the comparisons to 
identify the Pareto frontier.
The algorithm
first represents the ordering of the models 
along each objective coordinate using the {\em Borda scores} (to be defined shortly),
and then applies the empirical-gap elimination principle~\citep{kone2024bandit}
to the vector of Borda scores to determine the sampling and stopping rule.
We first introduce the notation used in the remainder of the paper.

\paragraph{Notation.}
For any $x\in \mathbb{R}$, denote $x^+ = \max\{x,0\}$.
For $a,b\in\mathbb R$, write $a\wedge b:=\min\{a,b\}$ and
$a\vee b:=\max\{a,b\}$.
For two nonnegative functions $f$ and $g$, we write $f = O(g)$ (or $f \lesssim g$) 
if there exists a universal constant 
$C > 0$ such that $f \le C g$; we write $f = \Omega(g)$ (or $f \gtrsim g$) if $g = O(f)$, 
and $f = \Theta(g)$ (or $f \asymp g$) if both $f = O(g)$ and $f = \Omega(g)$ hold.
The notation $f=\tilde O(g)$, $f = \tilde \Omega (g)$, and $f = \tilde \Theta(g)$ 
is defined analogously, up to logarithmic factors.
For \(p,q\in[0,1]\), define the binary relative entropy
$\kl(p,q)
    :=
    p\log\frac{p}{q}
    +(1-p)\log\frac{1-p}{1-q}$,
with the conventions \(0\log(0/x):=0\) for \(x\in[0,1]\) and \(x\log(x/0):=+\infty\) for \(x>0\).

\subsection{The vector-Borda reduction}
\label{sec:vector-borda}
We first define the coordinate-wise Borda scores that are central to the algorithm.
\begin{definition}[Coordinate-wise Borda scores]
\label{def:borda-vector}
For each model \(i\in[K]\) and objective \(r\in[d]\), define 
its Borda score as
\begin{equation}
    b_{i,r}(\theta)
    :=
    \frac{1}{K-1}
    \sum_{j\neq i}
    p_{ij}^{(r)}(\theta)
    =
    \frac{1}{K-1}
    \sum_{j\neq i}
    \sigma(\theta_{i,r}-\theta_{j,r}).\label{eq:vector-borda-coordinate}
\end{equation}
When the dependence on $\theta$ is clear from the context, 
we write $b_{i,r}$ in place of $b_{i,r}(\theta)$.
The {\em vector-Borda representation} of model \(i\) is
\(
    b_i
    :=
    (b_{i,1},\ldots,b_{i,d})
    \in[0,1]^d,
\)
and we write
\(
    b=(b_1,\ldots,b_K).
\)
\end{definition}
The key property of the vector-Borda transformation is that 
it preserves the ordering of the models along each coordinate
induced by the latent scores $\theta$. This property is 
formalized by the following lemma, with its proof provided in~Appendix~\ref{app:borda-order-preservation-proof}.

\begin{lemma}[Coordinate-wise order preservation]
\label{lem:borda-order-preservation}
For every \(i,j\in[K]\) and \(r\in[d]\),
\begin{equation}
    \operatorname{sign}(b_{i,r}-b_{j,r})
    =
    \operatorname{sign}(\theta_{i,r}-\theta_{j,r}).
    \label{eq:borda-order-preservation}
\end{equation}
Consequently,
\begin{equation}
    i\prec_b j
    \quad\Longleftrightarrow\quad
    i\prec_{\theta} j
    \quad 
    \text{ and }
    \quad
    \mathcal{P}(b)
    =
    \mathcal{P}(\theta)
    =
    \mathcal{P}^{\star}.
    \label{eq:borda-pareto-preservation}
\end{equation}
\end{lemma}
By~\Cref{lem:borda-order-preservation},
the latent strict Pareto frontier coincides with 
the Pareto frontier of the vector-valued Bernoulli means \(b_1,\ldots,b_K\). 
Hence, the learner can operate entirely in Borda space without estimating 
the latent score vectors \(\theta_1,\ldots,\theta_K\).

For a selected focal model $i$ and objective $r$, if we draw
$J\sim\mathrm{Unif}([K]\setminus\{i\})$ using fresh randomness,
issue the query $Q_t=(i,J,r)$, and observe $Z:=Y_t^{(r)}(i,J)$,
then conditional on the past and the selected $(i,r)$, averaging over $J$
gives \(Z\sim\operatorname{Bernoulli}(b_{i,r})\).
This observation underlies our proposed sampling rule 
and  
is the coordinate-wise analogue of the standard random-opponent Borda reduction \citep[Definition~1]{jamieson2015sparse}.

\subsection{Pareto classification gaps}
\label{sec:borda-preservation-gap}
Next, we define classification gaps that measure how difficult it is to determine each model’s Pareto status, while accounting for its role in identifying other models as non-Pareto.
For $x= (x_1,\ldots,x_K)\in (\mathbb R^d)^K$ and any models $i,j$, define
\[
    m_x(i,j)
    :=
    \min_{r\in[d]}(x_{j,r}-x_{i,r}),
    \qquad
    M_x(i,j)
    :=
    \max_{r\in[d]}(x_{i,r}-x_{j,r}).
\]
Thus,
\( M_x(i,j)=-m_x(i,j) \), and \(i\prec_x j \Longleftrightarrow m_x(i,j)>0\).

Our algorithm follows a successive-elimination scheme, 
where models are sequentially removed from an active set. Subsequent comparisons use only
remaining models as focal models, while opponents are drawn uniformly from
the full original set of models excluding the focal model. We therefore
define the classification gaps 
relative to an arbitrarily given set $S \subseteq [K]$
to characterize the difficulty of identifying the
Pareto frontier from the remaining models.
First, the strict Pareto set restricted to \(S\) is given by 
\[
\mathcal{P}_S(x)
:=
\left\{
i\in S:
\nexists j\in S\setminus\{i\}
\text{ such that }m_x(i,j)>0
\right\}.
\]
The definition of the classification gap is then given as follows.
\begin{definition}[Restricted-set classification gaps]
\label{def:classification-gap}
Let \(S\subseteq[K]\) with \(|S|\geq2\), and let
\(x_i\in\mathbb{R}^d\) for every \(i\in S\).
For every \(i\in S\), define the raw domination gap
\(
    \Delta_{i,S}^\star(x)
    :=
    \max_{j\in S\setminus\{i\}}m_x(i,j),
\)
and the auxiliary gap
\[
    \delta_{i,S}^\star(x) := \min_{j\in S\setminus\{i\}} \left\{ M_x(i,j) \wedge \left[ M_x(j,i)^+ + \bigl(\Delta_{j,S}^\star(x)\bigr)^+ \right] \right\}.
\]
The restricted-set classification gap is
\begin{equation}
    \Delta_{i,S}(x)
    :=
    \begin{cases}
        \Delta_{i,S}^\star(x),
        & i\notin \mathcal{P}_S(x), \\[1mm]
        \delta_{i,S}^\star(x),
        & i\in \mathcal{P}_S(x).
    \end{cases}
    \label{eq:restricted-classification-gap}
\end{equation}
\end{definition}
Intuitively, for $i \notin \mathcal{P}_S(x)$, $\Delta^\star_{i,S}(x)$
measures the largest domination margin certifying $i$
is non-Pareto. For $i\in \mathcal{P}_S(x)$, 
$\delta^\star_{i,S}(x)$ is the smallest separation needed to 
certify $i$'s Pareto status or to correctly classify another model
relative to $i$. Thus, a larger
\(\Delta_{i,S}(x)\) indicates estimating model $i$ for identifying 
$\mathcal P_S(x)$ is easier.

In the following, we omit the set subscript when \(S=[K]\) and write
$\mathcal P(x):=\mathcal P_{[K]}(x)$,
$\Delta_i^\star(x):=\Delta_{i,[K]}^\star(x)$,
$\delta_i^\star(x):=\delta_{i,[K]}^\star(x)$,
$\Delta_i(x):=\Delta_{i,[K]}(x)$.
For the vector-Borda and latent representations, respectively, let
$\Delta_i^{\Borda}:=\Delta_i(b)$,
$\Delta_i^\theta:=\Delta_i(\theta)$,
and
    $\Delta_{\min}^{\Borda}
    :=
    \min_{i\in[K]}\Delta_i^{\Borda}$.

\subsection{Phased vector-Borda elimination}
\label{sec:vbege-algorithm}
VB-EGE proceeds in phases, maintaining an active set \(A\) of models 
whose Pareto status remains undetermined
and an accepted set \(B\) of models certified as Pareto.
At each phase $m \ge 1$, for each pair $(i,r) \in A\times [d]$, 
the algorithm collects $n_m - n_{m-1}$ new observations by sampling
\begin{align}\label{eq:phase-sample}    
Z_{i,r,s} := Y_{t(i,r,s)}^{(r)}(i,J_{i,r,s}),
\text{ where }
J_{i,r,s} \overset{\text{i.i.d.}}{\sim} \text{Unif}([K]\backslash\{i\}), 
\quad \forall s = n_{m-1}+1, \dots, n_m,
\end{align} 
where $t(i,r,s)$ denotes the global round at which 
the $s$th comparison associated with model $i$ and coordinate $r$
is conducted, $n_0 = 0$, and $\{n_m\}_{m \ge 1}$ is an increasing sequence of  
sample-size parameters specified below.
Using all the observations collected through phase $m$, 
the algorithm updates the vector-Borda estimates as 
\begin{align*}
\widehat b_{i,r}^{(m)} := \frac{1}{n_m} \sum^{n_m}_{s=1} Z_{i,r,s}, 
\quad (i,r) \in A \times [d].
\end{align*}
Write $\widehat b_i^{(m)}=(\widehat b_{i,1}^{(m)},\ldots,\widehat b_{i,d}^{(m)})$;
all empirical Pareto sets and gaps below are evaluated using these vectors
for models in the current active set $A$.
It then updates the restricted-set Pareto frontier and the 
corresponding classification gap with respect to the current active set $A$: 
\begin{equation}
\widehat{\mathcal S}_A
:=
\mathcal P_A\!\big(\widehat b^{(m)}\big),
\qquad
\widehat\Delta_{i,A}
:=
\Delta_{i,A}\!\big(\widehat b^{(m)}\big),
\quad i\in A.
\label{eq:empirical-active-quantities}
\end{equation}
If all the estimated classification gaps in $A$ are small---that is, 
$\max_{i\in A} \widehat \Delta_{i,A} \le C_\thr r_m$,
where $C_\thr$ is a constant and $r_m$ is a confidence-radius parameter 
specified below---then the available evidence is not sufficient 
to certify the status of any additional model and 
the algorithm proceeds to the next phase without elimination.
Otherwise, it selects $i_m \in \arg\max_{i\in A}\widehat\Delta_{i,A}$
and removes it from $A$. The selected model $i_m$ is chosen anew at each
iteration of the inner loop within phase $m$, with ties resolved by prioritizing empirically
non-Pareto models, as specified in Algorithm~\ref{alg:vbege}.
If $i_m \in \widehat{\mathcal S}_A$, it is certified as Pareto and
added to $B$; otherwise, it is rejected as non-Pareto.
After each removal, if only one active model remains, the algorithm returns $B\cup A$.
Otherwise, it recomputes the empirical Pareto set and classification gaps
over the updated active set and repeats the removal step, using the same
phase-$m$ estimates without additional sampling.
It proceeds to phase $m+1$ only when no empirical gap exceeds the threshold.



The sampling schedule and removal thresholds are chosen so that, with high probability, 
every model added to $B$ is truly Pareto and every rejected model is truly non-Pareto. 
To be specific, at each phase 
$m\ge 1$, we define 
\begin{equation}
    \varepsilon_m := 2^{-m}, \qquad L_m := \log\frac{C_{\log}Kdm^2}{\delta}, \qquad n_m := \left\lceil \frac{C_{\mathrm{samp}}L_m}{\varepsilon_m^2} \right\rceil, \qquad r_m := \sqrt{ \frac{L_m}{2n_m} },
    \label{eq:vbege-schedule}
\end{equation}
where \( C=(C_{\mathrm{samp}},C_{\mathrm{thr}},C_{\log})\) are hyperparameters
controlling the sampling rate, removal threshold, and confidence factor of the algorithm. 
Unless stated otherwise, we use the default configuration
\begin{equation}
C=(C_{\mathrm{samp}},C_{\mathrm{thr}},C_{\log})=(2,4,4).    
\label{eq:theory-constants}
\end{equation}
The complete procedure is summarized in Algorithm~\ref{alg:vbege}.

\begin{algorithm}[h!]
\caption{Phased Vector-Borda Empirical-Gap Elimination \(\VBEGE(\delta;C)\)}
\label{alg:vbege}
\begin{algorithmic}[1]
\Require number of candidate models \(K\ge2\), 
number of objectives \(d\ge1\), 
error tolerance level \(\delta\in(0,1)\), and 
sampling hyperparameters \(C=(C_{\mathrm{samp}},C_{\mathrm{thr}},C_{\log}) \in (0,\infty)^2\times[1,\infty)\).

\State \textbf{Initialization:} \(A\gets[K]\), \(B\gets\varnothing\), \(n_0\gets0\).

\For{\(m=1,2,\ldots\)}
    \State Set \(\varepsilon_m,L_m,n_m,r_m\) according to~\eqref{eq:vbege-schedule}.

    \For{\((i,r)\in A\times[d]\)}
        \State Draw \(\{Z_{i,r,s}\}_{s=n_{m-1}+1}^{n_m}\) 
        according to~\eqref{eq:phase-sample}, and compute \(\displaystyle \widehat b_{i,r}^{(m)} \gets \frac1{n_m}\sum_{s=1}^{n_m}Z_{i,r,s} \).
    \EndFor
    
    \State Compute \(\widehat{\mathcal S}_A\) and \(\{\widehat\Delta_{i,A}\}_{i\in A}\) with \(\{\widehat b_i^{(m)}:i\in A\}\) according to \eqref{eq:empirical-active-quantities}.

    \While{\(|A|\ge2\) and \(\max_{i\in A}\widehat\Delta_{i,A}>C_{\mathrm{thr}}r_m\)}
        \State \(G\gets\arg\max_{i\in A}\widehat\Delta_{i,A}\).
        \State Choose \(i_m\in G\setminus\widehat{\mathcal S}_A\) if \(G\setminus\widehat{\mathcal S}_A\neq\varnothing\); otherwise choose \(i_m\in G\).
        \If{\(i_m\in\widehat{\mathcal S}_A\)}
            \State \(B\gets B\cup\{i_m\}\)
        \EndIf
        \State \(A\gets A\setminus\{i_m\}\).
        \State \textbf{if} \(|A|=1\) \textbf{ then return} \(B\cup A\).
        \State Recompute \(\widehat{\mathcal S}_A\) and  \(\{\widehat\Delta_{i,A}\}_{i\in A}\) over the updated \(A\) 
        with \(\{\widehat b_i^{(m)}:i\in A\}\).
    \EndWhile
\EndFor
\end{algorithmic}
\end{algorithm}

\begin{remark}[Sampling and tie-breaking conventions]
\label{rem:vbege-conventions}
VB-EGE samples only active models as focal models, but draws each opponent uniformly from the full original set of models \([K]\setminus\{i\}\), including previously removed models. Thus the mean of each focal observation remains the fixed global quantity $b_{i,r}$. 
When multiple models maximize \(\widehat\Delta_{i,A}\), VB-EGE 
prioritizes a maximizer outside the empirical Pareto set \(\widehat{\mathcal S}_A\), whenever one exists; otherwise, it selects an arbitrary maximizer.
On the uniform confidence event in
\Cref{lem:uniform-borda-confidence}, this tie-breaking rule ensures that each active non-Pareto model \(i\) retains an active Pareto model attaining its domination gap \(\Delta_i^{\Borda}\).
\end{remark}


\FloatBarrier

\section{Theoretical guarantees}
\label{sec:theory}

This section presents the theoretical guarantees for VB-EGE.
Recall that, for any algorithm $\mathsf A$, $\tau_{\mathsf A}$ is 
the total number of queries made before stopping.  
Each query compares two models with respect to 
a single objective and returns a binary outcome.
We write \(\tau=\tau_{\mathsf A}\) when the algorithm is clear from the context, 

For \(B,\Delta>0\), define the bounded-range class and its gap-separated subclass by
\begin{equation}
\begin{aligned}
    \Theta_{B,d}
    &:= 
    \left\{
        \theta\in(\mathbb R^d)^K:
        \max_{r\in[d]}
        \max_{i,j\in[K]}
        |\theta_{i,r}-\theta_{j,r}|
        \le 2B
    \right\}, \\[1mm]
    \Theta_{B,d}(\Delta)
    &:=
    \left\{
        \theta\in\Theta_{B,d}:
        \min_{i\in[K]}\Delta_i^\theta\ge\Delta
    \right\}.
\end{aligned}
\label{eq:theory-instance-classes}
\end{equation}
We also write \[
    \kappa_B
    :=
    \inf_{|x|\le2B}\sigma'(x)
    =
    \sigma(2B)\bigl(1-\sigma(2B)\bigr).
\]
In what follows, 
Section~\ref{sec:upper-bound} establishes the $\delta$-correctness 
of VB-EGE and an instance-dependent upper bound
on its sample complexity over the instance class $\Theta_{B,d}$.
Section~\ref{sec:lower-bound} provides a worst-case 
lower bound over $\Theta_{B,d}(\Delta)$. Together, 
these results show that 
VB-EGE is minimax-optimal over $\Theta_{B,d}(\Delta)$ 
up to logarithmic factors.

\subsection{Upper bound}
\label{sec:upper-bound}
The following theorem establishes the $\delta$-correctness 
of our proposed algorithm and an upper bound on 
the quantile of its stopping time.

\begin{theorem}[Quantile upper bound]
\label{thm:upper}
Fix \(B>0\) and \(\theta\in\Theta_{B,d}\) 
satisfying \(\min_{i\in[K]}\Delta_i^\theta>0\). 
Define for each $i\in[K]$ 
\[
    M_i^\theta
    :=
    1+
    \left\lceil
        \log_2\frac{4}{\kappa_B\Delta_i^\theta}
    \right\rceil
    ,\qquad
    L_i^\theta(\delta)
    :=
    \log\frac{4Kd(M_i^\theta)^2}{\delta}.
\]
Then VB-EGE (Algorithm~\ref{alg:vbege}) implemented 
with $\delta \in (0,1)$ and $C = (2,4,4)$ satisfies
\[
    \mathbb P_\theta\!\left(
        \widehat{\mathcal P}=\mathcal P(\theta),
        \tau
        \le
        dK+
        \frac{128d}{\kappa_B^2}
        \sum_{i=1}^K
        \frac{L_i^\theta(\delta)}
        {(\Delta_i^\theta)^2}
    \right)
    \ge1-\delta.
\]
Consequently, VB-EGE is $\delta$-correct on the positive-gap subclass of \(\Theta_{B,d}\), and  
\[
    Q_{\theta,\delta}(\VBEGE)
    \le
    dK+
    \frac{128d}{\kappa_B^2}
    \sum_{i=1}^K
    \frac{L_i^\theta(\delta)}
    {(\Delta_i^\theta)^2}
    =
    \widetilde O\!\left(
        \frac{d}{\kappa_B^2}
        \sum_{i=1}^K(\Delta_i^\theta)^{-2}
    \right).
\]
If \(B\) is bounded by an absolute constant, then
\(\kappa_B\) is bounded below by a positive absolute constant.
\end{theorem}
The proof of Theorem~\ref{thm:upper} is provided in~\ref{app:upper-bound-proof},
in which it first establishes exact recovery and finite stopping in Borda space, 
and then converts the bound to latent-score gaps.

\subsection{Lower bound}
\label{sec:lower-bound}
We now turn to the lower bound on the sample complexity
of any $\delta$-correct algorithm.
For the gap-separated instance class $\Theta_{B,d}(\Delta)$, 
define the minimax $(1-\delta)$-quantile sample complexity as
\[
    \mathbf Q_{B,d}(\Delta,\delta)
    :=
    \inf_{\mathsf A:\,
        \delta\text{-correct on }\Theta_{B,d}(\Delta)}
    \;
    \sup_{\theta\in\Theta_{B,d}(\Delta)}
    Q_{\theta,\delta}(\mathsf A).
\]
Thus, \(\mathbf Q_{B,d}(\Delta,\delta)\) is the smallest worst-case
\((1-\delta)\)-stopping-time quantile achievable by any
\(\delta\)-correct algorithm on \(\Theta_{B,d}(\Delta)\).
The following theorem gives a lower bound on $\mathbf Q_{B,d}(\Delta,\delta)$.

\begin{theorem}[Gap-separated minimax lower bound]
\label{thm:minimax-optimality}
Fix \(B>0\), \(0<\Delta\le\min\{1,B\}\), and \(0<\delta<1/3\).  There is
\[
    \mathbf Q_{B,d}(\Delta,\delta)
\ge 
\frac{d(K-1)}{\Delta^2}
\kl(1-2\delta,\delta).
\]
\end{theorem}
The proof of Theorem~\ref{thm:minimax-optimality} is deferred to Appendix~\ref{app:minimax-proof}. Here, we sketch the high-level idea:
the proof considers the hard instance, where for some $i\in [K]$, 
\begin{align}
\label{eq:hard-inst}
\theta_i = (0,\dots,0), \quad \theta_j = (-\Delta,\dots,-\Delta), \quad \forall j\in [K]\backslash \{i\}.
\end{align}
Raising any  model other than $i$ to $+\Delta$ on a single objective makes it Pareto, while preserving the gap and range constraints. A correct algorithm must rule out all $d(K-1)$ such possibilities.

Finally, applying Theorem 3.1 with \(\Delta_i^\theta \ge \Delta\) for every \(i\) and combining it with Theorem 3.2 yields the following minimax guarantee for fixed \(B\), up to logarithmic factors. 

\begin{corollary}[Gap-separated minimax optimality]
\label{cor:minimax-opt}
Under the same assumptions as in Theorem~\ref{thm:minimax-optimality}, 
\[
\mathbf Q_{B,d}(\Delta,\delta)
\le
\sup_{\theta\in\Theta_{B,d}(\Delta)}
Q_{\theta,\delta}(\VBEGE)
\le
dK+
\frac{128dK}{\kappa_B^2\Delta^2}
\log\frac{4Kd(\lceil\log_2 \frac{8}{\kappa_B \Delta} \rceil)^2}{\delta}.
\]
Consequently, for any fixed \(B>0\),
\[
    \mathbf Q_{B,d}(\Delta,\delta)
    =
    \widetilde\Theta\!\left(
        \frac{dK}{\Delta^2}
        \log\frac1\delta
    \right)
    \qquad\text{as }\delta\downarrow0,
\]
where \(\widetilde\Theta\) suppresses logarithmic factors in \(K\), \(d\), and \(\Delta^{-1}\). In particular, VB-EGE attains the minimax rate up to these logarithmic factors.
\end{corollary}

\section{Experiments}
\label{sec:experiments}
This section evaluates the scaling and sample efficiency of VB-EGE on synthetic data, 
as well as its sample efficiency on logged Arena comparisons. 

\subsection{Synthetic experiments}
\label{sec:synthetic-experiments}
In the scaling experiments, we generate data from the BT model 
corresponding to the hard instance in~\eqref{eq:hard-inst} and  
vary the values of $K,d,\Delta,\delta$.
Separately, we evaluate the methods on eight synthetic benchmarks 
generated from four families, which we call 
{\em Convex, Arena, Witness}, and {\em Two-group}, respectively.
Details of the data-generating process in both experiments are provided in Appendix~\ref{app:synthetic-benchmark-details}.


\begin{table}[htbp]
\centering
\caption{Parameters used in the scaling experiments.}
\label{tab:synthetic-main-settings}
\footnotesize
\begin{tabular}{@{}p{0.15\textwidth}p{0.69\textwidth}r@{}}
\toprule
Study & Parameter grid  \\
\midrule
Scaling in $K$ & \(K\in\{16,32,64,128\}\) at \(d=4,\Delta=1\)\\
Scaling in $d$ & \(d\in\{2,4,10\}\) at \(K=64,\Delta=1\)\\
Scaling in $\Delta$ & \(\Delta\in\{0.5,0.75,1,1.25,1.5\}\) at \(K=64,d=10\)  \\ 
Scaling in $\delta$ & \(\delta\in\{0.2,0.1,0.05,0.02,0.01,0.005,0.002,0.001\}\) at \(K=64,d=10,\Delta=1\) \\
\bottomrule
\end{tabular}
\end{table}

We compare VB-EGE with three baselines using uniform sampling:
\begin{enumerate}[label=(\arabic*)]
\item \emph{Focal Borda}. At each phase, 
the algorithm collects new observations for {\em every} model-objective pair
$(i,r)\in[K]\times [d]$ and performs no elimination;
otherwise, it uses the same sampling schedule $\{n_m\}_{m\ge 1}$, 
Borda estimates, and certification rule as VB-EGE.

\item \emph{BT-MLE}. The algorithm uniformly samples from 
all \(dK(K-1)/2\) pair--objective tuples $(i,j,r)$, where $1 \le i<j\le K$ and $r\in[d]$.
At the end of each phase, it fits a separate ridge-regularized BT model for each objective
to estimate the latent scores. As a plug-in heuristic, 
it substitutes these estimates into the classification-gap function and 
compares the resulting gaps with the same threshold 
as VB-EGE to determine the estimated Pareto set and whether to stop.

\item \emph{Borda plug-in}. The algorithm uses the same sampling rule as BT-MLE 
but estimates the Borda scores directly using smoothed empirical win rates, 
whose exact definition is given in Appendix~\ref{app:synthetic-protocol}. The stopping and Pareto certification rule
follows from VB-EGE.

\end{enumerate}

All methods use $\delta=0.05$ and $C=(2,4,4)$, except when 
we evaluate the scaling with respect to $\delta$.
\Cref{tab:synthetic-main-settings} summarizes the parameter grids, 
with further implementation details in Appendix~\ref{app:synthetic-protocol}.
Each experiment is repeated over $500$ independent trials, and we report 
the mean stopping time, measured in objective-specific comparisons, 
and proportion of trials in which the Pareto set is 
incorrectly classified.

\begin{figure}[!t]
\centering
\includegraphics[width=\linewidth,trim={0bp 2bp 0bp 0.5bp},clip]{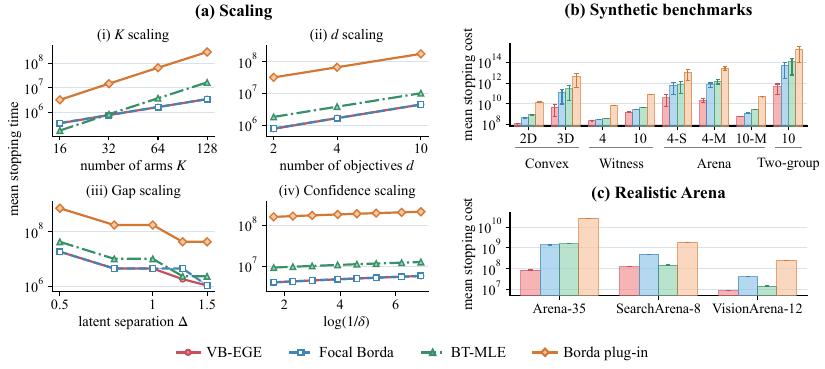}%
\begin{subcaptiongroup}%
\phantomcaption\label{fig:symmetric-hard-scaling}%
\phantomcaption\label{fig:synthetic-benchmark-overview}%
\phantomcaption\label{fig:realistic-main-benchmark}%
\end{subcaptiongroup}%
\caption{Mean stopping time (mean \(\pm\) one standard error) on logarithmic vertical axes. 
(a) Scaling in \(K\), \(d\), \(\Delta\), and \(\log(1/\delta)\). (b) Eight heterogeneous synthetic Pareto benchmark settings; S and M denote small and medium. (c) Three realistic Arena datasets, with stopping times measured in comparisons.}
\label{fig:synthetic-main-results}
\end{figure}

\paragraph{Results.}
Across all synthetic settings, we observe no errors in identifying
the Pareto set.
In Figure~\ref{fig:symmetric-hard-scaling}, the empirical log--log slopes of 
the mean stopping time with respect to $K$ and $d$ are both $1.09$ for VB-EGE,
consistent with the scaling predicted by our theory.
The mean stopping time also decreases as $\Delta$ increases and 
is nearly linear in $\log(1/\delta)$, with $R^2 \approx 0.99997$.
In these experiments, VB-EGE and Focal Borda have nearly identical performance, 
which is as expected because all dominated models are equally difficult to classify,
leaving little to gain from adaptive elimination.

Figure~\ref{fig:synthetic-benchmark-overview} shows the results 
from the eight heterogeneous benchmarks. 
VB-EGE achieves the lowest 
mean stopping cost in all settings.
For each baseline, we compute the ratio between its mean stopping time and that of VB-EGE.
Across the eight settings, the resulting ratio ranges from \(1.25\) to \(111\) for Focal Borda,
from \(1.79\) to \(261\) for BT-MLE, and
from \(29.7\) to \(4.23\times10^3\) for Borda plug-in. 
Appendix~\ref{app:synthetic-experiments} presents additional results and diagnostics.

\subsection{Realistic Arena data experiments}
\label{sec:realistic-experiments}

We consider Arena-Human-Preference-140K (Arena-140K) \citep{chiang2024chatbot},
Search-Arena-24K \citep{miroyan2026search},
and VisionArena-Battle \citep{chou2025visionarena},
treating the released task tags, 
search intents, and vision categories as objectives, respectively.
Within each dataset, we retain only decisive comparisons---those in which 
either model is labeled the winner---and exclude the ties and ``both bad'' outcomes.
We then select the largest subset of models such that  
every pair has at least one retained comparison for every objective.
Additional preprocessing details are in Appendix~\ref{app:real-data-protocol}.

\begin{table}[htbp]
\centering
\caption{Main realistic Arena data settings. Rows are retained pair--objective records; the VisionArena-Battle records arise from \(12{,}937\) unique multi-label comparisons.}
\label{tab:realistic-main-settings}
\experimenttablesize
\begin{tabular}{@{}lrrrrr@{}}
\toprule
Dataset & Objectives \(d\) & Candidates & Retained \(K\) & Rows & \(\lvert\mathcal P_{\rm emp}\rvert\) \\
\midrule
Arena-140K        & 5 & 52 & 35 & \(73{,}135\) & 3 \\
Search-Arena-24K  & 9 & 13 &  8 & \(4{,}305\)  & 5 \\
VisionArena-Battle& 8 & 17 & 12 & \(23{,}458\) & 3 \\
\bottomrule
\end{tabular}
\end{table}

For each model and objective, we compute a Borda score by averaging its logged win rates against all other retained models, and 
take the strict Pareto frontier of the resulting Borda vectors as the {\em empirical ground truth}.
For VB-EGE and Focal Borda, each query selects a focal model and an objective, then draws an opponent uniformly from the other retained models. BT-MLE and the Borda plug-in sample unordered pair--objective cells uniformly. A binary outcome is then drawn uniformly {\em with replacement} from the selected cell (see Appendix~\ref{sec:arena-replay}). We evaluate VB-EGE and the three baseline methods 
described in Section~\ref{sec:synthetic-experiments}, using
\(\delta=0.05\), \(C=(2,4,4)\). Each experiment is repeated over $100$ independent trials.

\paragraph{Results.} Across all experiments, 
every method stopped and correctly identified the empirical Pareto frontier in every trial.
Figure~\ref{fig:realistic-main-benchmark} compares the mean stopping time, 
where VB-EGE achieves the lowest value on all three datasets.
The ratio of mean stopping times (baseline/VB-EGE) 
ranges from \(3.96\) to \(16.54\) for Focal Borda, from \(1.20\) to \(18.27\) for BT-MLE, 
and from \(15.20\) to \(301.36\) for Borda plug-in. 
Additional results and discussion are provided in Appendix~\ref{app:realistic-pareto-details} and Appendix~\ref{sec:arena-robustness}.

\FloatBarrier

\section{Limitations and future work}
\label{sec:future-work}
This paper introduced a fixed-confidence adaptive sampling framework 
for identifying Pareto-optimal models from pairwise comparison. 
Several directions remain for future work. 
First, although our guarantees are stated under the BT model, the order-preservation property 
of the vector-Borda reduction requires only that the comparison probability be strictly increasing in the latent score difference.
The reduction therefore applies more broadly to monotone comparison models. 
Second, our Bradley--Terry theory considers only decisive comparisons and does not model
the \texttt{tie} and \texttt{both\_bad} labels in Chatbot Arena \citep{chiang2024chatbot}.
Appendix~\ref{sec:arena-robustness} empirically shows that the Borda-level guarantee extends to half-credit observations and uses a fitted Davidson construction to assess target sensitivity. A model-based treatment of non-decisive outcomes with a recovery guarantee for the population Pareto frontier remains open.
Third, the Arena experiments use several forms of objective labeling, 
including mutually exclusive groups, primary-intent labels, and overlapping categories. 
An important open question is how sensitive the identified frontiers are to these choices. 
One extension is to allow soft or learned assignments, where each comparison context is associated with a mixture of capabilities.

Beyond VB-EGE, there are also several algorithmic directions worth investigating. 
For example, under the BT model, one could estimate the latent scores and guide 
sampling using likelihood-based methods.
Appendix~\ref{app:pareto-btglr-comparison} explores Pareto BT-GLR Track-and-Stop as a screened implementation of this approach, 
leaving a rigorous theoretical analysis for future work.
Another promising direction is graph-aware sampling that exploits nonuniform historical support when selecting opponents and objectives. Such methods would be particularly relevant for larger, evolving arenas in which comparison coverage is uneven and new models are introduced over time.

\subsection*{Reproducibility}
Code for reproducing the results in this paper is available at \url{https://github.com/Twist-Shan/Beyond_Scalar_Leaderboards}. Dataset links, pinned revisions, and preprocessing details are
provided in Appendix~\ref{app:real-data-protocol}.


\subsection*{AI Use Statement}
In this work, we used generative AI tools (including ChatGPT 6 Astra, Claude Opus 5.5, etc.) for organizing the paper outline, polishing the writing, and structuring the appendices; searching for relevant literature; implementing synthetic-data generation and sampling from realistic Arena datasets; and, after the upper-bound proof had been established, searching for and assisting with the derivation of a matching lower bound.
We have not used generative AI tools for designing the conceptual frameworks, proposing or refining hypotheses or interpreting results. We have reviewed all technical claims, proofs, experimental designs, implementations, and reported results. We take responsibility for the final content of this work, including text, claims or artifacts produced with the aid of generative AI.

\bibliographystyle{reference_style}
\bibliography{references}

@inproceedings{zheng2023judging,
  title = {Judging {LLM-as-a-Judge} with {MT-Bench} and {Chatbot Arena}},
  author = {Zheng, Lianmin and Chiang, Wei-Lin and Sheng, Ying and Zhuang, Siyuan and Wu, Zhanghao and Zhuang, Yonghao and Lin, Zi and Li, Zhuohan and Li, Dacheng and Xing, Eric P. and Zhang, Hao and Gonzalez, Joseph E. and Stoica, Ion},
  booktitle = {Advances in Neural Information Processing Systems},
  volume = {36},
  pages = {46595--46623},
  publisher = {Curran Associates, Inc.},
  year = {2023},
  doi = {10.52202/075280-2020}
}

@inproceedings{chiang2024chatbot,
  title = {{Chatbot Arena}: An open platform for evaluating {LLM}s by human preference},
  author = {Chiang, Wei-Lin and Zheng, Lianmin and Sheng, Ying and Angelopoulos, Anastasios Nikolas and Li, Tianle and Li, Dacheng and Zhu, Banghua and Zhang, Hao and Jordan, Michael and Gonzalez, Joseph E. and Stoica, Ion},
  booktitle = {Proceedings of the 41st International Conference on Machine Learning},
  series = {Proceedings of Machine Learning Research},
  volume = {235},
  pages = {8359--8388},
  publisher = {PMLR},
  year = {2024}
}

@article{bradley1952rank,
  title = {Rank analysis of incomplete block designs: {I}. {The} method of paired comparisons},
  author = {Bradley, Ralph Allan and Terry, Milton E.},
  journal = {Biometrika},
  volume = {39},
  number = {3--4},
  pages = {324--345},
  year = {1952},
  doi = {10.1093/biomet/39.3-4.324}
}

@inproceedings{kone2024bandit,
  title = {Bandit {Pareto} set identification: The fixed budget setting},
  author = {Kone, Cyrille and Kaufmann, Emilie and Richert, Laura},
  booktitle = {Proceedings of the 27th International Conference on Artificial Intelligence and Statistics},
  series = {Proceedings of Machine Learning Research},
  volume = {238},
  pages = {2548--2556},
  publisher = {PMLR},
  year = {2024}
}

@inproceedings{garivier2024sequential,
  title = {Sequential learning of the {Pareto} front for multi-objective bandits},
  author = {Cr\'{e}pon, \'{E}lise and Garivier, Aur\'{e}lien and Koolen, Wouter M.},
  booktitle = {Proceedings of the 27th International Conference on Artificial Intelligence and Statistics},
  series = {Proceedings of Machine Learning Research},
  volume = {238},
  pages = {3583--3591},
  publisher = {PMLR},
  year = {2024}
}

@inproceedings{auer2016pareto,
  title = {{Pareto} front identification from stochastic bandit feedback},
  author = {Auer, Peter and Chiang, Chao-Kai and Ortner, Ronald and Drugan, Madalina},
  booktitle = {Proceedings of the 19th International Conference on Artificial Intelligence and Statistics},
  series = {Proceedings of Machine Learning Research},
  volume = {51},
  pages = {939--947},
  publisher = {PMLR},
  year = {2016}
}

@inproceedings{jamieson2015sparse,
  title = {Sparse dueling bandits},
  author = {Jamieson, Kevin and Katariya, Sumeet and Deshpande, Atul and Nowak, Robert},
  booktitle = {Proceedings of the 18th International Conference on Artificial Intelligence and Statistics},
  series = {Proceedings of Machine Learning Research},
  volume = {38},
  pages = {416--424},
  publisher = {PMLR},
  year = {2015}
}

@article{kaufmann2016complexity,
  title = {On the complexity of best-arm identification in multi-armed bandit models},
  author = {Kaufmann, Emilie and Capp{\'e}, Olivier and Garivier, Aur{\'e}lien},
  journal = {Journal of Machine Learning Research},
  volume = {17},
  number = {1},
  pages = {1--42},
  year = {2016}
}

@inproceedings{ameli2025statistical,
  title = {A statistical framework for ranking {LLM}-based chatbots},
  author = {Ameli, Siavash and Zhuang, Siyuan and Stoica, Ion and Mahoney, Michael W.},
  booktitle = {International Conference on Learning Representations},
  year = {2025}
}

@article{liang2023holistic,
  title = {Holistic evaluation of language models},
  author = {Liang, Percy and Bommasani, Rishi and Lee, Tony and Tsipras, Dimitris and Soylu, Dilara and Yasunaga, Michihiro and Zhang, Yian and Narayanan, Deepak and Wu, Yuhuai and Kumar, Ananya and Newman, Benjamin and Yuan, Binhang and Yan, Bobby and Zhang, Ce and Cosgrove, Christian and Manning, Christopher D. and R{\'e}, Christopher and Acosta-Navas, Diana and Hudson, Drew A. and Zelikman, Eric and Durmus, Esin and Ladhak, Faisal and Rong, Frieda and Ren, Hongyu and Yao, Huaxiu and Wang, Jue and Santhanam, Keshav and Orr, Laurel and Zheng, Lucia and Yuksekgonul, Mert and Suzgun, Mirac and Kim, Nathan and Guha, Neel and Chatterji, Niladri and Khattab, Omar and Henderson, Peter and Huang, Qian and Chi, Ryan and Xie, Sang Michael and Santurkar, Shibani and Ganguli, Surya and Hashimoto, Tatsunori and Icard, Thomas and Zhang, Tianyi and Chaudhary, Vishrav and Wang, William and Li, Xuechen and Mai, Yifan and Zhang, Yuhui and Koreeda, Yuta},
  journal = {Transactions on Machine Learning Research},
  year = {2023}
}

@inproceedings{frick2025prompt,
  title = {{Prompt-to-Leaderboard}: Prompt-adaptive {LLM} evaluations},
  author = {Frick, Evan and Chen, Connor and Tennyson, Joseph and Li, Tianle and Chiang, Wei-Lin and Angelopoulos, Anastasios Nikolas and Stoica, Ion},
  booktitle = {Proceedings of the 42nd International Conference on Machine Learning},
  series = {Proceedings of Machine Learning Research},
  volume = {267},
  pages = {17672--17689},
  publisher = {PMLR},
  year = {2025}
}

@article{heckel2019active,
  title = {Active ranking from pairwise comparisons and when parametric assumptions do not help},
  author = {Heckel, Reinhard and Shah, Nihar B. and Ramchandran, Kannan and Wainwright, Martin J.},
  journal = {The Annals of Statistics},
  volume = {47},
  number = {6},
  pages = {3099--3126},
  year = {2019},
  doi = {10.1214/18-AOS1772}
}

@article{goldberger2026optimal,
  title = {Optimal top-$k$ identification from pairwise comparisons},
  author = {Goldberger, Motti and Rudi, Nils},
  journal = {arXiv preprint arXiv:2607.08979},
  year = {2026}
}

@inproceedings{kone2023adaptive,
  title = {Adaptive algorithms for relaxed {Pareto} set identification},
  author = {Kone, Cyrille and Kaufmann, Emilie and Richert, Laura},
  booktitle = {Advances in Neural Information Processing Systems},
  volume = {36},
  pages = {35190--35201},
  publisher = {Curran Associates, Inc.},
  year = {2023},
  doi = {10.52202/075280-1528}
}

@inproceedings{liu2025elo,
  title = {{am-ELO}: A stable framework for arena-based {LLM} evaluation},
  author = {Liu, Zirui and Li, Jiatong and Zhuang, Yan and Liu, Qi and Shen, Shuanghong and Ouyang, Jie and Cheng, Mingyue and Wang, Shijin},
  booktitle = {Proceedings of the 42nd International Conference on Machine Learning},
  series = {Proceedings of Machine Learning Research},
  volume = {267},
  pages = {38857--38868},
  publisher = {PMLR},
  year = {2025}
}

@article{frauen2026nonparametric,
  title = {Nonparametric {LLM} evaluation from preference data},
  author = {Frauen, Dennis and Deviyani, Athiya and van der Schaar, Mihaela and Feuerriegel, Stefan},
  journal = {arXiv preprint arXiv:2601.21816},
  year = {2026}
}

@article{li2026low,
  title = {Low rank for rank: Uncertainty-aware task-specific {LLM} ranking under sparse pairwise comparisons},
  author = {Li, Jiachun and Simchi-Levi, David and Sun, Will Wei},
  journal = {arXiv preprint arXiv:2605.29395},
  year = {2026}
}

@inproceedings{liu2026fewer,
  title = {Fewer battles, more gain: An information-efficient framework for arena-based {LLM} evaluation},
  author = {Liu, Zirui and Wang, Xianquan and Zhuang, Yan and Li, Jiatong and Liu, Qi and Shen, Shuanghong and Cheng, Mingyue and Wang, Shijin},
  booktitle = {International Conference on Learning Representations},
  year = {2026}
}

@article{shah2018simple,
  title = {Simple, robust and optimal ranking from pairwise comparisons},
  author = {Shah, Nihar B. and Wainwright, Martin J.},
  journal = {Journal of Machine Learning Research},
  volume = {18},
  number = {199},
  pages = {1--38},
  year = {2018}
}

@inproceedings{karagozlu2024learning,
  title = {Learning the {Pareto} set under incomplete preferences: Pure exploration in vector bandits},
  author = {Karag{\"o}zl{\"u}, Efe Mert and Y{\i}ld{\i}r{\i}m, Ya{\c{s}}ar Cahit and Ararat, {\c{C}}a{\u{g}}{\i}n and Tekin, Cem},
  booktitle = {Proceedings of the 27th International Conference on Artificial Intelligence and Statistics},
  series = {Proceedings of Machine Learning Research},
  volume = {238},
  pages = {3070--3078},
  publisher = {PMLR},
  year = {2024}
}

@inproceedings{kone2025bandit,
  title = {Bandit {Pareto} set identification in a multi-output linear model},
  author = {Kone, Cyrille and Kaufmann, Emilie and Richert, Laura},
  booktitle = {Proceedings of the 28th International Conference on Artificial Intelligence and Statistics},
  series = {Proceedings of Machine Learning Research},
  volume = {258},
  pages = {1189--1197},
  publisher = {PMLR},
  year = {2025}
}

@inproceedings{shukla2024preference,
  title = {Preference-based pure exploration},
  author = {Shukla, Apurv and Basu, Debabrota},
  booktitle = {Advances in Neural Information Processing Systems},
  volume = {37},
  pages = {17313--17347},
  publisher = {Curran Associates, Inc.},
  year = {2024},
  doi = {10.52202/079017-0550}
}

@inproceedings{das2026frappe,
  title = {{FraPPE}: Fast and efficient preference-based pure exploration},
  author = {Das, Udvas and Shukla, Apurv and Basu, Debabrota},
  booktitle = {Advances in Neural Information Processing Systems},
  volume = {38},
  pages = {10190--10236},
  publisher = {Curran Associates, Inc.},
  year = {2025},
  doi = {10.52202/085713-0312}
}

@article{mannor2004sample,
  title = {The sample complexity of exploration in the multi-armed bandit problem},
  author = {Mannor, Shie and Tsitsiklis, John N.},
  journal = {Journal of Machine Learning Research},
  volume = {5},
  pages = {623--648},
  year = {2004}
}

@article{evendar2006action,
  title = {Action elimination and stopping conditions for the multi-armed bandit and reinforcement learning problems},
  author = {Even-Dar, Eyal and Mannor, Shie and Mansour, Yishay},
  journal = {Journal of Machine Learning Research},
  volume = {7},
  number = {39},
  pages = {1079--1105},
  year = {2006}
}

@inproceedings{audibert2010best,
  title = {Best arm identification in multi-armed bandits},
  author = {Audibert, Jean-Yves and Bubeck, S{\'e}bastien and Munos, R{\'e}mi},
  booktitle = {Proceedings of the 23rd Conference on Learning Theory},
  pages = {41--53},
  year = {2010}
}

@inproceedings{karnin2013almost,
  title = {Almost optimal exploration in multi-armed bandits},
  author = {Karnin, Zohar and Koren, Tomer and Somekh, Oren},
  booktitle = {Proceedings of the 30th International Conference on Machine Learning},
  series = {Proceedings of Machine Learning Research},
  volume = {28},
  pages = {1238--1246},
  publisher = {PMLR},
  year = {2013}
}

@inproceedings{jamieson2014lil,
  title = {{lil'UCB}: An optimal exploration algorithm for multi-armed bandits},
  author = {Jamieson, Kevin and Malloy, Matthew and Nowak, Robert and Bubeck, S{\'e}bastien},
  booktitle = {Proceedings of the 27th Conference on Learning Theory},
  series = {Proceedings of Machine Learning Research},
  volume = {35},
  pages = {423--439},
  publisher = {PMLR},
  year = {2014}
}

@inproceedings{garivier2016optimal,
  title = {Optimal best arm identification with fixed confidence},
  author = {Garivier, Aur{\'e}lien and Kaufmann, Emilie},
  booktitle = {Proceedings of the 29th Conference on Learning Theory},
  series = {Proceedings of Machine Learning Research},
  volume = {49},
  pages = {998--1027},
  publisher = {PMLR},
  year = {2016}
}

@article{russo2020simple,
  title = {Simple {Bayesian} algorithms for best-arm identification},
  author = {Russo, Daniel},
  journal = {Operations Research},
  volume = {68},
  number = {6},
  pages = {1625--1647},
  year = {2020},
  doi = {10.1287/opre.2019.1911}
}

@inproceedings{jourdan2022top,
  title = {Top two algorithms revisited},
  author = {Jourdan, Marc and Degenne, R{\'e}my and Baudry, Dorian and de Heide, Rianne and Kaufmann, Emilie},
  booktitle = {Advances in Neural Information Processing Systems},
  volume = {35},
  pages = {26791--26803},
  publisher = {Curran Associates, Inc.},
  year = {2022},
  doi = {10.52202/068431-1943}
}

@inproceedings{tirinzoni2022elimination,
  title = {On elimination strategies for bandit fixed-confidence identification},
  author = {Tirinzoni, Andrea and Degenne, R{\'e}my},
  booktitle = {Advances in Neural Information Processing Systems},
  volume = {35},
  pages = {18586--18598},
  publisher = {Curran Associates, Inc.},
  year = {2022},
  doi = {10.52202/068431-1350}
}

@inproceedings{bubeck2013multiple,
  title = {Multiple identifications in multi-armed bandits},
  author = {Bubeck, S{\'e}bastien and Wang, Tengyao and Viswanathan, Nitin},
  booktitle = {Proceedings of the 30th International Conference on Machine Learning},
  series = {Proceedings of Machine Learning Research},
  volume = {28},
  pages = {258--265},
  publisher = {PMLR},
  year = {2013}
}

@inproceedings{kaufmann2013information,
  title = {Information complexity in bandit subset selection},
  author = {Kaufmann, Emilie and Kalyanakrishnan, Shivaram},
  booktitle = {Proceedings of the 26th Conference on Learning Theory},
  series = {Proceedings of Machine Learning Research},
  volume = {30},
  pages = {228--251},
  publisher = {PMLR},
  year = {2013}
}

@inproceedings{chen2014combinatorial,
  title = {Combinatorial pure exploration of multi-armed bandits},
  author = {Chen, Shouyuan and Lin, Tian and King, Irwin and Lyu, Michael R. and Chen, Wei},
  booktitle = {Advances in Neural Information Processing Systems},
  volume = {27},
  pages = {379--387},
  publisher = {Curran Associates, Inc.},
  year = {2014}
}

@inproceedings{locatelli2016optimal,
  title = {An optimal algorithm for the thresholding bandit problem},
  author = {Locatelli, Andrea and Gutzeit, Maurilio and Carpentier, Alexandra},
  booktitle = {Proceedings of the 33rd International Conference on Machine Learning},
  series = {Proceedings of Machine Learning Research},
  volume = {48},
  pages = {1690--1698},
  publisher = {PMLR},
  year = {2016}
}

@inproceedings{katz2018feasible,
  title = {Feasible arm identification},
  author = {Katz-Samuels, Julian and Scott, Clay},
  booktitle = {Proceedings of the 35th International Conference on Machine Learning},
  series = {Proceedings of Machine Learning Research},
  volume = {80},
  pages = {2535--2543},
  publisher = {PMLR},
  year = {2018}
}

@inproceedings{degenne2019pure,
  title = {Pure exploration with multiple correct answers},
  author = {Degenne, R{\'e}my and Koolen, Wouter M.},
  booktitle = {Advances in Neural Information Processing Systems},
  volume = {32},
  pages = {14524--14533},
  publisher = {Curran Associates, Inc.},
  year = {2019}
}

@inproceedings{nakamura2024fixed,
  title = {Fixed-budget real-valued combinatorial pure exploration of multi-armed bandit},
  author = {Nakamura, Shintaro and Sugiyama, Masashi},
  booktitle = {Proceedings of the 27th International Conference on Artificial Intelligence and Statistics},
  series = {Proceedings of Machine Learning Research},
  volume = {238},
  pages = {1225--1233},
  publisher = {PMLR},
  year = {2024}
}

@inproceedings{carlsson2024pure,
  title = {Pure exploration in bandits with linear constraints},
  author = {Carlsson, Emil and Basu, Debabrota and Johansson, Fredrik D. and Dubhashi, Devdatt},
  booktitle = {Proceedings of the 27th International Conference on Artificial Intelligence and Statistics},
  series = {Proceedings of Machine Learning Research},
  volume = {238},
  pages = {334--342},
  publisher = {PMLR},
  year = {2024}
}

@inproceedings{pmlr-v258-cho25a,
  title = {Reward maximization for pure exploration: Minimax optimal good arm identification for nonparametric multi-armed bandits},
  author = {Cho, Brian M. and Meier, Dominik and Gan, Kyra and Kallus, Nathan},
  booktitle = {Proceedings of the 28th International Conference on Artificial Intelligence and Statistics},
  series = {Proceedings of Machine Learning Research},
  volume = {258},
  pages = {2251--2259},
  publisher = {PMLR},
  year = {2025}
}

@inproceedings{asudeh2015crowdsourcing,
  title = {Crowdsourcing {Pareto}-optimal object finding by pairwise comparisons},
  author = {Asudeh, Abolfazl and Zhang, Gensheng and Hassan, Naeemul and Li, Chengkai and Zaruba, Gergely V.},
  booktitle = {Proceedings of the 24th ACM International on Conference on Information and Knowledge Management},
  pages = {753--762},
  year = {2015}
}

@inproceedings{groz2020skyline,
  title = {Skyline computation with noisy comparisons},
  author = {Groz, Beno{\^\i}t and Mallmann-Trenn, Frederik and Mathieu, Claire and Verdugo, Victor},
  booktitle = {Combinatorial Algorithms},
  series = {Lecture Notes in Computer Science},
  volume = {12126},
  pages = {289--303},
  publisher = {Springer},
  year = {2020},
  doi = {10.1007/978-3-030-48966-3_22}
}

@inproceedings{audiffren2017bandits,
  title = {Bandits dueling on partially ordered sets},
  author = {Audiffren, Julien and Ralaivola, Liva},
  booktitle = {Advances in Neural Information Processing Systems},
  volume = {30},
  pages = {2130--2139},
  publisher = {Curran Associates, Inc.},
  year = {2017}
}

@article{davidson1970extending,
  title = {On extending the {Bradley-Terry} model to accommodate ties in paired comparison experiments},
  author = {Davidson, Roger R.},
  journal = {Journal of the American Statistical Association},
  volume = {65},
  number = {329},
  pages = {317--328},
  year = {1970},
  doi = {10.1080/01621459.1970.10481082}
}

@inproceedings{miroyan2026search,
  title = {{Search Arena}: Analyzing search-augmented {LLM}s},
  author = {Miroyan, Mihran and Wu, Tsung-Han and King, Logan and Li, Tianle and Pan, Jiayi and Hu, Xinyan and Chiang, Wei-Lin and Angelopoulos, Anastasios N. and Darrell, Trevor and Norouzi, Narges and Gonzalez, Joseph E.},
  booktitle = {International Conference on Learning Representations},
  year = {2026}
}

@inproceedings{chou2025visionarena,
  title = {{VisionArena}: 230k real world user-{VLM} conversations with preference labels},
  author = {Chou, Christopher and Dunlap, Lisa and Mashita, Koki and Mandal, Krishna and Darrell, Trevor and Stoica, Ion and Gonzalez, Joseph E. and Chiang, Wei-Lin},
  booktitle = {Proceedings of the IEEE/CVF Conference on Computer Vision and Pattern Recognition (CVPR)},
  pages = {3877--3887},
  year = {2025}
}

@article{xue2026costaware,
  title = {Cost-Aware Multi-Objective Bandits: Theory and Application to Budgeted {LLM} Configuration Evaluation},
  author = {Xue, Bo and Hong, Zhi and Li, Jiayi and Wan, Yuanyu and Cheng, Ji and Qiu, Shuang},
  journal = {arXiv preprint arXiv:2608.04333},
  year = {2026},
  url = {https://arxiv.org/abs/2608.04333}
}

\clearpage
\appendix
\startcontents[appendices]

\section*{Appendix Contents}

\begingroup
\setcounter{tocdepth}{2}
\printcontents[appendices]{l}{1}{}
\endgroup

\clearpage

\section{Additional related work}
\label{app:additional_related_work}

\paragraph{Best arm identification}
Best-arm identification is a canonical pure-exploration problem in
stochastic bandits, where samples are collected to identify an optimal
arm rather than to maximize cumulative reward. Early work established
PAC sample-complexity bounds and confidence-based elimination and
stopping procedures for identifying a near-optimal arm~\citep{mannor2004sample,evendar2006action}.
Subsequent work developed sharper procedures for best-arm identification
under fixed-budget and fixed-confidence formulations, including
Successive Rejects~\citep{audibert2010best}, Sequential Halving~\citep{karnin2013almost}, and lil'UCB~\citep{jamieson2014lil}.
\citet{kaufmann2016complexity} studied the instance-dependent complexity
of the two dominant fixed-budget and fixed-confidence formulations,
while \citet{garivier2016optimal} derived a tight fixed-confidence lower
bound and introduced the asymptotically optimal Track-and-Stop strategy. More recent work has developed alternative adaptive-allocation principles, notably Top-Two methods~\citep{russo2020simple,jourdan2022top}. Elimination strategies can further reduce the computational cost of fully adaptive fixed-confidence identification while preserving sample-complexity guarantees~\citep{tirinzoni2022elimination}.

\paragraph{Set-valued and structured identification.}
Beyond best-arm identification, bandit identification has been studied
for richer terminal targets. Top-\(m\) and subset-selection problems
seek several of the highest-mean arms~\citep{bubeck2013multiple,kaufmann2013information}, while thresholding
and good-arm identification aim to determine which arms satisfy a
prescribed performance criterion~\citep{locatelli2016optimal,pmlr-v258-cho25a}. Combinatorial pure exploration generalizes the answer space to structured decision classes, including fixed-size subsets, matchings, spanning trees, and paths \citep{chen2014combinatorial}, with recent work developing fixed-budget methods for more general real-valued combinatorial action classes~\citep{nakamura2024fixed}. Related extensions consider vector-valued observations and constrained decision spaces: \citet{katz2018feasible} study feasible-arm identification with multivariate means, while
\citet{carlsson2024pure} consider fixed-confidence identification when
the optimal policy is subject to linear constraints and may mix several
arms. More generally, \citet{degenne2019pure}
develop a fixed-confidence framework for pure-exploration problems that
may admit multiple correct answers.

\section{Technical proofs}
\label{app:proofs}
\subsection{Supporting results}
\subsubsection{Properties of the classification gaps}
\begin{lemma}[Basic inequalities for pairwise margins]
\label{lem:margin-calculus}
Let
\(
    x=(x_1,\ldots,x_K),
    \quad
    y=(y_1,\ldots,y_K)
    \in(\mathbb{R}^d)^K.
\)
Then, for all \(i,j,k\in[K]\),

\begin{enumerate}[label=(\roman*)]
    \item
    \(
        m_x(i,k)
        \geq
        m_x(i,j)+m_x(j,k);
    \)

    \item if
    \(
        \max_{u\in[K]}\max_{r\in[d]}
        |x_{u,r}-y_{u,r}|
        \leq \rho,
    \)
    then
    \[
        |m_x(i,j)-m_y(i,j)|
        \leq 2\rho,
        \qquad
        |M_x(i,j)-M_y(i,j)|
        \leq 2\rho;
    \]

    \item for all \(a,b\in\mathbb{R}\),
    \(
        |a^+-b^+|
        \leq
        |a-b|.
    \)
\end{enumerate}
\end{lemma}

\begin{proof}
For part~(i), coordinate-wise,
\[
    x_{k,r}-x_{i,r}
    =
    (x_{k,r}-x_{j,r})
    +
    (x_{j,r}-x_{i,r}),
\]
Taking the minimum over \(r\) and using
\(
    \min_r(a_r+b_r)
    \geq
    \min_r a_r+\min_r b_r
\)
gives
\[
m_x(i,k)\ge m_x(i,j)+m_x(j,k).
\]

For part~(ii), every coordinate difference satisfies
\[
    \left|
        (x_{j,r}-x_{i,r})
        -
        (y_{j,r}-y_{i,r})
    \right|
    \leq 2\rho.
\]
Both the minimum and maximum maps are \(1\)-Lipschitz with respect to the \(\ell_\infty\) norm, which proves the two inequalities.

Part~(iii) follows because the map \(a\mapsto a^+\) is \(1\)-Lipschitz.
\end{proof}

\begin{lemma}[Pareto witness for a non-Pareto model]
\label{lem:pareto-witness}
If \(i\notin\calP(x)\), then
\[
    \Delta_i^\star(x)
    =
    \max_{p\in\calP(x)}m_x(i,p)
    >
    0.
\]
In particular, there exists $i_x^\star \in \calP(x)$ such that
\(
    m_x(i,i_x^\star)
    =
    \Delta_i^\star(x).
\)
\end{lemma}

\begin{proof}
Since \(i\notin\calP(x)\), there exists a model that strictly dominates \(i\). Hence
\[
    \Delta_i^\star(x)
    =
    \max_{j\in[K]\setminus\{i\}}m_x(i,j)
    >
    0.
\]
Because the model set is finite, we may choose
\(
    i_x^\star
    \in
    \arg\max_{j\in[K]\setminus\{i\}}m_x(i,j).
\)
Then \( m_x(i,i_x^\star) = \Delta_i^\star(x) > 0, \) so \(i\prec_x i_x^\star\).

We claim that \(i_x^\star\in\calP(x)\). Suppose otherwise. Then there exists
\(p\in[K]\) such that \(i_x^\star\prec_x p\), and therefore
\(
    m_x(i_x^\star,p)>0.
\)
By \Cref{lem:margin-calculus},
\[
    m_x(i,p)
    \geq
    m_x(i,i_x^\star)+m_x(i_x^\star,p)
    >
    m_x(i,i_x^\star)
    =
    \Delta_i^\star(x).
\]
This contradicts the definition of \(i_x^\star\). Hence \(i_x^\star\in\calP(x)\), and because it already attains the maximum over all $j\neq i$, it also attains the maximum over $p\in \calP(x)$.
\end{proof}


For a Bradley--Terry instance \(\theta\in\Theta_{B,d}\), let
\(b=b(\theta)\) denote its coordinate-wise Borda vectors, and write
\[
    \Delta_i^{\mathrm B}:=\Delta_i(b(\theta)),
    \qquad
    \Delta_i^\theta:=\Delta_i(\theta).
\]
Recall that
\[
    \kappa_B
    =
    \inf_{|x|\le2B}\sigma'(x)
    =
    \sigma(2B)\bigl(1-\sigma(2B)\bigr)>0.
\]

\begin{lemma}[Borda gaps dominate latent gaps]
\label{lem:borda-latent-gap}
If $\theta\in\Theta_{B,d}$, then for every model $i$,
\[
    \Delta_i^{\Borda}
    \geq
    \kappa_B\Delta_i^\theta.
\]
\end{lemma}

\begin{proof}
Fix models $a,c$ and an objective $r$. For every \(u\in[K]\), define
\[
    \bar b_{u,r}
    :=
    \frac1K\sum_{\ell=1}^K
    \sigma(\theta_{u,r}-\theta_{\ell,r}).
\]
For each $\ell$, the mean-value theorem gives a point $\xi_{\ell}$ between $\theta_{a,r}-\theta_{\ell,r}$ and $\theta_{c,r}-\theta_{\ell,r}$ such that
\[
    \sigma(\theta_{c,r}-\theta_{\ell,r})
    -
    \sigma(\theta_{a,r}-\theta_{\ell,r})
    =
    \sigma'(\xi_{\ell})(\theta_{c,r}-\theta_{a,r}).
\]
Hence
\[
    \bar b_{c,r}-\bar b_{a,r}
    =
    \gamma_{acr}(\theta_{c,r}-\theta_{a,r}),
    \qquad
    \gamma_{acr}
    :=
    \frac1K\sum_{\ell=1}^K\sigma'(\xi_{\ell}).
\]
Because $\theta\in\Theta_{B,d}$, both endpoints defining each $\xi_{\ell}$ lie in $[-2B,2B]$, and therefore $\gamma_{acr}\geq\kappa_B$. Since
\[
    b_{u,r}
    =
    \frac{K\bar b_{u,r}-1/2}{K-1},
\]
we obtain
\[
    b_{c,r}-b_{a,r}
    =
    \frac{K}{K-1}
    \gamma_{acr}(\theta_{c,r}-\theta_{a,r}).
\]
In particular, because $K/(K-1)\geq1$, whenever $\theta_{c,r}\geq\theta_{a,r}$,
\begin{equation}
    b_{c,r}-b_{a,r}
    \geq
    \kappa_B(\theta_{c,r}-\theta_{a,r}).
    \label{eq:borda-latent-positive-difference}
\end{equation}

We then record three consequences. 

First, if $m_\theta(a,c)>0$, then every coordinate difference $\theta_{c,r}-\theta_{a,r}$ is positive, so \eqref{eq:borda-latent-positive-difference} gives
\begin{equation}
    m_b(a,c)
    \geq
    \kappa_Bm_\theta(a,c).
    \label{eq:borda-latent-domination-margin}
\end{equation}

Second, for all $a,c$,
\begin{equation}
    M_b(a,c)^+
    \geq
    \kappa_BM_\theta(a,c)^+.
    \label{eq:borda-latent-max-margin}
\end{equation}
If $M_\theta(a,c)\leq0$, this is trivial. Otherwise choose an objective attaining $M_\theta(a,c)=\theta_{a,r}-\theta_{c,r}>0$ and apply \eqref{eq:borda-latent-positive-difference} with the roles of $a$ and $c$ interchanged. 

Third, for every model $j$,
\begin{equation}
    (\Delta_j^\star(b))^+
    \geq
    \kappa_B(\Delta_j^\star(\theta))^+.
    \label{eq:borda-latent-raw-gap}
\end{equation}
If $\Delta_j^\star(\theta)\leq0$, this is trivial.  Otherwise choose a
model $c$ attaining $\Delta_j^\star(\theta)>0$ and apply
\eqref{eq:borda-latent-domination-margin} to $(j,c)$.

We now consider the two possible Pareto classes of model $i$.

If $i\notin\calP(\theta)$, coordinate-wise order preservation implies $i\notin\calP(b)$.  Choose a model $c$ attaining $\Delta_i^\star(\theta)>0$.  Then
\[
    \Delta_i^{\Borda}
    =
    \Delta_i^\star(b)
    \geq
    m_b(i,c)
    \geq
    \kappa_Bm_\theta(i,c)
    =
    \kappa_B\Delta_i^\theta.
\]

Suppose instead that $i\in\calP(\theta)$. Then $i\in\calP(b)$ by \Cref{lem:borda-order-preservation}.  Fix $j\neq i$.  Since $i$ is Pareto, $M_\theta(i,j)\geq0$.  If $M_\theta(i,j)>0$, applying \eqref{eq:borda-latent-positive-difference} at an objective attaining the positive maximum gives
\[
    M_b(i,j)
    \geq
    \kappa_BM_\theta(i,j).
\]
If $M_\theta(i,j)=0$, then some objective satisfies $\theta_{i,r}=\theta_{j,r}$; coordinate-wise order preservation gives $b_{i,r}=b_{j,r}$, and hence again $M_b(i,j)\geq0=\kappa_BM_\theta(i,j)$. Thus, in all cases,
\begin{equation}
    M_b(i,j)
    \geq
    \kappa_BM_\theta(i,j).
    \label{eq:borda-latent-pareto-first-component}
\end{equation}
Moreover, combining \eqref{eq:borda-latent-max-margin} and \eqref{eq:borda-latent-raw-gap} gives
\begin{equation}
    M_b(j,i)^++(\Delta_j^\star(b))^+
    \geq
    \kappa_B
    \left[
        M_\theta(j,i)^++(\Delta_j^\star(\theta))^+
    \right].
    \label{eq:borda-latent-pareto-second-component}
\end{equation}

Therefore, for every \(j\neq i\),
\begin{align*}
    M_b(i,j)
    \wedge
    \left[
        M_b(j,i)^+
        +
        \bigl(\Delta_j^\star(b)\bigr)^+
    \right] 
    \geq
    \kappa_B
    \left\{
        M_\theta(i,j)
        \wedge
        \left[
            M_\theta(j,i)^+
            +
            \bigl(\Delta_j^\star(\theta)\bigr)^+
        \right]
    \right\},
\end{align*}
Taking the minimum over \(j\neq i\) yields
\[
    \Delta_i^{\Borda}
    =
    \delta_i^\star(b)
    \geq
    \kappa_B\delta_i^\star(\theta)
    =
    \kappa_B\Delta_i^\theta.
\]
This proves the claim for every model $i$.
\end{proof}
\subsubsection{High-probability analysis of VB-EGE}
\label{app:borda-upper-bound}

We first state the upper-bound argument for a generic vector-Bernoulli mean instance. Fix 
\(
    b=(b_1,\ldots,b_K)\in([0,1]^d)^K.
\)
In this section, a query of model $i$ on objective $r$ reveals the
next unused observation from an independent stream
\[
    Z_{i,r,1},Z_{i,r,2},\ldots
    \overset{\mathrm{iid}}{\sim}
    \operatorname{Bernoulli}(b_{i,r}),
    \qquad i\in[K],\ r\in[d],
\]
These streams are independent of any additional internal randomization used by the algorithm.

Let $\mathbb P_b$ denote the induced law of the algorithm and the oracle. For compactness, write
\[
    \Delta_i^{\Borda}:=\Delta_i(b),
    \qquad
    \calP:=\calP(b),
    \qquad
    \Delta_{\min}^{\Borda}:=\min_{i\in[K]}\Delta_i^{\Borda}.
\]

Throughout this section, Algorithm~\ref{alg:vbege} is run with the common
configuration \(C=(2,4,4)\) specified in \Cref{eq:theory-constants}. Let $\tau$ count objective-specific binary comparisons, with $\tau=\infty$ if the procedure does not stop. Whenever \(\Delta_{\min}^{\Borda}>0\), define
\[
    M_i
    :=
    1+\left\lceil\log_2\frac{4}{\Delta_i^{\Borda}}\right\rceil,
    \qquad
    L_i(\delta)
    :=
    \log\frac{4KdM_i^2}{\delta}.
\]
Also define the $(1-\delta)$ stopping-time quantile under the
vector-Bernoulli instance $b$ by
\[
    Q_{b,\delta}(\VBEGE)
    :=
    \inf\left\{
        T\in\mathbb N_0:
        \mathbb P_b(\tau\leq T)\geq1-\delta
    \right\},
    \qquad
    \inf\varnothing:=\infty.
\]

\begin{proposition}[VB-EGE guarantee under vector-Bernoulli feedback]
\label{thm:borda-upper}
If $\Delta_{\min}^{\Borda}>0$, then Algorithm~\ref{alg:vbege}, run with
the configuration \(C\) in \eqref{eq:theory-constants}, satisfies
\begin{equation}
    \mathbb P_b\!\left(
        \widehat{\mathcal P}=\mathcal P(b),
        \quad
        \tau
        \leq
        dK
        +
        128d\sum_{i=1}^K
        \frac{L_i(\delta)}{(\Delta_i^{\Borda})^2}
    \right)
    \geq
    1-\delta.
    \label{eq:borda-upper-certificate}
\end{equation}
Consequently,
\[
    Q_{b,\delta}(\VBEGE)
    \leq
    dK
    +
    128d\sum_{i=1}^K
    \frac{L_i(\delta)}{(\Delta_i^{\Borda})^2}
    =
    \widetilde O\!\left(
        d\sum_{i=1}^K(\Delta_i^{\Borda})^{-2}
    \right).
\]
\end{proposition}

We first establish several lemmas used in the proof of \Cref{thm:borda-upper}. Under
\eqref{eq:theory-constants}, the phase schedule is
\[
    \varepsilon_m=2^{-m},
    \qquad
    L_m=\log\frac{4Kdm^2}{\delta},
    \qquad
    n_m
    =
    \left\lceil
        \frac{2L_m}{\varepsilon_m^2}
    \right\rceil,
    \qquad
    r_m=\sqrt{\frac{L_m}{2n_m}},
\]
and the removal threshold is $4r_m$.


For every phase \(m\), model \(i\), and objective \(r\), define the conceptual phase-\(m\) average
\[
    \widehat b_{i,r}^{(m)}
    :=
    \frac{1}{n_m}
    \sum_{s=1}^{n_m}Z_{i,r,s}.
\]
For an inactive model--objective pair, this quantity is used only in the analysis; the algorithm need not have revealed the corresponding prefix. At an inner-loop iteration with active set $A$, abbreviate
\[
\begin{aligned}
    \widehat m_A(i,j)
    &:=
    m_{\widehat b^{(m)}}(i,j),
    &
    \widehat M_A(i,j)
    &:=
    M_{\widehat b^{(m)}}(i,j),
    &
    \widehat\Delta_{i,A}^{\star}
    &:=
    \Delta_{i,A}^{\star}\!\left(\widehat b^{(m)}\right),
    &
    \widehat\delta_{i,A}^{\star}
    &:=
    \delta_{i,A}^{\star}\!\left(\widehat b^{(m)}\right).
\end{aligned}
\]
\begin{lemma}[Empirical gap identities]
\label{lem:empirical-sign-facts}
For every active set \(A\subseteq[K]\) with \(|A|\geq2\) and every \(i\in A\),
\[
    i\in\widehat{\mathcal S}_A
    \quad\Longleftrightarrow\quad
    \widehat\Delta_{i,A}^\star\leq0,
    \qquad
    i\notin\widehat{\mathcal S}_A
    \quad\Longleftrightarrow\quad
    \widehat\Delta_{i,A}^\star>0.
\]
and,
\[
    \widehat\Delta_{i,A}
    =
    \widehat\Delta_{i,A}^\star
    \vee
    \widehat\delta_{i,A}^\star.
\]
\end{lemma}

\begin{proof}
By definition,
\[
    \widehat\Delta_{i,A}^\star
    =
    \max_{j\in A\setminus\{i\}}
    \widehat m_A(i,j).
\]
Hence \(i\notin\widehat{\mathcal S}_A\) if and only if some \(j\in A\setminus\{i\}\) satisfies \(\widehat m_A(i,j)>0\), which is equivalent to \(\widehat\Delta_{i,A}^\star>0\). The first two equivalences follow.

Suppose first that \(i\in\widehat{\mathcal S}_A\). Then \(\widehat\Delta_{i,A}^\star\leq0\). Moreover, for every \(j\in A\setminus\{i\}\), empirical Pareto membership implies that \(j\) does not strictly dominate \(i\). Thus there exists a coordinate \(r\in[d]\) such that \(\widehat b_{i,r}\geq\widehat b_{j,r}\), and therefore
\[
    \widehat M_A(i,j)
    =
    \max_{r\in[d]}
    \bigl(\widehat b_{i,r}-\widehat b_{j,r}\bigr)
    \geq0.
\]
The other term,
\(
    \widehat M_A(j,i)^+
    +
    \bigl(\widehat\Delta_{j,A}^\star\bigr)^+,
\)
is also nonnegative. Hence every term in the minimum defining \(\widehat\delta_{i,A}^\star\) is nonnegative, so
\(
    \widehat\delta_{i,A}^\star\geq0
    \geq
    \widehat\Delta_{i,A}^\star.
\)
Therefore
\[
    \widehat\Delta_{i,A}^\star
    \vee
    \widehat\delta_{i,A}^\star
    =
    \widehat\delta_{i,A}^\star
    =
    \widehat\Delta_{i,A}.
\]

Now suppose that \(i\notin\widehat{\mathcal S}_A\). Then there exists \(j\in A\setminus\{i\}\) such that
\(
    \widehat m_A(i,j)>0,
\)
or equivalently,
\[
    \widehat M_A(i,j)
    =
    -\widehat m_A(i,j)
    <0.
\]
Using this \(j\) in the minimum defining \(\widehat\delta_{i,A}^\star\) gives
\begin{align*}
    \widehat\delta_{i,A}^\star
    &\leq
    \widehat M_A(i,j)
    \wedge
    \left[
        \widehat M_A(j,i)^+
        +
        \bigl(\widehat\Delta_{j,A}^\star\bigr)^+
    \right] \\
    &=
    \widehat M_A(i,j)
    <0,
\end{align*}
because the second term in the wedge is nonnegative. Since \(\widehat\Delta_{i,A}^\star>0\), it follows that
\[
    \widehat\Delta_{i,A}^\star
    \vee
    \widehat\delta_{i,A}^\star
    =
    \widehat\Delta_{i,A}^\star
    =
    \widehat\Delta_{i,A}.
\]
This proves the claim.
\end{proof}
\begin{lemma}[Uniform Borda concentration]
\label{lem:uniform-borda-confidence}
Define
\begin{equation}
    \calE
    :=
    \left\{
        \forall m\geq1,\ \forall i\in[K],\ \forall r\in[d]:
        \left|
            \widehat b_{i,r}^{(m)}-b_{i,r}
        \right|
        \leq r_m
    \right\}.
    \label{eq:uniform-borda-event}
\end{equation}
Then
\(
    \mathbb P_b(\calE)\geq1-\delta.
\)
Moreover, on \(\calE\), for every phase \(m\), active set \(A\), and distinct \(i,j\in A\),
\begin{equation}
    \left|
        \widehat m_A(i,j)-m_b(i,j)
    \right|
    \leq 2r_m,
    \qquad
    \left|
        \widehat M_A(i,j)-M_b(i,j)
    \right|
    \leq 2r_m.
    \label{eq:uniform-margin-deviation}
\end{equation}
\end{lemma}

\begin{proof}
For a fixed triple \((i,r,m)\), Hoeffding's inequality and the definition of \(r_m\) give
\[
    \mathbb P_b\!\left(
        \left|
            \widehat b_{i,r}^{(m)}-b_{i,r}
        \right|
        >r_m
    \right)
    \leq
    2\exp(-2n_mr_m^2)
    =
    2e^{-L_m}
    =
    \frac{\delta}{2Kdm^2}.
\]
Taking a union bound over \(i\in[K]\), \(r\in[d]\), and \(m\geq1\) yields
\[
    \mathbb P_b(\calE^c)
    \leq
    \frac{\delta}{2}
    \sum_{m=1}^{\infty}\frac{1}{m^2}
    <
    \delta.
\]
The two margin bounds follow from \Cref{lem:margin-calculus}.
\end{proof}

For every non-Pareto model \(i\notin\calP\), fix once and for all a Borda-Pareto witness supplied by \Cref{lem:pareto-witness}:
\[
    i^\star
    \in
    \arg\max_{p\in\calP}m_b(i,p),
    \qquad
    m_b(i,i^\star)=\Delta_i^{\Borda}>0.
\]
For an active set \(A\), define the witness invariant
\begin{equation}
    \mathsf W(A)
    :=
    \left\{
        \forall i\notin\calP,
        \quad
        i\in A\Longrightarrow i^\star\in A
    \right\}.
    \label{eq:witness-invariant}
\end{equation}
Initially, \(\mathsf W([K])\) holds.

\begin{lemma}[Every removed model is correctly classified]
\label{lem:safe-classification}
Assume that \(\calE\) and \(\mathsf W(A)\) hold when a model \(i\in A\) is removed in phase \(m\). Then:
\begin{enumerate}[label=(\roman*)]
    \item if \(i\notin\widehat{\mathcal S}_A\), then \(i\notin\calP\);
    \item if \(i\in\widehat{\mathcal S}_A\), then \(i\in\calP\).
\end{enumerate}
Thus every model removed without acceptance is truly non-Pareto, while every accepted model is truly Pareto.
\end{lemma}

\begin{proof}
For simplicity, we write \(r=r_m\).

First suppose \(i\notin\widehat{\mathcal S}_A\). Since \(i\) is removed, \(\widehat\Delta_{i,A}>4r\). By \Cref{lem:empirical-sign-facts}, \(\widehat\Delta_{i,A}=\widehat\Delta_{i,A}^\star\), so there exists \(j\in A\setminus\{i\}\) satisfies
\(
    \widehat m_A(i,j)>4r.
\)
Using \eqref{eq:uniform-margin-deviation},
\[
    m_b(i,j) \geq \widehat m_A(i,j) - 2r >2r>0.
\]
Hence \(i\prec_b j\), and therefore \(i\notin\calP(b)=\calP\).

Now suppose \(i\in\widehat{\mathcal S}_A\) but, toward a contradiction, \(i\notin\calP\). The witness invariant gives \(i^\star\in A\), and
\[
    m_b(i,i^\star)=\Delta_i^{\Borda}>0,
    \qquad
    M_b(i,i^\star)=-\Delta_i^{\Borda}.
\]
If \(\Delta_i^{\Borda}>2r\), then \eqref{eq:uniform-margin-deviation} implies
\[
    \widehat m_A(i,i^\star)
    \geq
    \Delta_i^{\Borda}-2r
    >
    0,
\]
contradicting \(i\in\widehat{\mathcal S}_A\). Thus necessarily \(\Delta_i^{\Borda}\leq2r\).

Since \(i\in\widehat{\mathcal S}_A\), its empirical classification gap is \(\widehat\Delta_{i,A}=\widehat\delta_{i,A}^\star\). Taking \(j=i^\star\) in the defining minimum gives
\begin{align*}
    \widehat\Delta_{i,A}
    &\leq
    \widehat M_A(i,i^\star)
    \wedge
    \left[
        \widehat M_A(i^\star,i)^+
        +
        \bigl(\widehat\Delta_{i^\star,A}^\star\bigr)^+
    \right] \\
    &\leq
    \widehat M_A(i,i^\star) \\
    &\leq
    M_b(i,i^\star)+2r \\
    &=
    -\Delta_i^{\Borda}+2r
    \leq
    2r,
\end{align*}
contradicting the removal condition \(\widehat\Delta_{i,A}>4r\). Therefore \(i\in\calP\).
\end{proof}

\begin{lemma}[Witness-invariant preservation]
\label{lem:witness-invariant-preservation}
Assume that $\calE$ and $\mathsf W(A)$ hold before a removal in phase $m$.
After the removal, the new active set $A'$ also satisfies
$\mathsf W(A')$.
\end{lemma}

\begin{proof}
Write $r=r_m$.  By \Cref{lem:safe-classification}, the only possible way to violate the invariant is to remove a true Pareto model $p\in\calP$ while leaving active a non-Pareto model $i\notin\calP$ whose fixed witness is $i^\star=p$.  Suppose, toward a contradiction, that this occurs. Immediately before the removal, $p,i\in A$.  Since the removed model $p$ is truly Pareto, \Cref{lem:safe-classification} implies that it cannot be empirically non-Pareto.  Thus $p\in\widehat{\mathcal S}_A$, and
\begin{equation}
    \widehat\Delta_{p,A}
    =
    \widehat\delta_{p,A}^\star
    >4r.
    \label{eq:removed-witness-gap}
\end{equation}

There are two cases to be considered.

\emph{Case 1: $p$ empirically strictly dominates $i$.}
Then $\widehat m_A(i,p)>0$, so
$i\notin\widehat{\mathcal S}_A$ and
$\widehat\Delta_{i,A}=\widehat\Delta_{i,A}^\star>0$.  Moreover,
$\widehat M_A(i,p)^+=0$.  Taking $j=i$ in the minimum defining
$\widehat\delta_{p,A}^\star$ yields
\[
    \widehat\Delta_{p,A}
    \leq
    \widehat M_A(i,p)^+
    +(\widehat\Delta_{i,A}^\star)^+
    =
    \widehat\Delta_{i,A}.
\]
Since \(p\) is the model selected for removal, it belongs to
\(
    G
    :=
    \arg\max_{u\in A}\widehat\Delta_{u,A}.
\)
If the preceding inequality is strict, then
\(\widehat\Delta_{i,A}>\widehat\Delta_{p,A}\), contradicting
\(p\in G\). If equality holds, then
\(i\in G\setminus\widehat{\mathcal S}_A\), so the
non-Pareto-first tie-breaking rule \Cref{rem:vbege-conventions} selects an empirical non-Pareto
maximizer rather than \(p\). In either case, the algorithm cannot remove
\(p\), which results in a contradiction.

\emph{Case 2: \(p\) does not empirically strictly dominate \(i\).} Then
\(
    \widehat m_A(i,p)\leq0,
\)
and hence
\[
    \widehat M_A(i,p)
    =
    -\widehat m_A(i,p)
    \geq0.
\]
Since
\(
    M_b(i,p)
    =
    -m_b(i,p)
    =
    -\Delta_i^{\Borda},
\)
the uniform margin deviation \eqref{eq:uniform-margin-deviation} gives
\begin{equation}
    0
    \leq
    \widehat M_A(i,p)
    =
    \widehat M_A(i,p)^+
    \leq
    2r-\Delta_i^{\Borda}.
    \label{eq:small-gap-left-behind}
\end{equation}
So
\(
    \Delta_i^{\Borda}\leq2r.
\)
Moreover, for every \(q\in A\setminus\{i\}\),
\[
    \widehat m_A(i,q)
    \leq
    m_b(i,q)+2r
    \leq
    \Delta_i^{\Borda}+2r,
\]
Therefore,
\begin{equation}
    \bigl(\widehat\Delta_{i,A}^\star\bigr)^+
    \leq
    \Delta_i^{\Borda}+2r.
    \label{eq:left-behind-raw-gap}
\end{equation}

Since \(p\in\widehat{\mathcal S}_A\), \(\widehat\Delta_{p,A}
=\widehat\delta_{p,A}^\star\). Taking \(j=i\) in the defining minimum gives
\begin{align*}
    \widehat\Delta_{p,A}
    &\leq
    \widehat M_A(p,i)
    \wedge
    \left[
        \widehat M_A(i,p)^+
        +
        \bigl(\widehat\Delta_{i,A}^\star\bigr)^+
    \right] \\
    &\leq
    \widehat M_A(i,p)^+
    +
    \bigl(\widehat\Delta_{i,A}^\star\bigr)^+ \\
    &\leq
    \bigl(2r-\Delta_i^{\Borda}\bigr)
    +
    \bigl(\Delta_i^{\Borda}+2r\bigr) \\
    &=4r,
\end{align*}
contradicting \eqref{eq:removed-witness-gap}. Hence the invariant is preserved.
\end{proof}

\begin{proposition}[Correctness on the confidence event]
\label{prop:correctness-confidence}
On $\calE$, Algorithm~\ref{alg:vbege} never misclassifies a removed model. If it stops, its recommendation equals $\calP(b)$.
\end{proposition}

\begin{proof}
The invariant holds initially and is preserved after every removal by \Cref{lem:witness-invariant-preservation}.  By induction over removals and \Cref{lem:safe-classification}, every model added to \(B\) belongs to \(\mathcal P\), while every model removed without being added to \(B\) lies outside \(\mathcal P\).

If the algorithm stops with a singleton active set $A=\{a\}$ and $a\notin\calP$, then the invariant would require the distinct witness $a^\star$ to remain active, contradicting $|A|=1$.  Thus the final active model is Pareto. Every model outside \(B\cup A\) was previously removed without acceptance and hence lies outside \(\mathcal P\). Conversely, every model in \(B\cup A\) is Pareto. Therefore the returned set \(B\cup A\) equals \(\mathcal P(b)\).
\end{proof}

\begin{lemma}[Removal once the radius is small]
\label{lem:small-radius-removal}
Assume $\calE$.  Consider a phase $m$ and a model $i$ that is active at the beginning of the inner loop.  If
\[
    r_m < \frac{\Delta_i^{\Borda}}{8},
\]
then either the algorithm stops during phase $m$, or model $i$ is removed during that phase.
\end{lemma}

\begin{proof}
Write \(r=r_m\), and suppose that \(i\) remains active at some inner-loop iteration with \(|A|\geq2\). By \Cref{lem:witness-invariant-preservation}, the witness invariant \(\mathsf W(A)\) holds.

If $i\notin\calP$, then its fixed witness $i^\star$ remains active. By \eqref{eq:uniform-margin-deviation},
\[
    \widehat m_A(i,i^\star)
    \geq
    \Delta_i^{\Borda}-2r
    >0.
\]
Thus $i\notin\widehat{\mathcal S}_A$, and \Cref{lem:empirical-sign-facts} gives
\[
    \widehat\Delta_{i,A}
    =
    \widehat\Delta_{i,A}^\star
    \geq
    \Delta_i^{\Borda}-2r
    >
    6r
    >
    4r.
\]

Now suppose $i\in\calP$. Then $\Delta_i^{\Borda}=\delta_i^\star(b)$.  Hence, for every $j\in A\setminus\{i\}$, both components in the defining wedge are at least $\Delta_i^{\Borda}$:
\begin{equation}
    M_b(i,j)\geq\Delta_i^{\Borda},
    \qquad
    M_b(j,i)^++(\Delta_j^\star(b))^+
    \geq
    \Delta_i^{\Borda}.
    \label{eq:true-pareto-gap-components}
\end{equation}
By \eqref{eq:uniform-margin-deviation},
\[
    \widehat M_A(i,j)
    \geq
    \Delta_i^{\Borda}-2r
    >0.
\]
Thus no active model empirically strictly dominates $i$, so $i\in\widehat{\mathcal S}_A$. 
For any active model \(j\in A\),
\[
\bigl(\widehat{\Delta}^{\star}_{j,A}\bigr)^+
\ge
\bigl(\Delta_j^\star(b)\bigr)^+ - 2r.
\]
Indeed, if \(j\in\mathcal P\), then
\(\Delta_j^\star(b)\le 0\), so the claim is immediate since the
left-hand side is nonnegative. If \(j\notin\mathcal P\), the
witness invariant ensures that its fixed Pareto witness
\(j^\star\) remains in \(A\). Hence
\[
\widehat{\Delta}^{\star}_{j,A}
=
\max_{q\in A\setminus\{j\}}\widehat m_A(j,q)
\ge
\widehat m_A(j,j^\star)
\ge
m_b(j,j^\star)-2r
=
\Delta_j^\star(b)-2r.
\]
Since \(\Delta_j^\star(b)>0\) in this case,
\[
\bigl(\widehat{\Delta}^{\star}_{j,A}\bigr)^+
\ge
\bigl(\Delta_j^\star(b)\bigr)^+ - 2r.
\]
Combining with \eqref{eq:true-pareto-gap-components} yields
\[
    \widehat M_A(j,i)^+
    +(\widehat\Delta_{j,A}^\star)^+
    \geq
    \Delta_i^{\Borda}-4r > 4r.
\]
Together with $\widehat M_A(i,j)\geq\Delta_i^{\Borda}-2r$, this shows that every bracket in the definition of $\widehat\delta_{i,A}^\star$ is at least $\Delta_i^{\Borda}-4r$. Therefore
\[
    \widehat\Delta_{i,A}
    =
    \widehat\delta_{i,A}^\star
    \geq
    \Delta_i^{\Borda}-4r > 4r.
\]

In either case, as long as $i$ remains active and $|A|\geq2$, its empirical gap exceeds the removal threshold. Hence the inner loop cannot terminate while retaining $i$: the algorithm must remove $i$, unless it stops first because only one active model remains.
\end{proof}

\begin{proof}[Proof of \Cref{thm:borda-upper}]
By \Cref{lem:uniform-borda-confidence}, $\mathbb P_b(\calE)\geq1-\delta$.  Work on $\calE$.  Correctness upon stopping follows from \Cref{prop:correctness-confidence}; it remains to bound the query count.

For each model $i$, define
\[
    m_i
    :=
    \min\left\{
        m\geq1:
        \varepsilon_m<\frac{\Delta_i^{\Borda}}{4}
    \right\}.
\]
This integer is well defined because $\varepsilon_m\downarrow0$. Moreover, $0<\Delta_i^{\Borda}\leq1$: for a non-Pareto model the raw domination gap is at most $1$, while for a Pareto model the defining wedge is bounded above by $M_b(i,j)\leq1$. Hence $m_i>1$. By minimality of $m_i$,
\[
    \varepsilon_{m_i-1}
    \geq
    \frac{\Delta_i^{\Borda}}{4},
    \qquad
    \varepsilon_{m_i}
    =
    \frac12\varepsilon_{m_i-1}
    \geq
    \frac{\Delta_i^{\Borda}}{8}.
\]
On the other hand, the defining property of $m_i$ gives
\[
    \varepsilon_{m_i}
    <
    \frac{\Delta_i^{\Borda}}{4}.
\]
Since $n_m\geq2L_m/\varepsilon_m^2$,
\[
    r_m
    =
    \sqrt{\frac{L_m}{2n_m}}
    \leq
    \frac{\varepsilon_m}{2}.
\]
Therefore
\[
    r_{m_i}
    <
    \frac{\Delta_i^{\Borda}}{8}.
\]
By \Cref{lem:small-radius-removal}, if the algorithm reaches phase \(m_i\) and model \(i\) is still active, then during that phase either the algorithm stops or \(i\) is removed. Thus no coordinate of model \(i\) is sampled after phase \(m_i\).

In particular, the algorithm stops after finitely many phases. Indeed, let
\[
    \overline m:=\max_{i\in[K]}m_i<\infty.
\]
If the algorithm had not stopped by the end of phase \(\overline m\), any model still active at that time would have remained active beyond its phase \(m_i\), contradicting \Cref{lem:small-radius-removal}.

Consequently, the algorithm stops by phase \(\overline m\), and model \(i\) is never sampled after phase \(m_i\). Each of its \(d\) coordinates therefore receives at most \(n_{m_i}\) observations, so
\[
    \tau\le d\sum_{i=1}^K n_{m_i}.
\]
It remains to bound \(n_{m_i}\). By its definition and the lower bound \(\varepsilon_{m_i}\geq\Delta_i^{\Borda}/8\),
\begin{equation*}
    n_{m_i}
    \leq
    1+
    \frac{2}{\varepsilon_{m_i}^2}
    \log\frac{4Kdm_i^2}{\delta} 
    \leq
    1+
    \frac{128}{(\Delta_i^{\Borda})^2}
    \log\frac{4Kdm_i^2}{\delta}.
\end{equation*}
Furthermore, \(m_i\leq M_i\), because
\[
    \varepsilon_{M_i}
    =
    2^{-M_i}
    \leq
    \frac{\Delta_i^{\Borda}}{8}
    <
    \frac{\Delta_i^{\Borda}}{4}.
\]
Consequently,
\[
    n_{m_i}
    \leq
    1+
    \frac{128L_i(\delta)}
         {(\Delta_i^{\Borda})^2}.
\]

Combining the preceding two displays gives
\[
    \tau
    \le dK+128d\sum_{i=1}^K
    \frac{L_i(\delta)}{(\Delta_i^{\Borda})^2}.
\]
Thus, on \(\calE\), the algorithm returns \(\calP(b)\) within the stated
query budget. Since \(\mathbb P_b(\calE)\ge1-\delta\),
\eqref{eq:borda-upper-certificate} follows. The quantile statement follows directly from the definition of \(Q_{b,\delta}(\VBEGE)\).
\end{proof}
\subsubsection{Change-of-measure tools}
\begin{lemma}[Monotonicity of binary relative entropy]
\label{lem:binary-kl-monotonicity}
Let
\[
    0\leq q\leq b<a\leq p\leq1.
\]
Then
\[
    \kl(p,q)
    \geq
    \kl(a,b),
\]
where binary relative entropy is extended to the boundary by the usual conventions.
\end{lemma}

\begin{proof}
For \(p,q\in(0,1)\),
\[
    \frac{\partial}{\partial p}\kl(p,q)
    =
    \log\frac{p(1-q)}{q(1-p)},
\]
which is positive whenever \(p>q\). Moreover,
\[
    \frac{\partial}{\partial q}\kl(p,q)
    =
    \frac{q-p}{q(1-q)},
\]
which is negative whenever \(p>q\). Thus \(\kl(p,q)\) is increasing in its first argument and decreasing in its second argument on the region \(p>q\). The boundary cases follow by taking the corresponding one-sided limits; cases in which the left-hand side is infinite are immediate.
\end{proof}
\begin{lemma}[Logistic KL upper bound]
\label{lem:logistic-kl-upper}
For all \(x,y\in\R\),
\[
    \kl(\sigma(x),\sigma(y))
    \leq
    \frac{(x-y)^2}{8}.
\]
\end{lemma}

\begin{proof}
The Bernoulli distribution with mean \(\sigma(u)\) is the one-parameter
exponential family with log-partition function \(A(u)=\log(1+e^u)\). Since
\(A''(u)=\sigma(u)(1-\sigma(u))\leq1/4\), the function \(A\) is globally
\(1/4\)-smooth. The corresponding Bregman-divergence representation gives
\[
    \kl(\sigma(x),\sigma(y))
    =
    A(y)-A(x)-A'(x)(y-x)
    \leq
    \frac{(y-x)^2}{8}.
\]
\end{proof}
For \(1\leq i<j\leq K\), objective \(r\in[d]\), and \(u\in\mathbb N_0\) with \(u\le\tau\), define
\[
    N_{ij,r}^{\mathrm{pair}}(u)
    :=
    \sum_{t=1}^{u}
    \1\!\left\{
        R_t=r,\,
        \{I_t,J_t\}=\{i,j\}
    \right\}.
\]
Thus \(N_{ij,r}^{\mathrm{pair}}(u)\) counts comparisons of the unordered pair \(\{i,j\}\) on objective \(r\) among the first \(u\) queries.

For every model \(i\in[K]\), define its incident-comparison count by
\[
    N_{i,r}^{\mathrm{inc}}(u)
    :=
    \sum_{j<i}N_{ji,r}^{\mathrm{pair}}(u)
    +
    \sum_{j>i}N_{ij,r}^{\mathrm{pair}}(u).
\]

\begin{lemma}[Bounded-horizon transportation]
\label{lem:bounded-horizon-transport}
Fix \(T\in\mathbb N_0\). Let \(\theta\) and \(\lambda\) be two instances, and let \(G\) be an event measurable with respect to the transcript stopped at \(\tau\wedge T\), including the terminal recommendation if the algorithm stops by time \(T\). Then
\begin{equation}
    \sum_{r=1}^d
    \sum_{1\leq i<j\leq K}
    \E_\theta\!\left[
        N_{ij,r}^{\mathrm{pair}}(\tau\wedge T)
    \right]
    \kl\!\left(
        p_{ij}^{(r)}(\theta),
        p_{ij}^{(r)}(\lambda)
    \right)
    \geq
    \kl\!\left(
        \P_\theta(G),
        \P_\lambda(G)
    \right).
    \label{eq:bounded-horizon-transport}
\end{equation}
\end{lemma}

\begin{proof}
Let \(\P_\theta^T\) and \(\P_\lambda^T\) denote the laws of the transcript
stopped at \(\tau\wedge T\). View the stopped transcript as a length-\(T\)
adaptive experiment with a common cemetery action and observation after
stopping, which contributes zero KL divergence. At each time, the query is
predictable from the past and the algorithmic randomization, whose law is the
same under both instances. The chain rule for KL divergence therefore gives
\[
    \KL(\P_\theta^T,\P_\lambda^T)
    =
    \E_\theta\left[
        \sum_{t=1}^{\tau\wedge T}
        \kl\left(
            p_{I_tJ_t}^{(R_t)}(\theta),
            p_{I_tJ_t}^{(R_t)}(\lambda)
        \right)
    \right].
\]
For a reverse-orientation query,
\(p_{ji}^{(r)}=1-p_{ij}^{(r)}\), and
\(\kl(1-p,1-q)=\kl(p,q)\). Grouping both orientations by unordered pair
turns the right-hand side into the left-hand side of
\eqref{eq:bounded-horizon-transport}. Finally, applying data processing to
the binary statistic \(\1_G\) gives
\[
    \KL\!\left(
        \P_\theta^T
        \,\middle\|\,
        \P_\lambda^T
    \right)
    \geq
    \kl\left(
        \P_\theta(G),
        \P_\lambda(G)
    \right),
\]
which proves the claim.
\end{proof}

\subsection{Proofs of main results}
\subsubsection{Proof of \Cref{lem:borda-order-preservation}}
\label{app:borda-order-preservation-proof}

\begin{proof}[Proof of \Cref{lem:borda-order-preservation}]
For every model \(u\in[K]\) and objective \(r\in[d]\), define the self-included Borda coordinate
\[
    \bar b_{u,r}
    :=
    \frac{1}{K}
    \sum_{\ell=1}^{K}
    \sigma(\theta_{u,r}-\theta_{\ell,r}).
\]
Since \(\sigma(0)=1/2\),
\[
    b_{u,r}
    =
    \frac{K\bar b_{u,r}-1/2}{K-1}.
\]
In particular, \(b_{u,r}\) is a strictly increasing affine function of \(\bar b_{u,r}\).

Fix \(i,j\in[K]\) and \(r\in[d]\). Suppose first that \(\theta_{i,r}>\theta_{j,r}\). Then, for every \(\ell\in[K]\),
\[
    \theta_{i,r}-\theta_{\ell,r}
    >
    \theta_{j,r}-\theta_{\ell,r}.
\]
Because the logistic function is strictly increasing,
\[
    \sigma(\theta_{i,r}-\theta_{\ell,r})
    >
    \sigma(\theta_{j,r}-\theta_{\ell,r})
    \qquad
    \text{for every }\ell\in[K].
\]
Averaging over \(\ell\) yields \(\bar b_{i,r}>\bar b_{j,r}\), and hence \(b_{i,r}>b_{j,r}\).

Exchanging \(i\) and \(j\) proves that \(\theta_{i,r}<\theta_{j,r}\) implies \(b_{i,r}<b_{j,r}\). If \(\theta_{i,r}=\theta_{j,r}\), then the summands defining \(\bar b_{i,r}\) and \(\bar b_{j,r}\) coincide for every \(\ell\), so \(b_{i,r}=b_{j,r}\). Therefore,
\[
    \operatorname{sign}(b_{i,r}-b_{j,r})
    =
    \operatorname{sign}(\theta_{i,r}-\theta_{j,r}),
\]
which proves \eqref{eq:borda-order-preservation}.

Applying the sign identity coordinate-wise gives
\[
    \theta_{i,r}<\theta_{j,r}
    \ \text{for all }r\in[d]
    \quad\Longleftrightarrow\quad
    b_{i,r}<b_{j,r}
    \ \text{for all }r\in[d].
\]
Hence \(i\prec_\theta j\) if and only if \(i\prec_b j\). Since the strict Pareto frontier is determined entirely by this domination relation,
\[
    \mathcal P(b)
    =
    \mathcal P(\theta)
    =
    \mathcal P^\star,
\]
which proves \eqref{eq:borda-pareto-preservation}.
\end{proof}

\subsubsection{Proof of \Cref{thm:upper}}\label{app:upper-bound-proof}
\begin{proof}[Proof of \Cref{thm:upper}]
Conditional on $\mathcal F_{t-1}$ and $(I_t,R_t)=(i,r)$,
the opponent $J_t$ is uniform on $[K]\setminus\{i\}$.
By \eqref{eq:bt-model},
\[
\mathbb P_\theta(Y_t=1\mid\mathcal F_{t-1},I_t=i,R_t=r)
=
\frac{1}{K-1}\sum_{j\ne i}p_{ij}^{(r)}(\theta)
=
b_{i,r}(\theta).
\]
Since the opponent and comparison randomness are fresh at every query, the oracle can be realized by mutually independent pre-generated streams
\[
    Z_{i,r,1},Z_{i,r,2},\ldots
    \overset{\mathrm{iid}}{\sim}
    \operatorname{Bernoulli}\!\left(b_{i,r}(\theta)\right),
\]
revealing the next unused observation whenever the algorithm queries \((i,r)\). Algorithm~\ref{alg:vbege} discards the opponent identities and uses only these binary outcomes, so its stopping time and recommendation have the same law under this stream representation. Hence \Cref{thm:borda-upper} applies with \(b=b(\theta)\) under \(\mathbb P_\theta\). By \Cref{lem:borda-order-preservation},
\[
    \calP(b(\theta))=\calP(\theta).
\]

Let $M_i$ and $L_i(\delta)$ be the Borda-space quantities from Appendix~\ref{app:borda-upper-bound}.  By \Cref{lem:borda-latent-gap}, $\Delta_i^{\Borda}\geq\kappa_B\Delta_i^\theta$, and therefore
\[
    M_i
    =
    1+\left\lceil\log_2\frac{4}{\Delta_i^{\Borda}}\right\rceil
    \leq
    1+\left\lceil
        \log_2\frac{4}{\kappa_B\Delta_i^\theta}
    \right\rceil
    =
    M_i^\theta.
\]
It follows that $L_i(\delta)\leq L_i^\theta(\delta)$ and
\[
    \frac{L_i(\delta)}{(\Delta_i^{\Borda})^2}
    \leq
    \frac{L_i^\theta(\delta)}
    {\kappa_B^2(\Delta_i^\theta)^2}.
\]
Substituting this inequality into \eqref{eq:borda-upper-certificate}, with $b=b(\theta)$, gives
\[
    \mathbb P_\theta\!\left(
        \widehat{\mathcal P}=\mathcal P(\theta),
        \quad
        \tau
        \leq
        dK
        +
        \frac{128d}{\kappa_B^2}
        \sum_{i=1}^K
        \frac{L_i^\theta(\delta)}{(\Delta_i^\theta)^2}
    \right)
    \geq
    1-\delta.
\]
The displayed event gives the stated bound on $Q_{\theta,\delta}(\VBEGE)$.  Absorbing the logarithmic factors $L_i^\theta(\delta)$ yields the $\widetilde O$ expression. If $B$ is bounded by an absolute constant, i.e., $B=O(1)$, then $\kappa_B$ is bounded below by a positive absolute constant, i.e., $\kappa_B=\Theta(1)$.
\end{proof}

\subsubsection{Proof of \Cref{thm:minimax-optimality}}
\label{app:lower-bound-proof}
\label{app:minimax-proof}

We first establish a quantile lower bound for a symmetric instance and then apply it to the gap-separated class.

Fix \(\Delta>0\) and consider the symmetric instance \(\theta^\Delta\) defined by
\[
    \theta^\Delta_{1,r}=0,
    \qquad
    \theta^\Delta_{i,r}=-\Delta,
    \qquad
    i\in\{2,\ldots,K\},\ r\in[d].
\]
The first model strictly dominates every other model, and hence
\(\mathcal P(\theta^\Delta)=\{1\}\).

To obtain a lower bound, we construct alternatives under which one
of the dominated models becomes Pareto. 
For \(i\in\{2,\ldots,K\}\), \(r\in[d]\), and \(\eta>0\), define the one-coordinate alternative
\[
    \lambda^{i,r,\eta}_{a,s}
    :=
    \begin{cases}
        \eta,
        &(a,s)=(i,r),\\
        \theta^\Delta_{a,s},
        &\text{otherwise}.
    \end{cases}
\]

The alternative changes only the $i$th model's score on objective \(r\),
moving it from \(-\Delta\) to \(\eta>0\). Consequently, model \(i\)
is Pareto under \(\lambda^{i,r,\eta}\), so that
$\mathcal P(\lambda^{i,r,\eta})
\neq
\mathcal P(\theta^\Delta)$.

The parameter \(\eta\) measures how far the perturbed coordinate lies
beyond the Pareto decision boundary at \(0\). Smaller \(\eta\) yields a harder alternative, since the Pareto frontier still changes while the largest
affected Bradley--Terry logit shift, \(\Delta+\eta\), becomes smaller.

\begin{lemma}[Quantile lower bound]
\label{lem:quantile-lower}
Let \(\delta\in(0,1/2)\), and let \(\mathsf A\) be \(\delta\)-correct over an instance class \(\mathcal C\) containing \(\theta^\Delta\). Suppose that there exists \(\bar\eta>0\) such that,
for every \(i\in\{2,\ldots,K\}\) and \(r\in[d]\), the class
\(\mathcal C\) contains at least one alternative
\(\lambda^{i,r,\eta_{i,r}}\) with \(0<\eta_{i,r}\leq\bar\eta\).
Then, for every \(\alpha\in(0,1-2\delta)\),
\[
    Q_{\theta^\Delta,\alpha}(\mathsf A)
    \ge
    \frac{4d(K-1)}{(\Delta+\bar\eta)^2}
    \kl(1-\delta-\alpha,\delta).
\]

If, moreover, arbitrarily small frontier-changing perturbations
are admissible, namely, for every \(i\in\{2,\ldots,K\}\) and \(r\in[d]\) there exists a sequence \(\eta_n^{i,r}\downarrow0\) such that \( \lambda^{i,r,\eta_n^{i,r}}\in\mathcal C \text{ for every }n, \)
then
\[
    Q_{\theta^\Delta,\alpha}(\mathsf A)
    \ge
    \frac{4d(K-1)}{\Delta^2}
    \kl(1-\delta-\alpha,\delta).
\]
In particular, if \(\alpha=\delta<1/3\), then
\[
    Q_{\theta^\Delta,\delta}(\mathsf A)
    \ge
    \frac{4d(K-1)}{\Delta^2}
    \kl(1-2\delta,\delta)
    =
    \Omega\!\left(
        \frac{dK}{\Delta^2}\log\frac1\delta
    \right)
    \qquad
    \text{as }\delta\downarrow0.
\]
\end{lemma}

The lower bound holds under the full adaptive ordered-pair oracle of \Cref{sec:problem}, and therefore also applies to the uniform-opponent implementation used by VB-EGE. The proof uses a bounded-horizon change-of-measure argument \citep[Lemma~1]{kaufmann2016complexity}.

\begin{proof}[Proof of \Cref{lem:quantile-lower}]
It suffices to lower-bound every finite \(T\) satisfying
\begin{equation}
    \P_{\theta^\Delta}(\tau\leq T)
    \geq
    1-\alpha.
    \label{eq:lower-quantile-horizon}
\end{equation}
If no such \(T\) exists, the quantile is infinite and the claim is immediate.

Fix a suboptimal model \(i\in\{2,\ldots,K\}\), objective \(r\in[d]\), and an admissible \(\eta>0\). Under \(\theta^\Delta\), model \(1\) strictly dominates all others, so \(\calP(\theta^\Delta)=\{1\}\). Under \(\lambda^{i,r,\eta}\), model \(1\) does not dominate \(i\), because \(\lambda_{i,r}^{i,r,\eta}=\eta>0=\lambda_{1,r}^{i,r,\eta}\). No other model dominates \(i\), since its coordinate \(r\) is smaller than \(\eta\). Thus \(i\) is Pareto under the alternative and
\begin{equation}
    \calP(\lambda^{i,r,\eta})
    \neq
    \{1\}
    =
    \calP(\theta^\Delta).
    \label{eq:alternative-changes-frontier}
\end{equation}

Consider the stopped event
\[
    G_T
    :=
    \{\tau\leq T,\ \widehat\calP=\{1\}\}.
\]
By \(\delta\)-correctness and
\eqref{eq:lower-quantile-horizon},
\[
    \P_{\theta^\Delta}(G_T)
    \geq
    1-\delta-\alpha.
\]
By \eqref{eq:alternative-changes-frontier}, outputting \(\{1\}\) under
\(\lambda^{i,r,\eta}\) is an error, so
\[
    \P_{\lambda^{i,r,\eta}}(G_T)
    \leq
    \delta.
\]
Since
\[
    \mathbb P_{\theta^\Delta}(G_T)
    \geq
    1-\delta-\alpha
    >
    \delta
    \geq
    \mathbb P_{\lambda^{i,r,\eta}}(G_T),
\]
\Cref{lem:binary-kl-monotonicity} yields
\begin{equation}
    \kl\!\left(
        \mathbb P_{\theta^\Delta}(G_T),
        \mathbb P_{\lambda^{i,r,\eta}}(G_T)
    \right)
    \geq
    \ell_{\delta,\alpha}^{\mathrm{hp}}
    :=
    \kl(1-\delta-\alpha,\delta).
    \label{eq:event-kl-lower-bound}
\end{equation}

The two instances differ only in objective \(r\) of model \(i\). Thus all queries outside objective \(r\), or not involving \(i\), contribute zero KL in \Cref{lem:bounded-horizon-transport}. For a comparison in objective \(r\) involving \(i\), the Bradley--Terry logit changes by at most \(\Delta+\eta\): it changes from \(-\Delta\) to \(\eta\) against model \(1\), and from \(0\) to \(\Delta+\eta\) against any other suboptimal model. By \Cref{lem:logistic-kl-upper}, every affected comparison contributes at most \((\Delta+\eta)^2/8\). Combining \Cref{lem:bounded-horizon-transport} with \eqref{eq:event-kl-lower-bound} yields
\[
    \ell_{\delta,\alpha}^{\mathrm{hp}}
    \leq
    \frac{(\Delta+\eta)^2}{8}
    \E_{\theta^\Delta}\left[N_{i,r}^{\mathrm{inc}}(\tau\wedge T)\right].
\]
Therefore
\[
    \E_{\theta^\Delta}\left[N_{i,r}^{\mathrm{inc}}(\tau\wedge T)\right]
    \geq
    \frac{8\ell_{\delta,\alpha}^{\mathrm{hp}}}
         {(\Delta+\eta)^2}.
\]
By the assumption of \Cref{lem:quantile-lower}, for every
\(i\in\{2,\ldots,K\}\) and \(r\in[d]\), one can choose
\(\eta_{i,r}\in(0,\bar\eta]\) such that
\(\lambda^{i,r,\eta_{i,r}}\in\mathcal C\). Therefore,
\[
    \mathbb E_{\theta^\Delta}
    \!\left[
        N_{i,r}^{\mathrm{inc}}(\tau\wedge T)
    \right]
    \ge
    \frac{8\ell_{\delta,\alpha}^{\mathrm{hp}}}
    {(\Delta+\eta_{i,r})^2}
    \ge
    \frac{8\ell_{\delta,\alpha}^{\mathrm{hp}}}
    {(\Delta+\bar\eta)^2}.
\]
Summing over \(i\in\{2,\ldots,K\}\) and \(r\in[d]\) gives
\[
    \sum_{i=2}^K\sum_{r=1}^d
    \mathbb E_{\theta^\Delta}
    \!\left[
        N_{i,r}^{\mathrm{inc}}(\tau\wedge T)
    \right]
    \ge
    \frac{8d(K-1)\ell_{\delta,\alpha}^{\mathrm{hp}}}
    {(\Delta+\bar\eta)^2}.
\]

On every sample path,
\[
    \sum_{i=2}^K\sum_{r=1}^d
    N_{i,r}^{\mathrm{inc}}(\tau\wedge T)
    \leq
    2(\tau\wedge T)
    \leq
    2T,
\]
because one objective-specific binary comparison is incident to at most two
models in \(\{2,\ldots,K\}\). Combining the two
bounds gives
\[
    T
    \ge
    \frac{4d(K-1)}
    {(\Delta+\bar\eta)^2}
    \kl(1-\delta-\alpha,\delta).
\]
Taking the infimum over all finite \(T\) satisfying
\[
    \mathbb P_{\theta^\Delta}(\tau\le T)\ge1-\alpha
\]
proves the first assertion.

The sequence assumption in the second part of \Cref{lem:quantile-lower} implies the preceding admissibility condition for every \(\bar\eta>0\). Applying the first bound with arbitrary \(\bar\eta>0\) and letting \(\bar\eta\downarrow0\) gives
\[
    Q_{\theta^\Delta,\alpha}(\mathsf A)
    \ge
    \frac{4d(K-1)}{\Delta^2}
    \kl(1-\delta-\alpha,\delta).
\]
The specialization \(\alpha=\delta<1/3\) follows from
\[
    \kl(1-2\delta,\delta)
    =
    \Theta\!\left(\log\frac1\delta\right)
    \qquad
    \text{as }\delta\downarrow0.
\]
\end{proof}

\begin{remark}[Admissible alternatives]
\label{rem:admissible-alternatives}
For the bounded-range class $\Theta_{B,d}$, the small-perturbation condition in the second part of \Cref{lem:quantile-lower} holds whenever \(0<\Delta<2B\). For example, one may take
$\eta_n:=\frac{2B-\Delta}{n+1}\downarrow0$.

This condition cannot hold on the gap-separated subclass \(\Theta_{B,d}(\Delta)\). Indeed, for every \(\eta>0\),
$\Delta_i\!\left(\lambda^{i,r,\eta}\right)
    \le\eta$, so the alternatives eventually leave \(\Theta_{B,d}(\Delta)\) as \(\eta\downarrow0\). However, if \(\Delta\le B\), then
\[
    \theta^\Delta,\lambda^{i,r,\Delta}
    \in\Theta_{B,d}(\Delta)
    \qquad
    \text{for every }i\ge2,\ r\in[d].
\]
The first part of \Cref{lem:quantile-lower} therefore applies with $\bar\eta=\Delta$, which is the construction used in \Cref{thm:minimax-optimality}.
\end{remark}

We now apply the preceding lemma to prove the minimax lower bound.

\begin{proof}[Proof of \Cref{thm:minimax-optimality}]
Let \(\mathsf A\) be any algorithm that is \(\delta\)-correct on \(\Theta_{B,d}(\Delta)\), and consider the symmetric instance \(\theta^\Delta\) from \Cref{lem:quantile-lower}. Directly from \Cref{def:classification-gap},
\[
    \mathcal P(\theta^\Delta)=\{1\},
    \qquad
    \Delta_a(\theta^\Delta)=\Delta
    \quad\text{for every }a\in[K].
\]

For every \(i\ge2\) and \(r\in[d]\), let \(\lambda^{i,r,\Delta}\) replace only
\(\theta^\Delta_{i,r}=-\Delta\) by \(+\Delta\). A direct calculation gives
\[
    \mathcal P(\lambda^{i,r,\Delta})
    =
    \begin{cases}
        \{i\}, & d=1,\\
        \{1,i\}, & d\ge2,
    \end{cases}
\]
and
\[
    \Delta_1(\lambda^{i,r,\Delta})
    =
    \Delta_i(\lambda^{i,r,\Delta})
    =
    \Delta,
    \qquad
    \Delta_k(\lambda^{i,r,\Delta})
    \ge\Delta
    \quad
    \text{for every }k\notin\{1,i\}.
\]
Moreover,
\[
    \max_{s\in[d]}\max_{a,b\in[K]}
    |\theta^\Delta_{a,s}-\theta^\Delta_{b,s}|
    =\Delta,
\]
whereas
\[
    \max_{s\in[d]}\max_{a,b\in[K]}
    |\lambda^{i,r,\Delta}_{a,s}
      -\lambda^{i,r,\Delta}_{b,s}|
    \le2\Delta.
\]
Since \(\Delta\le B\),
\[
    \theta^\Delta,\lambda^{i,r,\Delta}
    \in\Theta_{B,d}(\Delta).
\]

Since \(\lambda^{i,r,\Delta}\in\Theta_{B,d}(\Delta)\) for every \(i\ge2\) and \(r\in[d]\), the first part of \Cref{lem:quantile-lower} applies with \(\bar\eta=\Delta\) and \(\alpha=\delta\). Hence
\[
    \sup_{\theta\in\Theta_{B,d}(\Delta)}
    Q_{\theta,\delta}(\mathsf A)
    \ge
    Q_{\theta^\Delta,\delta}(\mathsf A)
    \ge
    \frac{d(K-1)}{\Delta^2}
    \kl(1-2\delta,\delta).
\]
Taking the infimum over all such \(\mathsf A\) proves the minimax lower bound.
\end{proof}

\subsubsection{Proof of \Cref{cor:minimax-opt}}

\begin{proof}[Proof of \Cref{cor:minimax-opt}]
For every \(\theta\in\Theta_{B,d}(\Delta)\),
\[
\begin{aligned}
    &\Delta_i^\theta\ge\Delta,
    \qquad
    M_i^\theta\le \left\lceil\log_2\frac{8}{\kappa_B\Delta}\right\rceil,\\
    &L_i^\theta(\delta)
    \le
    \log\frac{4Kd\left(\left\lceil\log_2\frac{8}{\kappa_B\Delta}\right\rceil\right)^2}{\delta}.
\end{aligned}
\]
Hence \Cref{thm:upper} gives
\[
    \sup_{\theta\in\Theta_{B,d}(\Delta)}
    Q_{\theta,\delta}(\VBEGE)
    \le
    dK+
    \frac{128dK}{\kappa_B^2\Delta^2}
    \log\frac{4Kd\left(\left\lceil\log_2\frac{8}{\kappa_B\Delta}\right\rceil\right)^2}{\delta}.
\]
The joint recovery statement in \Cref{thm:upper} also shows that VB-EGE is \(\delta\)-correct on \(\Theta_{B,d}(\Delta)\). It is therefore admissible in the infimum defining \(\mathbf Q_{B,d}(\Delta,\delta)\), and hence
\[
    \mathbf Q_{B,d}(\Delta,\delta)
    \le
    \sup_{\theta\in\Theta_{B,d}(\Delta)}
    Q_{\theta,\delta}(\VBEGE).
\]

Combining this upper bound with \Cref{thm:minimax-optimality} gives the minimax rate. Since \(K\ge2\), we have \(K-1\ge K/2\). Finally, for fixed \(B\), \(\kappa_B\) is bounded below by a positive constant, while
\[
    \kl(1-2\delta,\delta)
    =
    \Theta\!\left(\log\frac1\delta\right)
    \qquad
    \text{as }\delta\downarrow0.
\]
Moreover, because \(\Delta\le1\) and \(\delta<1/3\), the additive \(dK\) term is absorbed by the displayed upper-bound term. Hence VB-EGE matches the polynomial dependence \(dK\Delta^{-2}\) up to logarithmic factors in \(K\), \(d\), and \(\Delta^{-1}\), and matches the \(\log(1/\delta)\) confidence dependence.
\end{proof}

\section{Additional details and results for synthetic experiments}
\label{app:synthetic-experiments}


\subsection{Details for the data-generating process}
\label{app:synthetic-benchmark-details}
In the scaling experiments, the parameter grids are specified in Table~\ref{tab:synthetic-main-settings}.

In the Convex-3D, Witness, and Arena settings, we first generate $s$ Pareto models 
and then construct the dominated models by lowering the scores accordingly.
The latent scores corresponding to the 
Pareto models are  given by $\theta_i = 0.15\mathbf 1_d+0.75z_i$, where 
\(z_i\overset{\text{iid}}{\sim}\operatorname{Dirichlet}(\alpha\mathbf 1_d)\), $\forall i\in [s]$; 
thus \(\alpha\) controls their concentration on the trade-off simplex.
For Convex-2D, the $s$ Pareto anchors are deterministic and equally spaced on a decreasing line segment:
\[
t_p=\frac{p-1}{s-1}, \qquad
\theta_p=(0.20+0.65t_p,\;0.85-0.65t_p), \qquad p\in[s].
\]
In the synthetic Arena and Convex settings, each dominated arm $i\notin[s]$ is 
independently assigned an anchor
$P_i\sim\operatorname{Unif}([s])$ and its latent score is generated as 
\begin{align}
\label{eq:benchmark-dominated}
\theta_i=\theta_{P_i}-m_i \text{ where }m_{i,r}\overset{\mathrm{iid}}{\sim}\operatorname{Unif}[\ell,u]
\quad \forall i \notin [s].
\end{align}
For Convex-2D, $[\ell,u]=[0.03,0.18]$.
For the Witness setting, exactly \(q\) dominated arms are attached to each Pareto anchor 
\(p\in[s]\), so $K=s(q+1)$. Candidate scores are generated according to~\eqref{eq:benchmark-dominated} with $P_i=p$ and retained only when $p$ is their unique Pareto dominator.
In the Two-group-10 setting, we generate \(K_{\mathrm{low}}\) low-group arms 
and \(K_{\mathrm{high}}\) high-group arms, with $K = K_{\mathrm{low}}+K_{\mathrm{high}}$.
Each coordinate is drawn independently from \(\operatorname{Unif}[0.20,0.45]\)
for the low-group arms and from \(\operatorname{Unif}[0.55,0.75]\) for the high-group arms.
Because every high-group coordinate exceeds every low-group coordinate, 
each low-group arm is strictly dominated by every high-group arm. 
The strict Pareto frontier therefore consists entirely of high-group arms, 
with its membership varying across realizations.
The specific parameters used in each setting are reported in Table~\ref{tab:synthetic-benchmark-settings}.



For Convex-3D and Arena-10, we generate 50 sets of latent model scores and simulate comparisons independently 10 times for each setting. The other benchmarks use 500 independently generated instances.  For benchmarks with repeated observations of the same instance, standard errors are computed from the 50 instance-level averages.


\begin{table}[htb]
\centering
\caption{Parameters used in the synthetic benchmark experiments.}  
\label{tab:synthetic-benchmark-settings}
\experimenttablesize
\setlength{\tabcolsep}{4pt}
\begin{tabular}{@{}llrrrp{0.30\textwidth}@{}}
\toprule
Setting & Generator & \(K\) & \(d\) & \(|\calP|\) & Parameters \\
\midrule
Convex-2D & Convex frontier & 60 & 2 & 15 & 
\(s=15\), margins \([0.03,0.18]\) \\
Convex-3D & Convex frontier & 60 & 3 & 15 &  \(s=15\), simplex \(\alpha=1\), margins \([0.03,0.18]\) \\
Witness-4 & Unique witness & 40 & 4 & 8 & 
\(s=8,q=4\), margins \([0.04,0.16]\), \(\alpha=0.7\) \\
Witness-10 & Unique witness & 80 & 10 & 16 & 
\(s=16,q=4\), margins \([0.035,0.14]\), \(\alpha=0.7\) \\
Arena-4 small & Synthetic Arena & 32 & 4 & 8 &  \(s=8\), margins \([0.08,0.25]\), \(\alpha=0.7\) \\
Arena-4 medium & Synthetic Arena & 64 & 4 & 12 &  \(s=12\), margins \([0.08,0.25]\), \(\alpha=0.7\) \\
Arena-10 medium & Synthetic Arena & 64 & 10 & 16 &  \(s=16\), margins \([0.06,0.20]\), \(\alpha=0.7\) \\
Two-group-10 & High-dimensional two-group & 64 & 10 & 15.73 &  \(K_{\mathrm{low}}=48,K_{\mathrm{high}}=16\) \\
\bottomrule
\end{tabular}
\end{table}

\subsection{Implementation details for the baseline models}
\label{app:synthetic-protocol}

For {\em BT-MLE}, at each phase, 
we refit a centered ridge-regularized Bradley--Terry model 
per objective using all accumulated comparisons.
For numerical stability, we use a ridge penalty coefficient of \(10^{-8}\) and 
an optimization tolerance of \(10^{-9}\). If the optimizer fails to converge, 
any fitted score is nonfinite or exceeds \(20\) in absolute value, 
or the comparison graph is disconnected, we refit with coefficient \(10^{-4}\).

For {\em Borda plug-in}, let \(N_{ij,r}^{\mathrm{pair}}(t)\) denote the number of comparisons between models \(i\) and \(j\) on objective \(r\) up to query \(t\), and let \(W_{ij,r}^{\mathrm{on}}(t)\) denote the number of these comparisons won by \(i\).
For \(i<j\), we estimate the probability that \(i\) beats \(j\) by
\[
    \widetilde p_{ij}^{(r)}(t)
    :=\frac{W_{ij,r}^{\mathrm{on}}(t)+1/2}
    {N_{ij,r}^{\mathrm{pair}}(t)+1},
    \qquad
    \widetilde p_{ji}^{(r)}(t)
    :=1-\widetilde p_{ij}^{(r)}(t).
\]
The added counts prevent estimates of exactly zero or one.
We estimate each model's Borda score on objective \(r\) by averaging its estimated win probabilities over all other models:
\[
    \widehat b_{i,r}(t)
    :=\frac{1}{K-1}\sum_{j\ne i}\widetilde p_{ij}^{(r)}(t).
\]

\subsection{Additional results from synthetic experiments}
\label{app:synthetic-benchmark-add-results}
\Cref{tab:synthetic-scaling-results} and~\Cref{tab:synthetic-benchmark-results}
report the numerical results underlying (i)-(iii) of~\Cref{fig:symmetric-hard-scaling} 
and~\Cref{fig:synthetic-benchmark-overview}, respectively.
\Cref{tab:confidence-symmetric-results} provides the results for  
panel (iv) of~\Cref{fig:symmetric-hard-scaling},
while~\Cref{tab:confidence-arena-results} reports  
confidence for the Arena-10 benchmark.
Here, $q_{0.95}(\tau)$ denotes the $95$th quantile of the stopping time.
Fitting a linear model of the mean stopping time on $\log(1/\delta)$
gives \(R^2\approx0.99997\) for the hard-instance setting and 
\(0.9978\) for the Arena-10 benchmark, with the same ordering at the 95th percentile. 
These results support logarithmic scaling of stopping cost with $1/\delta$,
although we note that with only $300 - 500$ replications, 
the experiments cannot reliably estimate failure probabilities on the order of 
$10^{-3}$.


\begin{table}[htb]
\centering
\caption{Fixed-confidence scaling results. Parentheses contain one standard error for \VBEGE; remaining columns report each baseline's mean comparison count divided by that of VB-EGE.}
\label{tab:synthetic-scaling-results}
\experimenttablesize
\begin{tabular}{@{}llrrrr@{}}
\toprule
Sweep & Value & VB mean \(\tau\) (SE) & Focal/VB & MLE/VB & Plug-in/VB \\
\midrule
\(K\) & 16  & \(3.63\times10^5\) (678) & 1.02 & 0.51 & 9.03 \\
      & 32  & \(7.85\times10^5\) (411) & 1.00 & 1.08 & 19.1 \\
      & 64  & \(1.66\times10^6\) (457) & 1.00 & 2.29 & 40.1 \\
      & 128 & \(3.51\times10^6\) (301) & 1.00 & 4.80 & 83.2 \\
\addlinespace
\(d\) & 2  & \(7.65\times10^5\) (\(1.63\times10^3\)) & 1.03 & 2.38 & 41.7 \\
      & 4  & \(1.66\times10^6\) (405) & 1.00 & 2.29 & 40.1 \\
      & 10 & \(4.46\times10^6\) (0) & 1.00 & 2.27 & 39.5 \\
\addlinespace
\(\Delta\) & 0.50 & \(1.84\times10^7\) (\(2.78\times10^3\)) & 1.00 & 2.31 & 39.3 \\
           & 0.75 & \(4.46\times10^6\) (0) & 1.00 & 2.27 & 39.5 \\
           & 1.00 & \(4.46\times10^6\) (0) & 1.00 & 2.27 & 39.5 \\
           & 1.25 & \(1.82\times10^6\) (\(2.04\times10^4\)) & 2.45 & 1.28 & 23.4 \\
           & 1.50 & \(1.07\times10^6\) (212) & 1.01 & 2.17 & 39.8 \\
\bottomrule
\end{tabular}
\end{table}

\begin{table}[htb]
\centering
\caption{Mean comparison counts for VB-EGE and ratios of mean comparison counts (baseline/VB-EGE). Parentheses contain one standard error. Ratio standard errors use the paired delta method, with instance-level averages as the independent units when multiple trials share the same latent instance.}
\label{tab:synthetic-benchmark-results}
\experimenttablesize
\setlength{\tabcolsep}{4pt}
\begin{tabular}{@{}lrrrr@{}}
\toprule
Setting & VB mean \(\tau\) (SE) & Focal/VB (SE) & MLE/VB (SE) & Plug-in/VB (SE) \\
\midrule
Convex-2D & \(1.21\times10^8\) (\(9.57\times10^5\)) & 3.55 (0.10) & 6.90 (0.21) & 118 (3.6) \\
Convex-3D & \(4.46\times10^9\) (\(3.86\times10^9\)) & 28.1 (1.6) & 62.5 (3.7) & 1015 (60) \\
Witness-4 & \(2.28\times10^8\) (\(7.27\times10^5\)) & 1.25 (0.0040) & 1.79 (0.0057) & 29.7 (0.094) \\
Witness-10 & \(1.60\times10^9\) (\(3.26\times10^6\)) & 1.81 (0.060) & 2.97 (0.0061) & 49.0 (0.10) \\
Arena-4 small & \(3.83\times10^{10}\) (\(3.30\times10^{10}\)) & 15.7 (0.35) & 18.3 (0.80) & 302 (14) \\
Arena-4 medium & \(2.20\times10^{10}\) (\(9.61\times10^9\)) & 31.6 (0.25) & 69.0 (5.7) & 1224 (9.6) \\
Arena-10 medium & \(6.15\times10^8\) (\(5.87\times10^6\)) & 2.05 (0.020) & 4.80 (0.046) & 79.0 (0.75) \\
Two-group-10 & \(4.48\times10^{11}\) (\(3.69\times10^{11}\)) & 111 (16) & 261 (36) & 4227 (590) \\
\bottomrule
\end{tabular}
\end{table}

\begin{table}[htb]
\centering
\begin{minipage}[t]{0.49\textwidth}
\vspace{0pt}
\centering
\caption{Confidence scaling on the symmetric \(K=64,d=10\) task.}
\label{tab:confidence-symmetric-results}
\experimenttablesize
\setlength{\tabcolsep}{3pt}
\resizebox{\linewidth}{!}{%
\begin{tabular}{@{}lrrrrr@{}}
\toprule
\(\delta\) & VB mean \(\tau\) (SE) & VB \(q_{0.95}(\tau)\) &
Focal/VB & MLE/VB & Plug-in/VB \\
\midrule
0.200 & \(4.01\times10^6\) (0) & \(4.01\times10^6\) & 1.00 & 2.30 & 40.4 \\
0.100 & \(4.23\times10^6\) (0) & \(4.23\times10^6\) & 1.00 & 2.29 & 39.9 \\
0.050 & \(4.46\times10^6\) (0) & \(4.46\times10^6\) & 1.00 & 2.27 & 39.5 \\
0.020 & \(4.76\times10^6\) (0) & \(4.76\times10^6\) & 1.00 & 2.25 & 39.0 \\
0.010 & \(4.99\times10^6\) (0) & \(4.99\times10^6\) & 1.00 & 2.24 & 38.6 \\
0.005 & \(5.22\times10^6\) (0) & \(5.22\times10^6\) & 1.00 & 2.23 & 38.3 \\
0.002 & \(5.52\times10^6\) (0) & \(5.52\times10^6\) & 1.00 & 2.21 & 38.0 \\
0.001 & \(5.74\times10^6\) (0) & \(5.74\times10^6\) & 1.00 & 2.20 & 37.7 \\
\bottomrule
\end{tabular}%
}
\end{minipage}\hfill
\begin{minipage}[t]{0.49\textwidth}
\vspace{0pt}
\centering
\caption{Confidence scaling on the Arena-10 medium task.}
\label{tab:confidence-arena-results}
\experimenttablesize
\setlength{\tabcolsep}{3pt}
\resizebox{\linewidth}{!}{%
\begin{tabular}{@{}lrrrrr@{}}
\toprule
\(\delta\) & VB mean \(\tau\) (SE) & VB \(q_{0.95}(\tau)\) &
Focal/VB & MLE/VB & Plug-in/VB \\
\midrule
0.200 & \(5.62\times10^8\) (\(3.69\times10^6\)) & \(6.57\times10^8\) & 2.09 & 4.99 & 82.7 \\
0.100 & \(5.89\times10^8\) (\(3.63\times10^6\)) & \(6.91\times10^8\) & 2.10 & 5.01 & 81.8 \\
0.050 & \(6.21\times10^8\) (\(3.63\times10^6\)) & \(7.10\times10^8\) & 2.09 & 4.95 & 80.9 \\
0.020 & \(6.49\times10^8\) (\(4.46\times10^6\)) & \(7.69\times10^8\) & 2.12 & 4.93 & 81.1 \\
0.010 & \(6.81\times10^8\) (\(4.20\times10^6\)) & \(7.86\times10^8\) & 2.13 & 4.92 & 80.2 \\
0.005 & \(7.06\times10^8\) (\(4.67\times10^6\)) & \(8.20\times10^8\) & 2.12 & 4.87 & 79.9 \\
\bottomrule
\end{tabular}%
}
\end{minipage}
\end{table}






\begin{figure}[htb]
\centering
\includegraphics[width=0.5\linewidth]{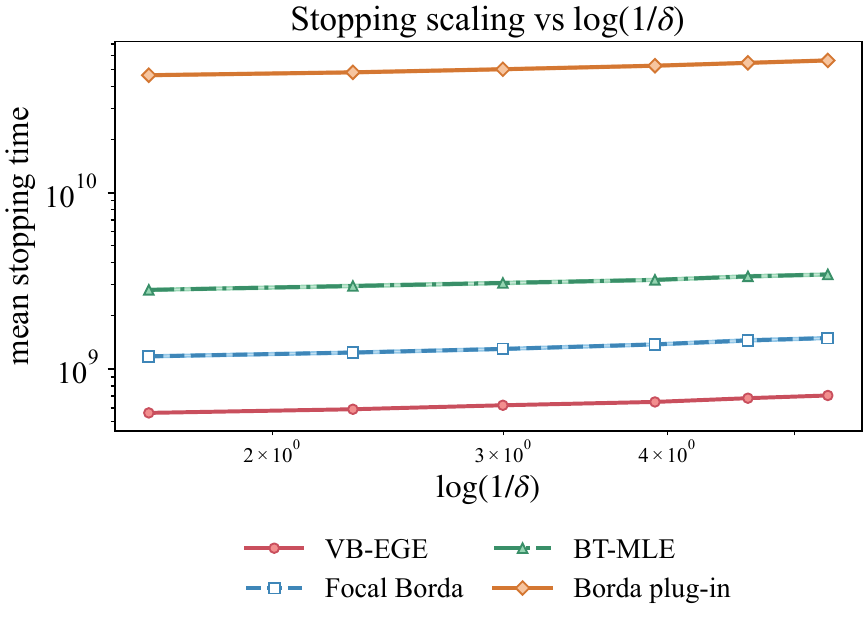}
\caption{Arena-10 medium mean stopping time (mean \(\pm\) one standard error) versus \(\log(1/\delta)\).}
\label{fig:confidence-scaling-details}
\end{figure}

\subsection{Hyperparameter sensitivity analysis}
\label{app:synthetic-constants}

\begin{figure}[H]
\centering
\begin{subfigure}[t]{0.3\textwidth}
\centering
\includegraphics[width=\linewidth]{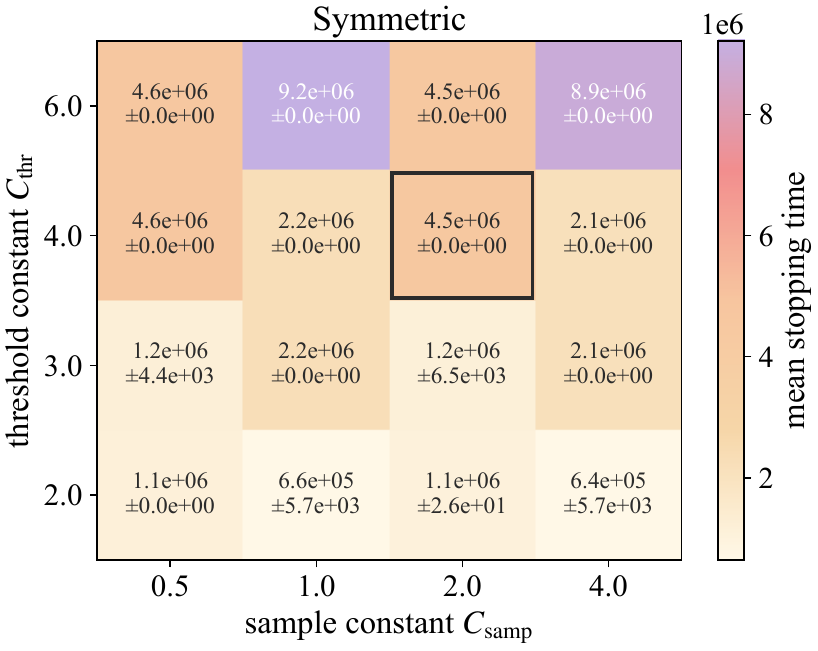}
\caption{Symmetric \(K=64,d=10\).}
\end{subfigure}
\hfill
\begin{subfigure}[t]{0.3\textwidth}
\centering
\includegraphics[width=\linewidth]{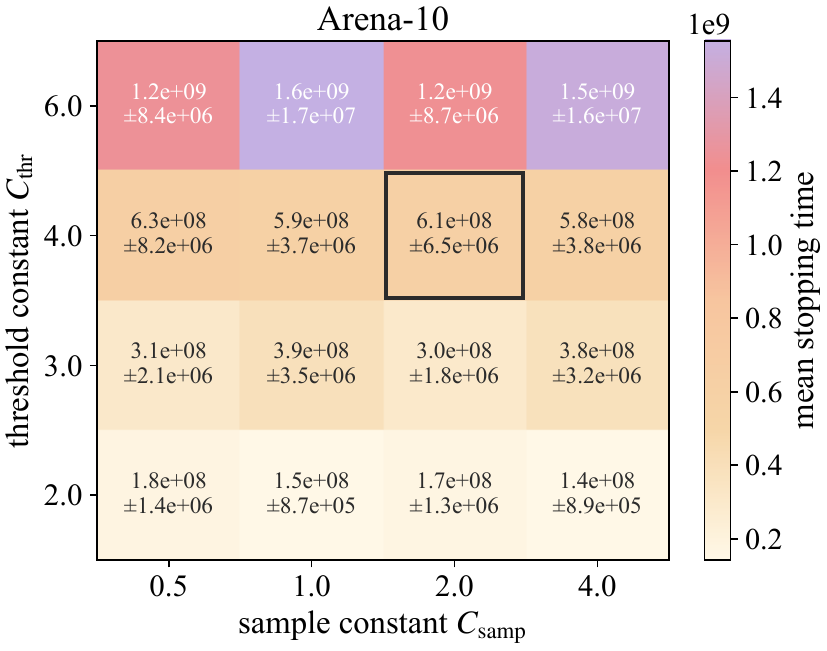}
\caption{Arena-10 medium.}
\end{subfigure}
\hfill
\begin{subfigure}[t]{0.3\textwidth}
\centering
\includegraphics[width=\linewidth]{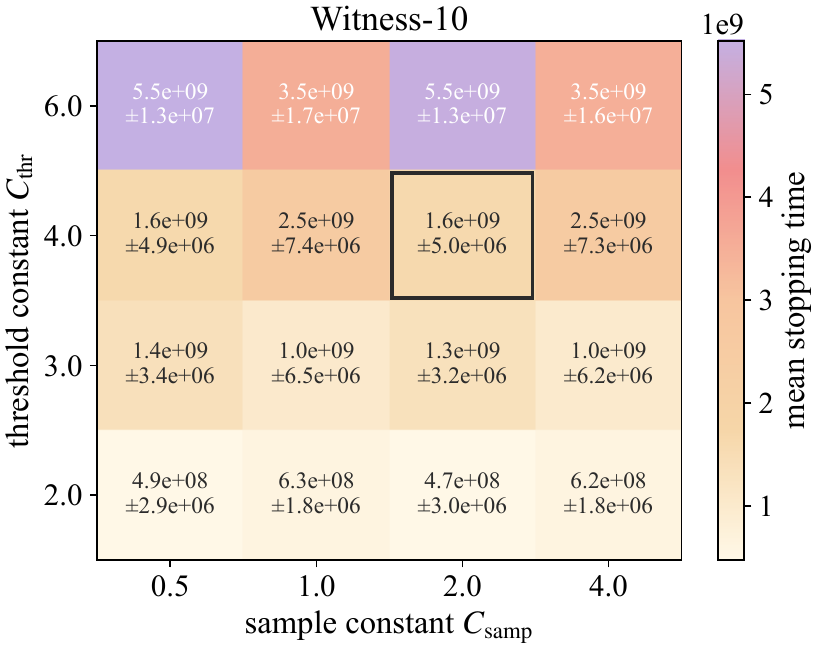}
\caption{Witness-10.}
\end{subfigure}
\caption{Mean stopping-time with respect to difference choices of \((C_{\mathrm{samp}},C_{\mathrm{thr}})\). 
Each cell reports stopping time (mean \(\pm\) one standard error). The black rectangle corresponds to the default choice.}
\label{fig:constant-sensitivity-heatmaps}
\end{figure}

We now examine sensitivity to the sampling and removal-threshold parameters by varying $C_{\mathrm{samp}}\in\{0.5,1,2,4\}$ and $C_{\mathrm{thr}}\in\{2,3,4,6\}$, while fixing $C_{\log}=4$ and $\delta=0.05$. 
We consider three settings: (1) the hard instance considered in the scaling studies with 
$K = 64$, $d = 10$, (2) the Arena-10 medium setting, and (3) the Witness-10 setting. 
We note that configurations other than the default are not covered by \Cref{thm:upper} as stated.
\Cref{fig:constant-sensitivity-heatmaps} present heatmaps 
corresponding to the sensitivity analysis results.
We see that smaller \(C_{\mathrm{thr}}\) generally yields smaller stopping time.

\subsection{Pareto-set size sensitivity analysis}
\label{app:pareto-size}

\begin{figure}[H]
\centering
\begin{subfigure}[t]{0.49\textwidth}
\centering
\includegraphics[width=\linewidth]{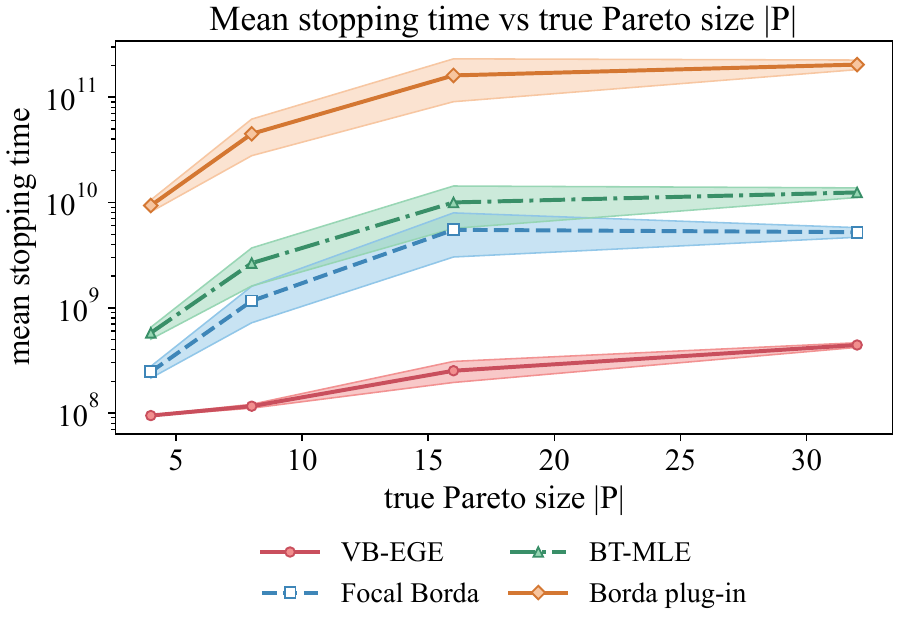}
\caption{Arena-4.}
\end{subfigure}
\hfill
\begin{subfigure}[t]{0.49\textwidth}
\centering
\includegraphics[width=\linewidth]{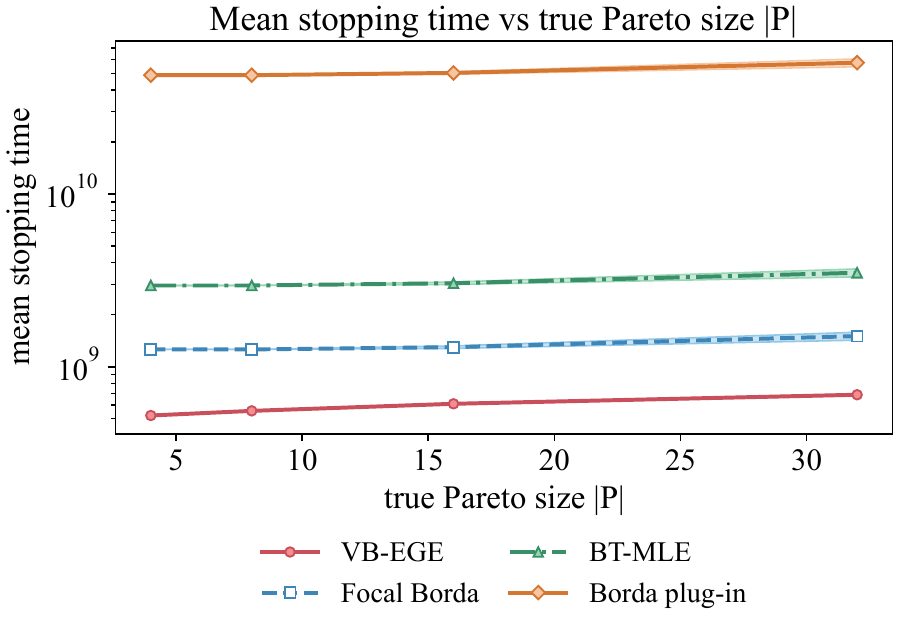}
\caption{Arena-10.}
\end{subfigure}
\caption{Mean stopping time (mean \(\pm\) one standard error) as the true Pareto set grows.}
\label{fig:pareto-size-results}
\end{figure}

We hold \(K=64\) and the synthetic Arena generator family fixed while varying \(|\calP|\in\{4,8,16,32\}\). Arena-4 uses margins \([0.08,0.25]\), Arena-10 uses \([0.06,0.20]\), and each parameter setting has 300 trials.
The results are reported in~\Cref{tab:pareto-size-results} and visualized in~\Cref{fig:pareto-size-results}, where we see that 
stopping cost increases with \(|\calP|\), while every tested ratio of mean comparison counts (baseline/VB-EGE) remains above one. 
The adaptive advantage therefore persists across frontier sizes \(4\)--\(32\).

\begin{table}[H]
\centering
\caption{Sensitivity to Pareto-set size.  Parentheses contain one standard error for \VBEGE; other entries report each baseline's mean comparison count divided by that of VB-EGE.}
\label{tab:pareto-size-results}
\experimenttablesize
\begin{tabular}{@{}lrrrrr@{}}
\toprule
\(d\) & \(|\calP|\) & VB mean \(\tau\) (SE) & Focal/VB & MLE/VB & Plug-in/VB \\
\midrule
4 & 4  & \(9.44\times10^7\) (\(7.02\times10^5\)) & 2.60 & 6.11 & 99.1 \\
4 & 8  & \(1.16\times10^8\) (\(4.91\times10^6\)) & 10.0 & 22.9 & 387 \\
4 & 16 & \(2.52\times10^8\) (\(5.70\times10^7\)) & 21.8 & 39.7 & 640 \\
4 & 32 & \(4.42\times10^8\) (\(2.22\times10^7\)) & 11.8 & 28.2 & 462 \\
\addlinespace
10 & 4  & \(5.21\times10^8\) (\(2.79\times10^6\)) & 2.41 & 5.66 & 93.5 \\
10 & 8  & \(5.55\times10^8\) (\(3.11\times10^6\)) & 2.27 & 5.32 & 87.9 \\
10 & 16 & \(6.10\times10^8\) (\(4.21\times10^6\)) & 2.13 & 4.99 & 82.4 \\
10 & 32 & \(6.88\times10^8\) (\(6.87\times10^6\)) & 2.19 & 5.09 & 83.7 \\
\bottomrule
\end{tabular}
\end{table}

\subsection{Sampling-mechanism diagnostics}
\label{app:sampling-diagnostics}



\begin{figure}[H]
\centering
\includegraphics[width=0.92\textwidth]{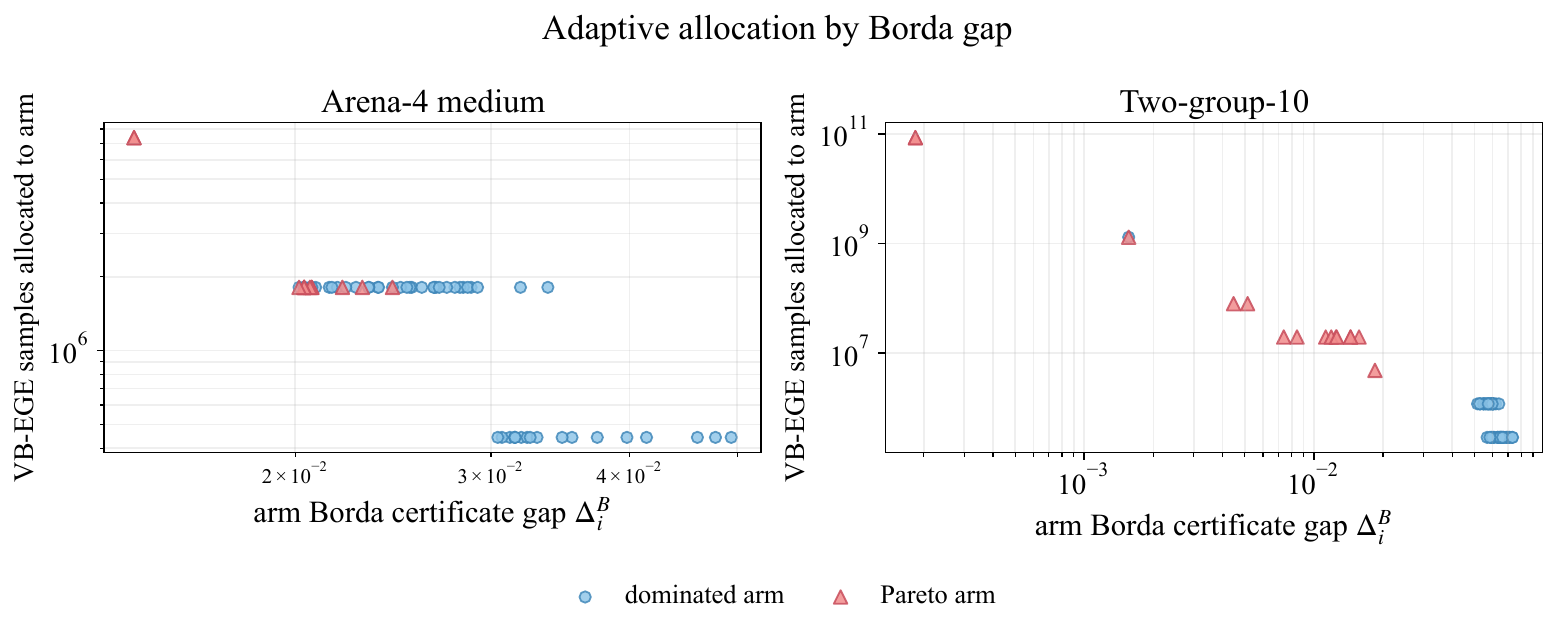}
\caption{Total \VBEGE\ allocation to each model versus its Borda
classification gap on representative Arena-4 medium and Two-group-10
instances.  Smaller-gap models receive more samples.}
\label{fig:allocation-versus-gap}
\end{figure}

Figure~\ref{fig:allocation-versus-gap} plots the per-arm 
sample allocation against the Borda classification gap, 
showing that VB-EGE collects more observations for models with smaller gaps.

\FloatBarrier

\section{Additional details and results for real-data experiments}
\label{app:real-data}

This section describes preprocessing and sampling from the three real-world Arena datasets, followed by detailed results and sensitivity to the treatment of non-decisive outcomes.

\subsection{Datasets, preprocessing, and retained models}
\label{app:real-data-protocol}

\paragraph{Dataset versions and definition of the objectives.}
We fix the dataset versions before preprocessing. 
Comparisons are assigned to objectives solely from the released metadata, 
without using the prompt or response text.
\begin{itemize}
    \item \textbf{Arena-Human-Preference-140K (Arena-140K):}
    \href{https://huggingface.co/datasets/lmarena-ai/arena-human-preference-140k}
    {\texttt{lmarena-ai/arena-human-preference-140k}}.
    \par Revision:
    \nolinkurl{6322995ab34d7c2693e3f47dd13fa5caa0789a74}.
    We construct five mutually exclusive prompt categories: \texttt{general}, \texttt{code}, \texttt{math}, \texttt{creative\_writing}, and \texttt{instruction\_following}. 
    The four specialized categories are determined by the following metadata fields: 
    \begin{itemize}
        \item \texttt{is\_code} for \texttt{code},
        \item \texttt{category\_tag.math\_v0.1.math} for \texttt{math}, 
        \item \texttt{category\_tag.creative\_writing\_v0.1.creative\_writing} for \texttt{creative\_writing},
        \item \texttt{category\_tag.if\_v0.1.if} for \texttt{instruction\_following}.
    \end{itemize}   
    Because the released indicators may overlap, we assign each comparison 
    to the first applicable category according to the prespecified priority ordering:
    \[
        \texttt{code}
        \;>\;
        \texttt{math}
        \;>\;
        \texttt{creative\_writing}
        \;>\;
        \texttt{instruction\_following}.
    \]
    Comparisons with no active specialized indicators are assigned to the \texttt{general} category.
    For example, a comparison marked as both code and math is assigned to \texttt{code}. 

    \item \textbf{Search-Arena-24K:}
    \href{https://huggingface.co/datasets/lmarena-ai/search-arena-24k}
    {\texttt{lmarena-ai/search-arena-24k}}.
    \par Revision:
    \nolinkurl{fac8dcf86146c8773ef020095c5694c9b2bc98d7}.
    We use the nine released values of the categorical \texttt{primary\_intent} field as objectives: Factual Lookup, Info Synthesis, Recommendation, Analysis, Explanation, Guidance, Creative Generation, Other, and Text Processing. Since \texttt{primary\_intent} is single-valued, each retained comparison contributes to at most one objective. Comparisons with a missing or unsupported intent are excluded.

    \item \textbf{VisionArena-Battle:}
    \href{https://huggingface.co/datasets/lmarena-ai/VisionArena-Battle}
    {\texttt{lmarena-ai/VisionArena-Battle}}.
    \par Revision:
    \nolinkurl{a41d58f15caa3e2d9685774153d2bc815b943d96}.
    We use the eight released Boolean category indicators directly: \texttt{captioning}, \texttt{creative\_writing}, \texttt{diagram}, \texttt{entity\_recognition}, \texttt{homework}, \texttt{humor}, \texttt{is\_code}, and \texttt{ocr}. These annotations are multi-label by construction. A comparison contributes once to every pair--objective cell whose corresponding indicator is true. Consequently, the number of pair--objective assignments may exceed the number of unique source comparisons; these repeated assignments represent overlapping category membership.
\end{itemize}

We retain only comparisons in which either model wins. 
In Arena-140K, we remove \(21{,}528\) \texttt{tie} and \(15{,}749\) \texttt{both\_bad} records.
In Search-Arena-24K, we remove \(2{,}282\) \texttt{tie}, \(1{,}442\) \texttt{both\_bad} records, \(10{,}751\) missing outcomes, and \(1{,}285\) comparisons whose two model names refer to the same model after name normalization. 
In VisionArena-Battle, we remove \(5{,}216\) \texttt{tie} and \(5{,}905\) \texttt{both\_bad} records; another \(2{,}762\) decisive records match none of the eight released category indicators and are therefore removed.

\paragraph{Selection of the retained models.}
After removing the non-decisive comparisons, we take as candidates all models that appear in at least one decisive comparison on every objective.
We connect two candidate models when their pair has at least one decisive comparison on every objective, and retain a maximum-cardinality clique. If several maximum cliques exist, we choose the one with the largest total number of within-clique comparisons across objectives, then the lexicographically smallest sorted list of model names.
The resulting retained sets have \(35\), \(8\), and \(12\) models for Arena-140K, Search-Arena-24K, and VisionArena-Battle, respectively. We call these tasks \emph{Arena-35}, \emph{SearchArena-8}, and \emph{VisionArena-12}; the suffix gives the number of retained models.

Having determined the retained models, we then remove comparisons involving discarded models.
This step removes \(25{,}197\), \(4{,}004\), and \(5{,}467\) objective records from Arena-140K, Search-Arena-24K, and VisionArena-Battle, respectively. 
The details are summarized in \Cref{tab:realistic-data-audit}.
The resulting empirical strict frontiers are presented in~\Cref{tab:realistic-frontier-members}.

\begin{table}[H]
\centering
\caption{Preprocessing counts for the three Arena datasets. ``Assigned comparisons'' counts source comparisons assigned to at least one evaluation objective; ``Objective records'' counts the resulting pair--objective records. As in the main text, ``Rows'' counts pair--objective records involving only retained models. The \(23{,}458\) retained VisionArena-Battle records correspond to \(12{,}937\) unique source comparisons.}
\label{tab:realistic-data-audit}
\footnotesize
\resizebox{\textwidth}{!}{%
\begin{tabular}{@{}lrrrrrrr@{}}
\toprule
Dataset & Source & Decisive & Assigned comparisons & Objective records & Candidates & \(K\) & Rows \\
\midrule
Arena-140K & \(135{,}609\) & \(98{,}332\) & \(98{,}332\) & \(98{,}332\) & 52 & 35 & \(73{,}135\) \\
Search-Arena-24K & \(24{,}069\)  & \(8{,}309\)  & \(8{,}309\) & \(8{,}309\)  & 13 &  8 & \(4{,}305\) \\
VisionArena-Battle & \(29{,}849\)  & \(18{,}728\) & \(15{,}966\) & \(28{,}925\) & 17 & 12 & \(23{,}458\) \\
\bottomrule
\end{tabular}%
}
\end{table}

\begin{table}[H]
\centering
\caption{Full-dimensional strict empirical Pareto sets on the three retained model sets.}
\label{tab:realistic-frontier-members}
\footnotesize
\begin{tabular}{@{}p{0.23\textwidth}>{\raggedright\arraybackslash}p{0.71\textwidth}@{}}
\toprule
Dataset & Pareto members \\
\midrule
Arena-35 & \texttt{chatgpt-4o-latest-20250326}; \texttt{gemini-2.5-pro}; \texttt{grok-3-preview-02-24} \\
\addlinespace
SearchArena-8 & \texttt{gemini-2.5-pro-exp-03-25-grounding}; \texttt{sonar}; \texttt{sonar-pro-high}; \texttt{sonar-reasoning}; \texttt{sonar-reasoning-pro-high} \\
\addlinespace
VisionArena-12 & \texttt{claude-3-5-sonnet-20240620}; \texttt{gemini-1.5-pro-exp-0801}; \texttt{gpt-4o-2024-05-13} \\
\bottomrule
\end{tabular}
\end{table}

\subsection{Sampling from logged Arena comparisons}
\label{sec:arena-replay}
We evaluate the four methods by sampling {\em with replacement} from the retained logs,
where each draw counts as one comparison.
We evaluate the methods by treating the logged comparisons as the ground truth and comparing the methods' outputs against the empirical Pareto frontier derived from these logs.

\paragraph{Ground truth and inferential target.}
Let $\mathcal D_{\log}$ denote the fixed log of decisive comparisons. 
For any $1\le i < j\le K$ and $r\in[d]$, let  $N_{ij,r}^{\log}$ be the number of records for that pair and objective, 
where $N_{ij,r}^{\log}\ge 1$ by the choice of retained models.
These records form a \emph{pair--objective cell}, defined as 
\begin{equation}
    \mathcal D_{ij,r}:=(Y_{ij,r,s})_{s=1}^{N_{ij,r}^{\log}},
    \label{eq:arena-cell}
\end{equation}
where $Y_{ij,r,s}=1$ if $i$ wins and $0$ if $j$ wins. For the reverse orientation, set $Y_{ji,r,s}=1-Y_{ij,r,s}$ and $N_{ji,r}^{\log}=N_{ij,r}^{\log}$. 
Define
\begin{equation}
    \widehat p_{ij,r}:=\frac{1}{N_{ij,r}^{\log}}
    \sum_{s=1}^{N_{ij,r}^{\log}}Y_{ij,r,s},
    \qquad \widehat p_{ji,r}=1-\widehat p_{ij,r},
    \label{eq:empirical-arena-win-rate}
\end{equation}
and the empirical uniform-opponent Borda coordinates
\begin{equation}
    b_{i,r}^{\mathrm{emp}}:=\frac{1}{K-1}\sum_{j\ne i}\widehat p_{ij,r}.
    \label{eq:empirical-arena-borda-coordinate}
\end{equation}
Writing $b_i^{\mathrm{emp}}=(b_{i,1}^{\mathrm{emp}},\ldots,b_{i,d}^{\mathrm{emp}})$ and $b^{\mathrm{emp}}=(b_i^{\mathrm{emp}})_{i=1}^K$, the ground-truth Pareto set is
\begin{equation}
    \mathcal P_{\mathrm{emp}}:=\mathcal P(b^{\mathrm{emp}}).
    \label{eq:empirical-arena-pareto-target}
\end{equation}

\paragraph{Uniform-opponent queries.}
For a focal query $(i,r)$, \VBEGE\ draws an opponent from all other retained models, including models already removed from its active set, then draws a record from that cell:
\begin{equation}
    J\sim\operatorname{Unif}([K]\setminus\{i\}),
    \qquad S\mid J=j\sim\operatorname{Unif}(\{1,\ldots,N_{ij,r}^{\log}\}).
    \label{eq:arena-replay-record}
\end{equation}
It returns
\begin{equation}
    Z:=Y_{iJ,r,S}.
    \label{eq:arena-replay-outcome}
\end{equation}
Both samling steps are performed with replacement, using 
fresh randomness independent of any other draw and of the algorithm's internal randomization.
The following lemma establishes the the sampled $Z$ is conditionally unbiased for the empirical Borda coordinate,
where  the subscript $\mathrm{samp}$ denotes this randomness conditional on $\mathcal D_{\log}$.

\begin{lemma}[Conditional Bernoulli representation]
\label{lem:arena-replay-validity}
For every fixed logged data $\mathcal D_{\log}$ and 
focal query $(i,r)$,
\begin{equation}
    Z\mid\mathcal D_{\log},i,r
    \sim\operatorname{Bernoulli}(b_{i,r}^{\mathrm{emp}}).
    \label{eq:arena-replay-bernoulli}
\end{equation}
The sampling randomization can be represented by mutually independent streams, one for each model--objective pair, with
\begin{equation}
    Z_{i,r,1},Z_{i,r,2},\ldots\mid\mathcal D_{\log}
    \overset{\mathrm{iid}}{\sim}\operatorname{Bernoulli}(b_{i,r}^{\mathrm{emp}}).
    \label{eq:arena-replay-iid-stream}
\end{equation}
\end{lemma}
\begin{proof}
Conditional on the log and $J=j$, the mean of $Z$ is $\widehat p_{ij,r}$. Averaging over the uniform opponent gives
\[
    \mathbb E_{\mathrm{samp}}[Z\mid\mathcal D_{\log},i,r]
    =\frac{1}{K-1}\sum_{j\ne i}\widehat p_{ij,r}
    =b_{i,r}^{\mathrm{emp}}.
\]
Since $Z$ is binary, this determines its Bernoulli law. Fresh, independent opponent and record draws give the independent streams. Adaptive selection of $(i,r)$ reveals the next unused observation of the corresponding stream, so this representation also applies to adaptive queries.
\end{proof}

For fixed $n\geq1$, let $\overline Z_{i,r}(n):=n^{-1}\sum_{s=1}^n Z_{i,r,s}$; the phase-$m$ estimate is $\widehat b_{i,r}^{(m)}=\overline Z_{i,r}(n_m)$.

\begin{remark}[Recovery conditional on the logged data]
\label{rem:arena-replay-guarantee}
If $\min_i\Delta_i(b^{\mathrm{emp}})>0$, \Cref{lem:arena-replay-validity} lets us apply \Cref{thm:borda-upper} with $b=b^{\mathrm{emp}}$: with probability at least $1-\delta$ conditional on $\mathcal D_{\log}$, \VBEGE\ with $C=(2,4,4)$ recovers $\mathcal P_{\mathrm{emp}}$ and satisfies the same sample-complexity bound with gaps $\Delta_i(b^{\mathrm{emp}})$.
\end{remark}

\paragraph{Opponent weighting and baseline sampling.}
For \VBEGE\ and Focal Borda, each focal query $(i,r)$ 
selects an opponent uniformly from the other retained models, then samples a logged outcome uniformly with replacement from the corresponding pair--objective cell. 
Focal Borda allocates these queries uniformly over model--objective pairs. BT-MLE and the Borda plug-in allocate comparisons uniformly over unordered pair--objective cells, sampling a logged outcome uniformly with replacement from each selected cell. BT-MLE fits scores to the resulting pair counts, while the plug-in averages the estimated pairwise win rates over opponents.

For efficiency, the implementations sample batch counts directly. A batch of $n$ focal queries to $(i,r)$ has a $\operatorname{Binomial}(n,b_{i,r}^{\mathrm{emp}})$ win count, while a batch in cell $(i,j,r)$ has a $\operatorname{Binomial}(n,\widehat p_{ij,r})$ win count. These draws preserve the joint law of the counts used by the methods, which make their decisions at batch boundaries.

\paragraph{Population-level unbiasedness under an additional logging assumption.}
The recovery guarantee above conditions on the observed log. To relate its Borda means to population means, write $\mathbf N=(N_{ij,r}^{\log})_{i<j,r}$ and let $\mathbb E_{\log}$ denote expectation over the logged outcomes conditional on these counts.

\begin{lemma}[Population Borda unbiasedness under cell-wise iid logging]
\label{lem:arena-population-borda-unbiased}
Suppose that, for every cell,
\begin{equation}
    Y_{ij,r,1},\ldots,Y_{ij,r,N_{ij,r}^{\log}}\mid\mathbf N
    \overset{\mathrm{iid}}{\sim}\operatorname{Bernoulli}(p_{ij,r}),
    \label{eq:arena-cellwise-iid-logging}
\end{equation}
where $p_{ij,r}$ is the population win probability conditional on a decisive outcome. 
Set $p_{ji,r}=1-p_{ij,r}$ and
\begin{equation}
    b_{i,r}^{\mathrm{pop}}:=\frac{1}{K-1}\sum_{j\ne i}p_{ij,r}.
    \label{eq:population-arena-borda-coordinate}
\end{equation}
Then
\[
    \mathbb E_{\log}[\widehat p_{ij,r}\mid\mathbf N]=p_{ij,r},
    \qquad
    \mathbb E_{\log}[b_{i,r}^{\mathrm{emp}}\mid\mathbf N]=b_{i,r}^{\mathrm{pop}}.
\]
For every fixed $n\geq1$, also $\mathbb E_{\log,\mathrm{samp}}[\overline Z_{i,r}(n)\mid\mathbf N]=b_{i,r}^{\mathrm{pop}}$.
\end{lemma}
\begin{proof}
The cell average has expectation $p_{ij,r}$ under \eqref{eq:arena-cellwise-iid-logging}. Averaging these expectations over opponents proves the Borda identity. 

The last assertion follows by iterated expectation and $\mathbb E_{\mathrm{samp}}[\overline Z_{i,r}(n)\mid\mathcal D_{\log}]=b_{i,r}^{\mathrm{emp}}$.
\end{proof}
Together with \Cref{rem:arena-replay-guarantee}, this lemma connects the two levels of the logged-data experiment. Conditional on the log, \VBEGE\ recovers $\mathcal P_{\mathrm{emp}}$ with probability at least $1-\delta$. 
Under cell-wise iid logging, $b^{\mathrm{emp}}$ is an unbiased estimate of $b^{\mathrm{pop}}$ and concentrates around it as the cell counts grow. 

Although unbiasedness does not necessarily transfer to $\mathcal P_{\mathrm{emp}}$
because $\mathcal P(\cdot)$ is discontinuous,
we nevertheless have $\mathcal P_{\mathrm{emp}}=\mathcal P(b^{\mathrm{pop}})$
if $\min_i\Delta_i(b^{\mathrm{pop}})>0$ and  
$\max_{i,r}|b^{\mathrm{emp}}_{i,r}-b^{\mathrm{pop}}_{i,r}|<\min_i\Delta_i(b^{\mathrm{pop}})/2$; 
the probability of the latter event tends to one as $\min_{i<j,r}N_{ij,r}^{\log}\to\infty$.

\subsection{Full Pareto-set results and method comparisons}
\label{app:realistic-pareto-details}

\subsubsection{Comparison of the four methods on the three Arena datasets}

\Cref{tab:realistic-main-benchmarks} gives the complete stopping-time comparison
in the three real-world Arena datasets.

\begin{table}[H]
\centering
\caption{Comparison of the four methods on the three Arena datasets. Entries are mean stopping times in millions of sampled comparisons (including repeated records), with each method's mean divided by the VB-EGE mean in parentheses. Every method stops and recovers the exact frontier in all 100 trials on each dataset.}
\label{tab:realistic-main-benchmarks}
\footnotesize
\setlength{\tabcolsep}{4pt}
\begin{tabular*}{\textwidth}{@{\extracolsep{\fill}}lrrr@{}}
\toprule
Method & Arena-35 & SearchArena-8 & VisionArena-12 \\
\midrule
\VBEGE & \(86.82\;(1.00\times)\) & \(124.34\;(1.00\times)\) & \(8.80\;(1.00\times)\) \\
Focal Borda & \(1{,}435.70\;(16.54\times)\) & \(492.74\;(3.96\times)\) & \(40.39\;(4.59\times)\) \\
BT-MLE & \(1{,}586.26\;(18.27\times)\) & \(149.06\;(1.20\times)\) & \(14.77\;(1.68\times)\) \\
Borda plug-in & \(26{,}164.10\;(301.36\times)\) & \(1{,}890.10\;(15.20\times)\) & \(251.62\;(28.60\times)\) \\
\bottomrule
\end{tabular*}
\end{table}

\subsubsection{Full-dimensional empirical Borda profiles}

\begin{figure}[H]
\centering
\includegraphics[width=0.85\textwidth,keepaspectratio]
{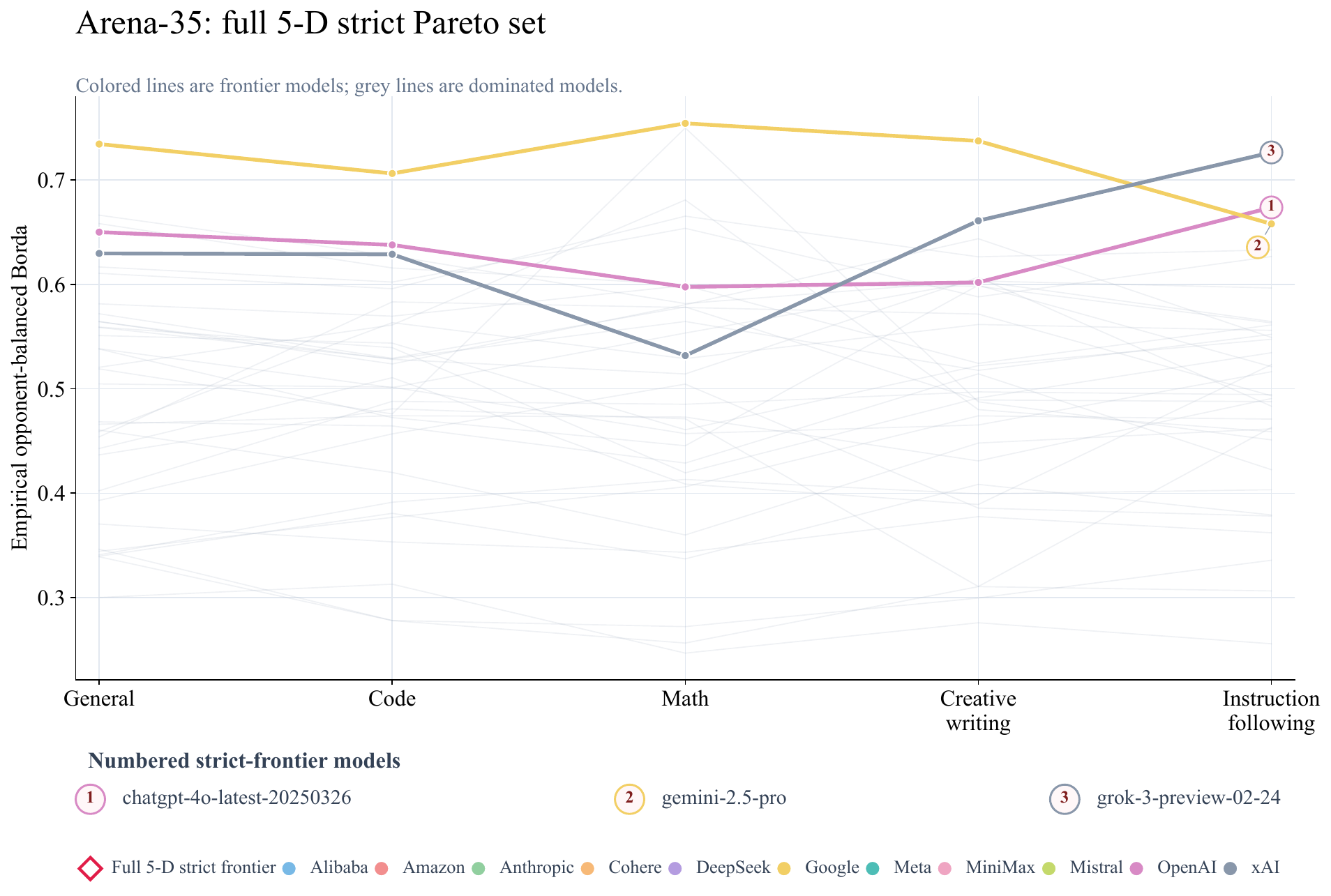}
\caption{Arena-35: empirical Borda profiles on all five objectives.}
\label{fig:realistic-arena35-pareto-full}
\end{figure}

\begin{figure}[H]
\centering
\includegraphics[width=0.85\textwidth,keepaspectratio]
{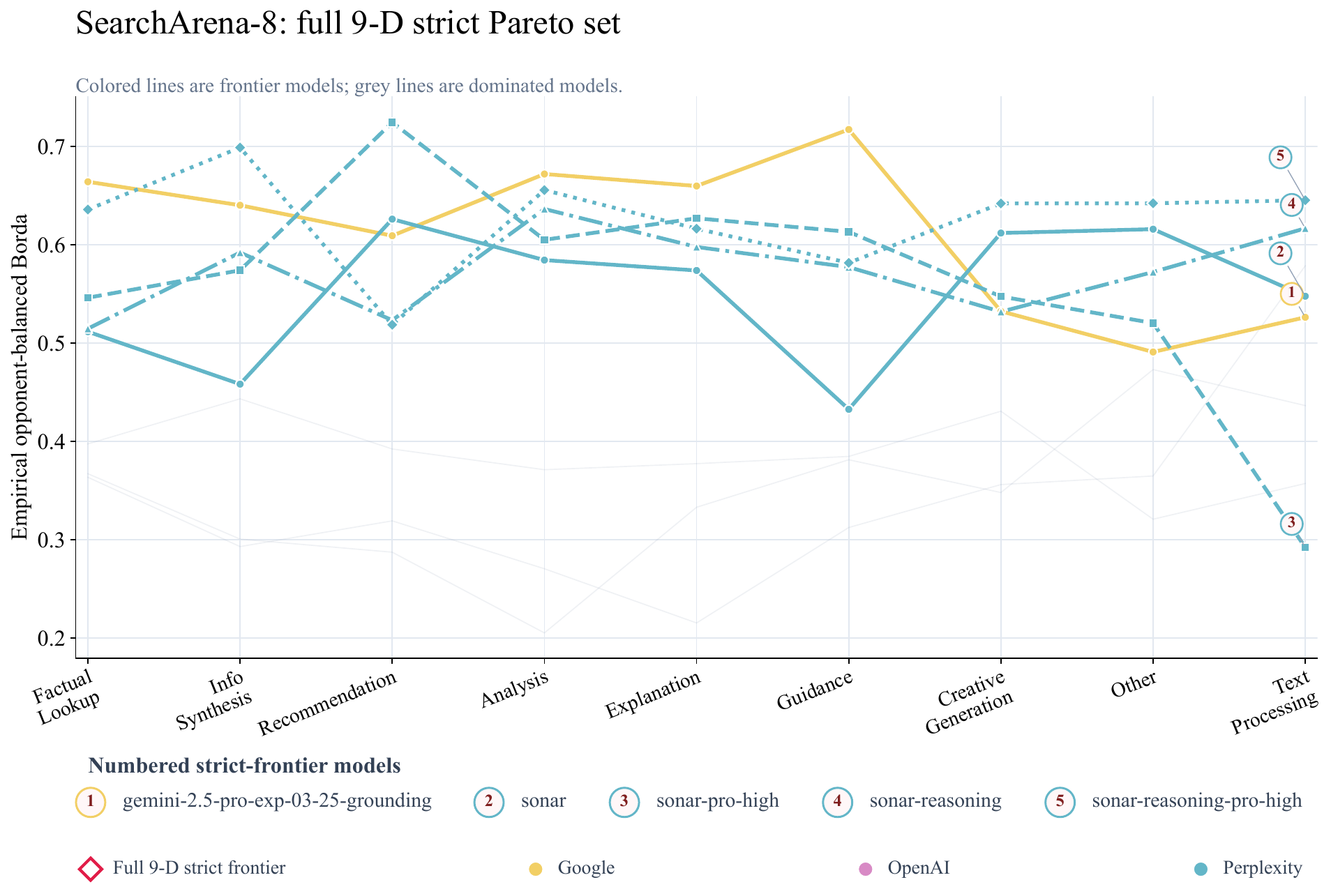}
\caption{SearchArena-8: empirical Borda profiles on all nine objectives.}
\label{fig:realistic-search8-pareto-full}
\end{figure}

\begin{figure}[H]
\centering
\includegraphics[width=0.8\textwidth,keepaspectratio]
{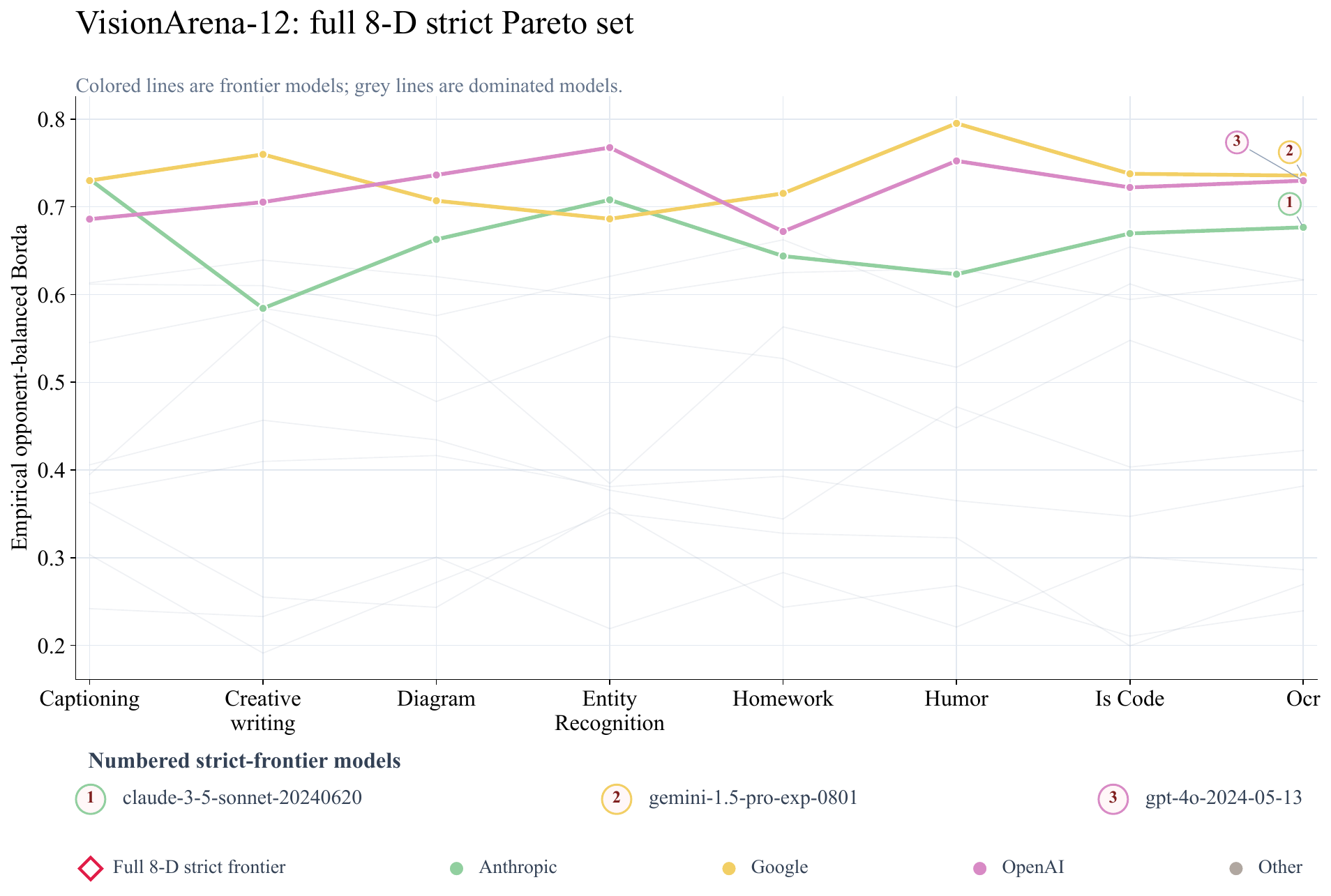}
\caption{VisionArena-12: empirical Borda profiles on all eight objectives.}
\label{fig:realistic-vision12-pareto-full}
\end{figure}

\Cref{fig:realistic-arena35-pareto-full,fig:realistic-search8-pareto-full,fig:realistic-vision12-pareto-full} present 
the Borda scores for all the models under evaluation. Pareto membership is computed using all five, nine, or eight objectives for Arena-35, SearchArena-8, and VisionArena-12, respectively. Colored lines correspond to strict-frontier members 
and faint gray lines are dominated models from the same retained set.  
Across all three datasets, the frontier profiles cross one another repeatedly, reflecting the complementary strengths of the Pareto models.

\subsubsection{Low-dimensional illustrative projections}
\label{sec:low-dimensional-projections}

\begin{figure}[H]
    \centering
    \captionsetup{font=large}
    \captionsetup[subfigure]{font=normalsize}
    \begin{subfigure}[t]{0.48\textwidth}
        \vspace{0pt}
        \centering
        \includegraphics[
            width=\linewidth,
            keepaspectratio
        ]{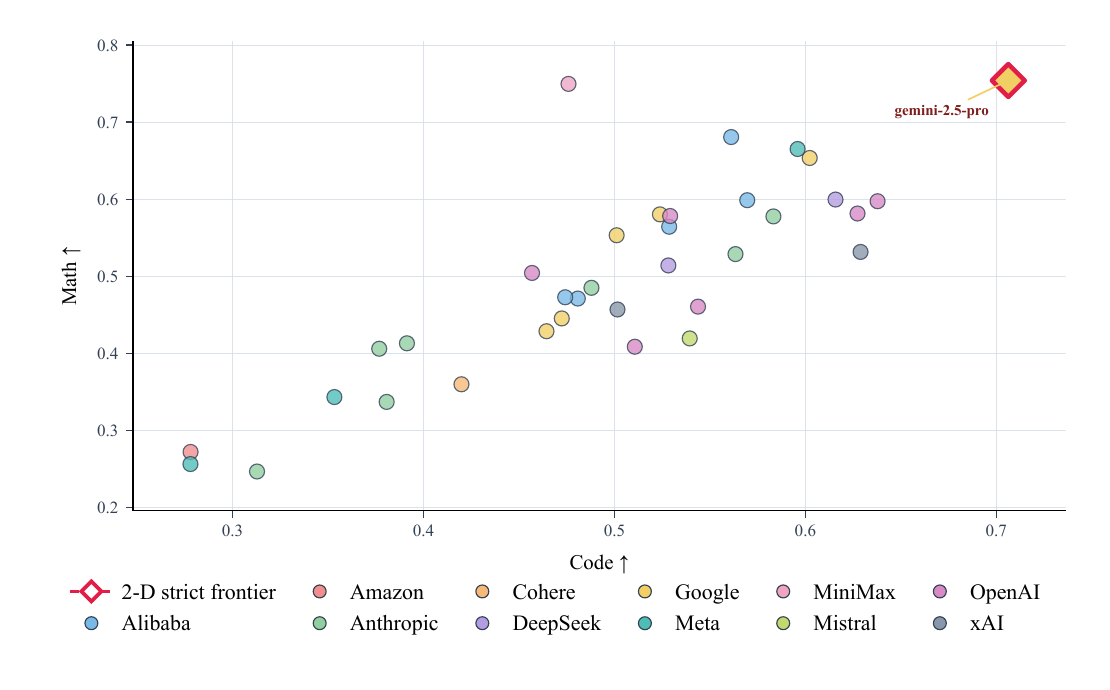}
        \caption{2D: Code versus Math.}
        \label{fig:realistic-arena35-pareto-2d}
    \end{subfigure}
    \hfill
    \begin{subfigure}[t]{0.48\textwidth}
        \vspace{0pt}
        \centering
        \includegraphics[
            width=\linewidth,
            keepaspectratio
        ]{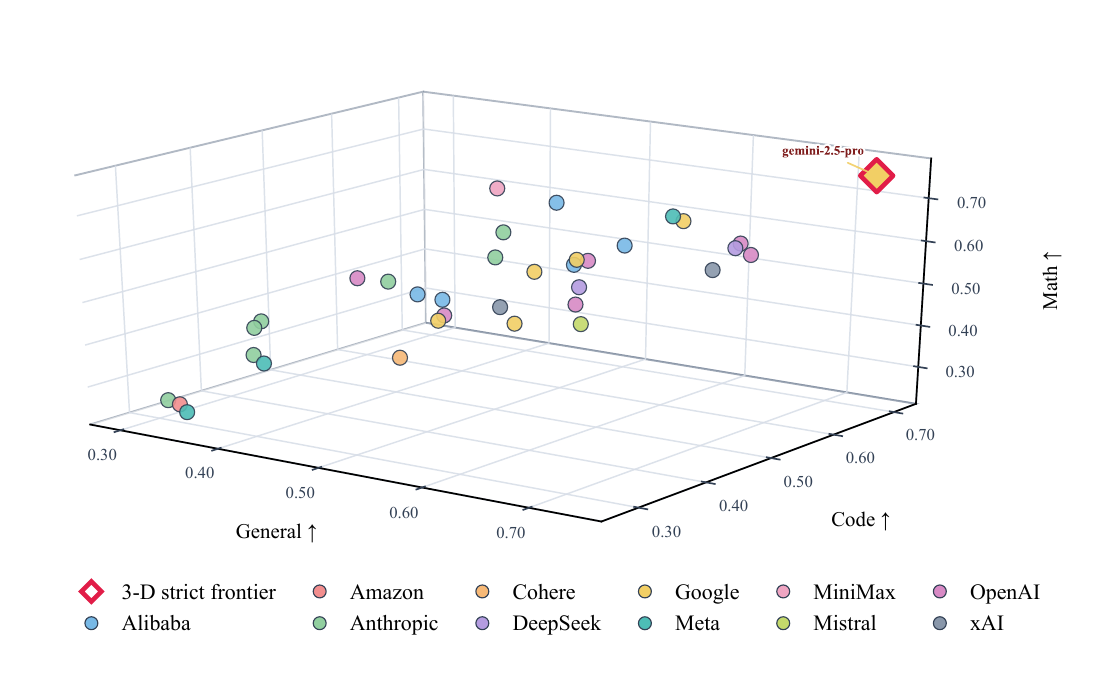}
        \caption{3D: General, Code, and Math.}
        \label{fig:realistic-arena35-pareto-3d}
    \end{subfigure}
    \caption{Arena-35 illustrative projections.}
    \label{fig:realistic-arena35-pareto-projections}
\end{figure}

\begin{figure}[H]
    \centering
    \captionsetup{font=large}
    \captionsetup[subfigure]{font=normalsize}
    \begin{subfigure}[t]{0.48\textwidth}
        \vspace{0pt}
        \centering
        \includegraphics[
            width=\linewidth,
            keepaspectratio
        ]{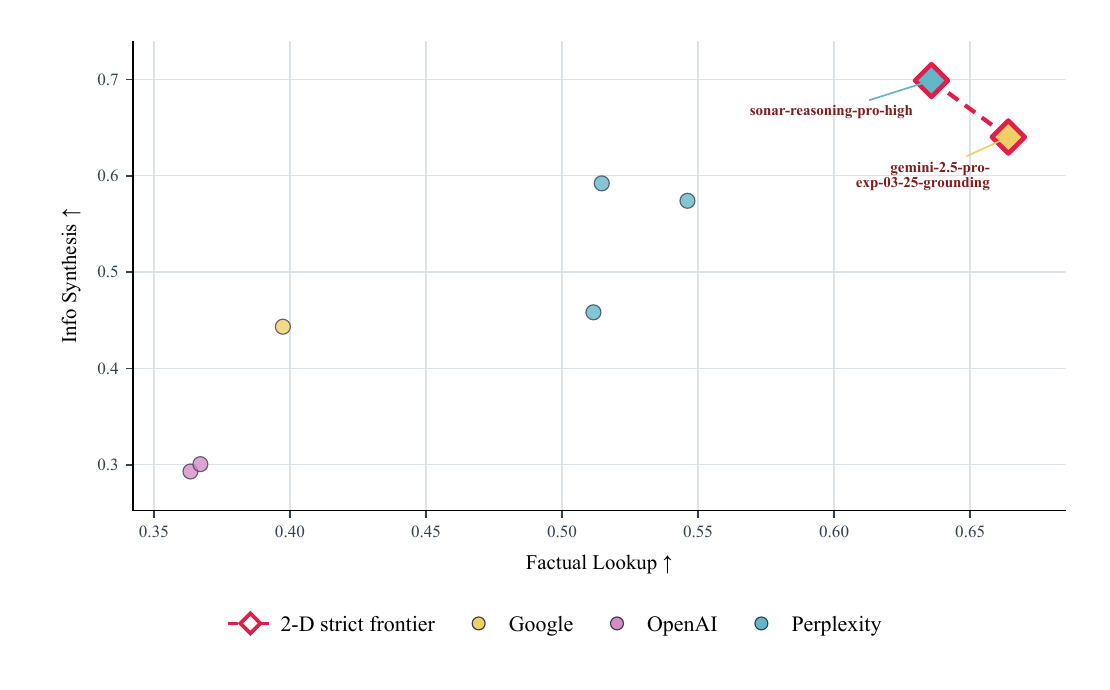}
        \caption{2D: Factual Lookup versus Info Synthesis.}
        \label{fig:realistic-search8-pareto-2d}
    \end{subfigure}
    \hfill
    \begin{subfigure}[t]{0.48\textwidth}
        \vspace{0pt}
        \centering
        \includegraphics[
            width=\linewidth,
            keepaspectratio
        ]{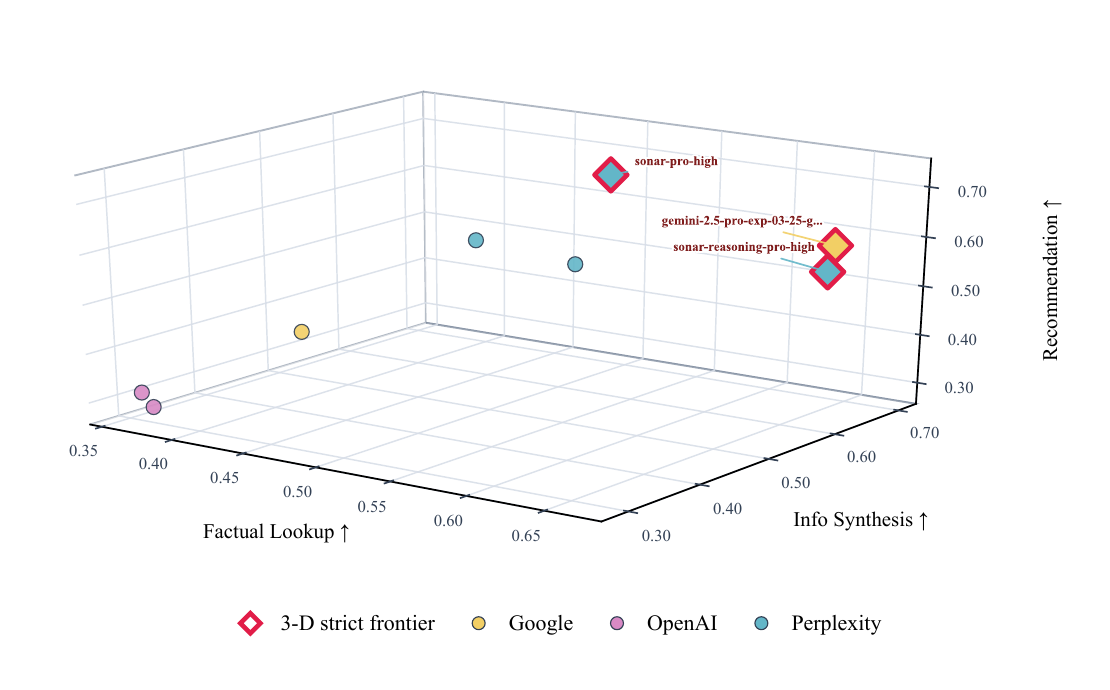}
        \caption{3D: Factual Lookup, Info Synthesis, and Recommendation.}
        \label{fig:realistic-search8-pareto-3d}
    \end{subfigure}
    \caption{SearchArena-8 illustrative projections.}
    \label{fig:realistic-search8-pareto-projections}
\end{figure}

\begin{figure}[H]
    \centering
    \captionsetup{font=large}
    \captionsetup[subfigure]{font=normalsize}
    \begin{subfigure}[t]{0.48\textwidth}
        \vspace{0pt}
        \centering
        \includegraphics[
            width=\linewidth,
            keepaspectratio
        ]{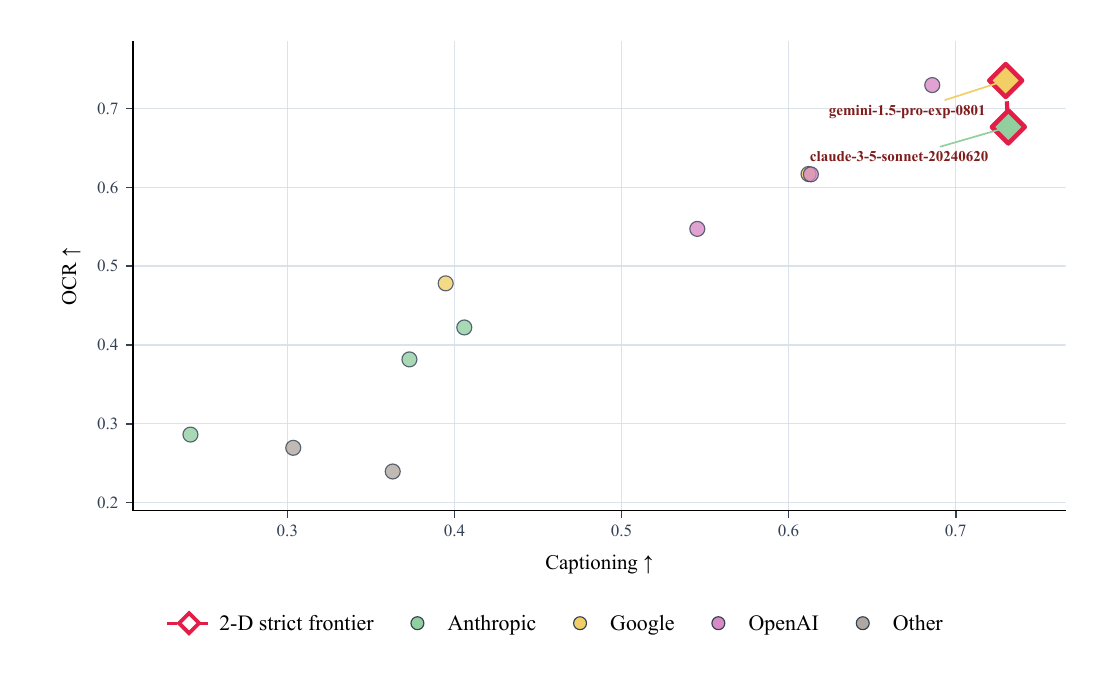}
        \caption{2D: Captioning versus OCR.}
        \label{fig:realistic-vision12-pareto-2d}
    \end{subfigure}
    \hfill
    \begin{subfigure}[t]{0.48\textwidth}
        \vspace{0pt}
        \centering
        \includegraphics[
            width=\linewidth,
            keepaspectratio
        ]{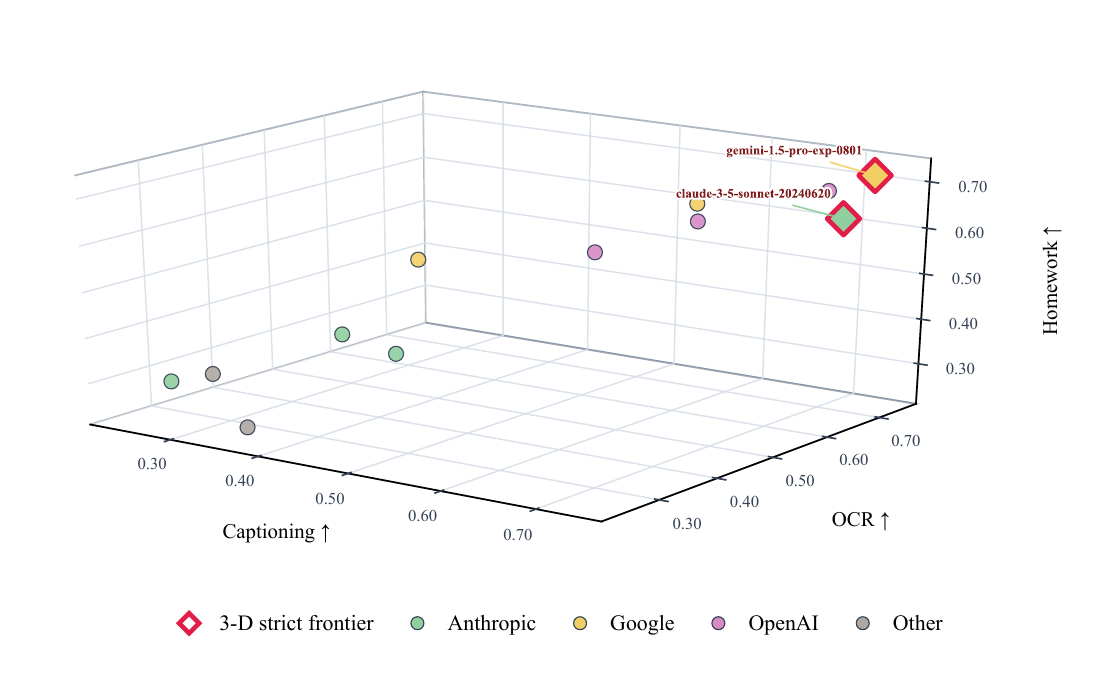}
        \caption{3D: Captioning, OCR, and Homework.}
        \label{fig:realistic-vision12-pareto-3d}
    \end{subfigure}
    \caption{VisionArena-12 illustrative projections.}
    \label{fig:realistic-vision12-pareto-projections}
\end{figure}

\Cref{fig:realistic-arena35-pareto-projections,fig:realistic-search8-pareto-projections,fig:realistic-vision12-pareto-projections} 
present representative two-dimensional and three-dimensional projections for each dataset, where we 
recompute Pareto membership in each projected space. Because strict dominance in the sense of \Cref{def:strict-pareto-frontier} is preserved when coordinates are removed, each projected frontier is a subset of the full-dimensional frontier.

The Arena-35 views here and in \Cref{fig:arena-pareto-frontiers} use the same 35 retained models and the same empirical Borda coordinates. The General, Code, and Math projection retains only \texttt{gemini-2.5-pro}. The Code, Math, and Instruction-Following projection in \Cref{fig:arena-pareto-frontiers} retains all three members of the full frontier: \texttt{chatgpt-4o-\allowbreak latest-\allowbreak 20250326}, \texttt{gemini-2.5-pro}, and \texttt{grok-3-preview-02-24}. All recovery results use the full-dimensional target.

\subsection{Sensitivity to \texttt{tie} and \texttt{both\_bad} handling}
\label{sec:arena-robustness}

The main experiments retain only decisive comparisons. 
This section examines the sensitivity of our results 
to different treatments of non-decisive outcomes. 
We consider two alternative treatmens in addition to 
removing the non-decisive items:
\[
    D=\text{decisive-only},\qquad
    H=\text{direct merged half-credit},\qquad
    F=\text{fitted Davidson half-credit}.
\]
All three use the same retained models, objective assignments, and uniform-opponent aggregation. 

\paragraph{Common setup and comparison metrics.}
For an unordered pair \(\{i,j\}\) and objective \(r\), let \(W_{ij,r}\) and \(W_{ji,r}\) be the directed win counts, and \(T_{ij,r}\) and \(N^{\mathrm{bad}}_{ij,r}\) the \texttt{tie} and \texttt{both\_bad} counts. We merge non-decisive outcomes into \(U_{ij,r}:=T_{ij,r}+N^{\mathrm{bad}}_{ij,r}\). \Cref{tab:realistic-fixed-roster-outcomes} reports their totals; non-decisive records account for \(27.5\%\) to \(38.0\%\) of each retained log.

For \(X\in\{D,H,F\}\), write \(b^X=(b_i^X)_{i=1}^K\) for its Borda vectors and define
\[
    b^D=b^{\mathrm{emp}},\qquad
    \mathcal P_X=\mathcal P(b^X),\qquad
    \Delta_{\min}^X=\min_i\Delta_i(b^X),\qquad
    \mathcal H_X=\sum_i\Delta_i(b^X)^{-2},
\]
where the definition of $b^{\mathrm{emp}}$ is to be 
given below for each treatment.

For two methods $X$ and $Y$, we define 
\[
    \operatorname{Jac}(X,Y)=\frac{|\mathcal P_X\cap\mathcal P_Y|}{|\mathcal P_X\cup\mathcal P_Y|},
\]
and let \(\rho_{\min}(X,Y)\) denote the minimum Spearman correlation across objectives. Stopping-time comparisons use the main experiment settings: 100 independent trials per construction and dataset, \(\delta=0.05\), and \(C=(2,4,4)\).

\begin{table}[H]
\centering
\caption{
    Outcome counts on the retained models, before non-decisive labels are merged. Each record is one comparison assigned to one objective, and the last column is the fraction of records labeled \texttt{tie} or \texttt{both\_bad}.
}
\label{tab:realistic-fixed-roster-outcomes}
\footnotesize
\begin{tabular}{@{}lrrrr@{}}
\toprule
Dataset & Decisive & \texttt{tie} & \texttt{both\_bad}
        & Non-decisive fraction \\
\midrule
Arena-35       & \(73{,}135\) & \(16{,}093\) & \(11{,}594\) & \(27.5\%\) \\
SearchArena-8  & \(4{,}305\)  & \(1{,}148\)  & \(676\)      & \(29.8\%\) \\
VisionArena-12 & \(23{,}458\) & \(6{,}346\)  & \(8{,}031\)  & \(38.0\%\) \\
\bottomrule
\end{tabular}
\end{table}

\subsubsection{Direct merged half-credit recoding}
\label{app:real-data-direct-tie-sensitivity}

\begin{figure}[H]
\centering
\includegraphics[width=0.98\textwidth,keepaspectratio]
{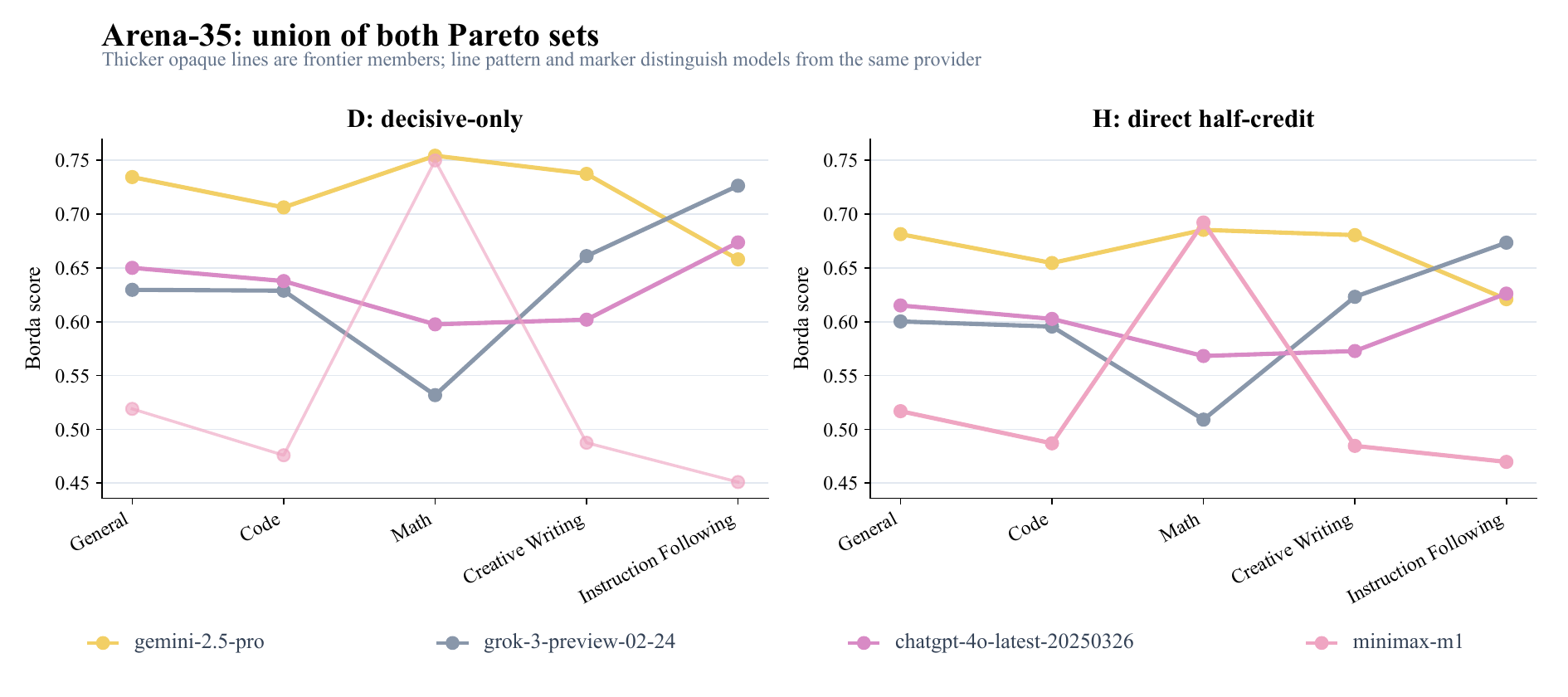}
\caption{Arena-35: Borda profiles of the models in \(\mathcal P_D\cup\mathcal P_H\) under each target; \texttt{minimax-m1} enters the strict Pareto frontier under \(H\). Left: decisive-only target \(D\); right: direct half-credit target \(H\).}
\label{fig:realistic-direct-arena35-sensitivity}
\end{figure}

\begin{figure}[H]
\centering
\includegraphics[width=0.98\textwidth,keepaspectratio]
{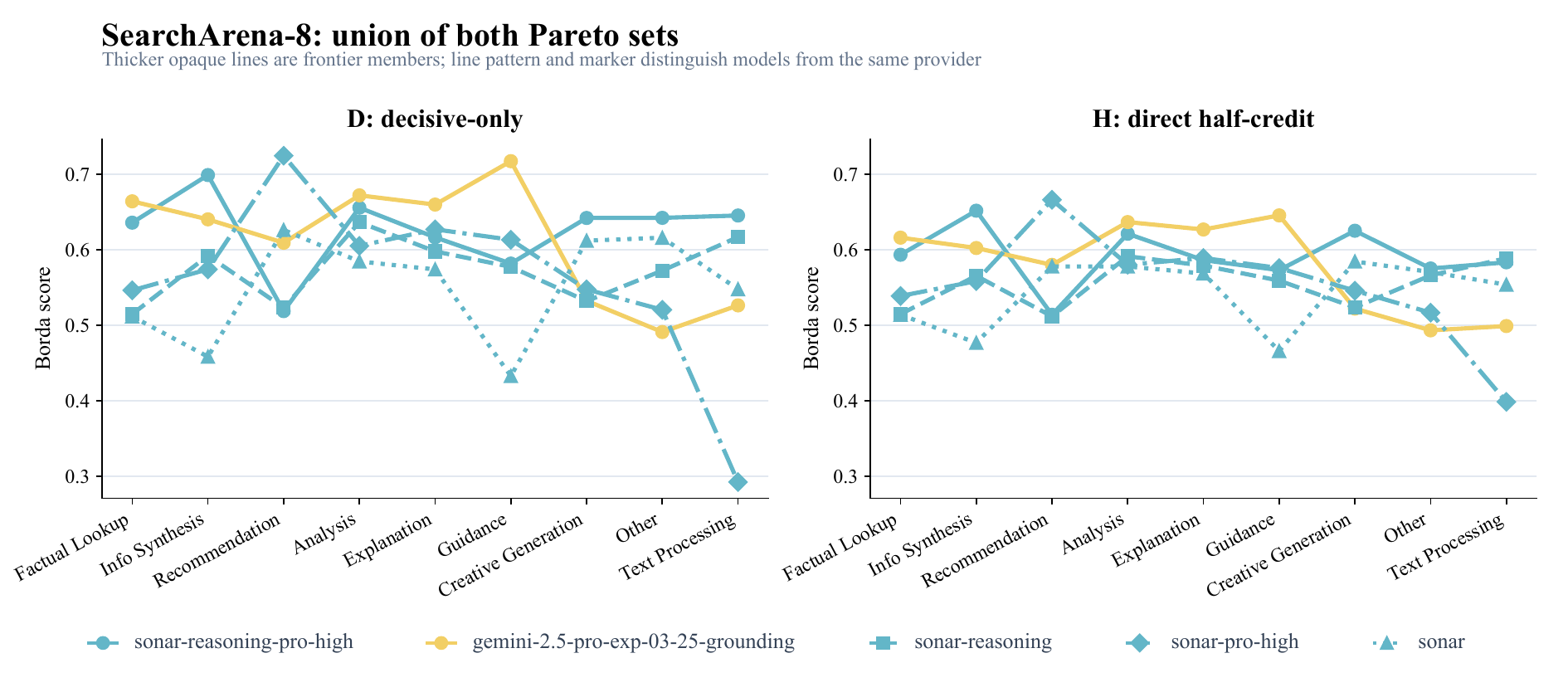}
\caption{SearchArena-8: Borda profiles of the models in \(\mathcal P_D\cup\mathcal P_H\) under each target; frontier membership is unchanged. Left: decisive-only target \(D\); right: direct half-credit target \(H\).}
\label{fig:realistic-direct-search8-sensitivity}
\end{figure}

\begin{figure}[H]
\centering
\includegraphics[width=0.98\textwidth,keepaspectratio]
{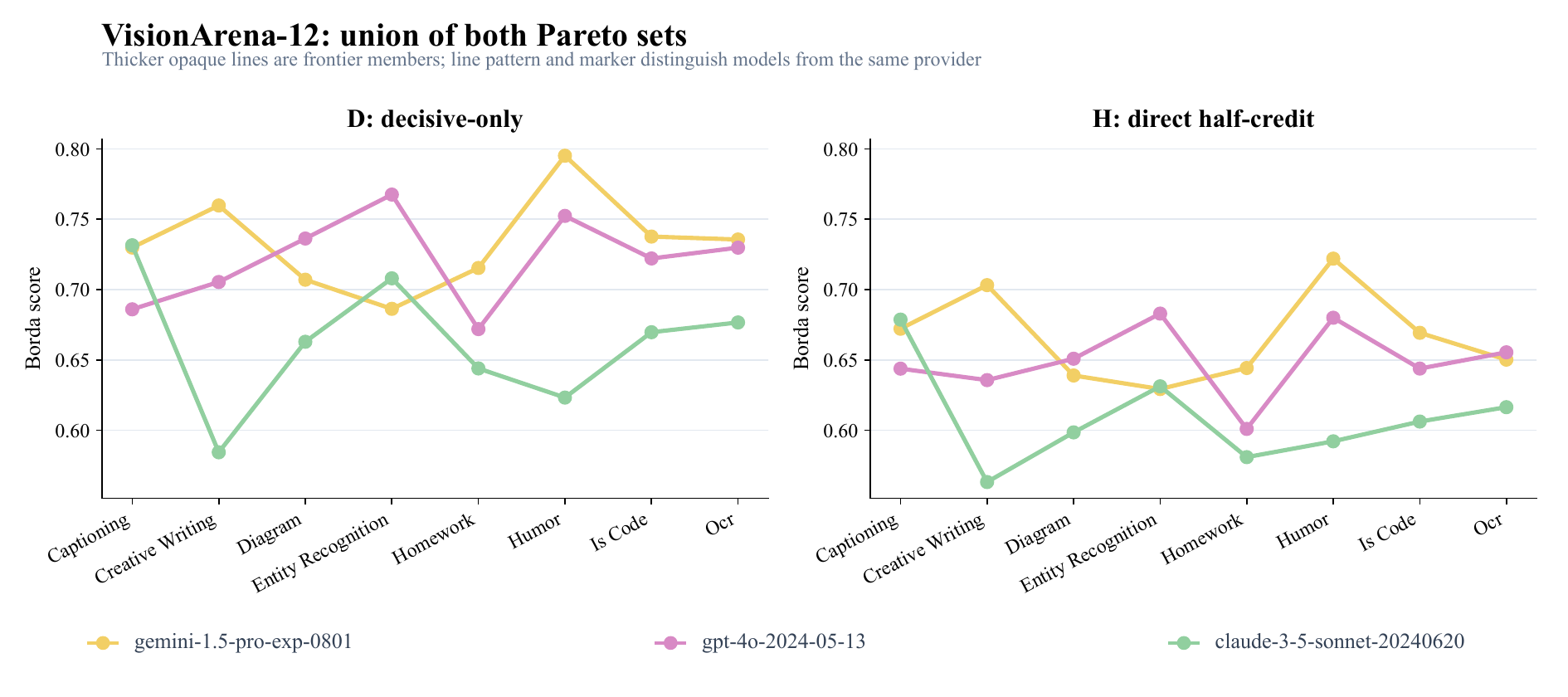}
\caption{VisionArena-12: Borda profiles of the models in \(\mathcal P_D\cup\mathcal P_H\) under each target; frontier membership is unchanged. Left: decisive-only target \(D\); right: direct half-credit target \(H\).}
\label{fig:realistic-direct-vision12-sensitivity}
\end{figure}

Under \(D\), the empirical win rate is
\[
    \widehat p^{D}_{ij,r}
    =
    \frac{W_{ij,r}}{W_{ij,r}+W_{ji,r}}
    =\widehat p_{ij,r}.
\]
The direct construction $H$ keeps each decisive outcome coded as \((1,0)\) and assigns \((1/2,1/2)\) to every \texttt{tie} and \texttt{both\_bad} outcome, so its directed cell mean is
\begin{equation}
    \widehat p^{H}_{ij,r}
    =
    \frac{W_{ij,r}+\frac12 U_{ij,r}}
         {W_{ij,r}+W_{ji,r}+U_{ij,r}},
    \qquad
    \widehat p^{H}_{ji,r}=1-\widehat p^{H}_{ij,r}.
    \label{eq:direct-half-credit-cell}
\end{equation}
Every pair--objective cell is estimated directly from its own logged counts, and the corresponding Borda mean is \(b^H_{i,r}=(K-1)^{-1}\sum_{j\ne i}\widehat p^H_{ij,r}\).
\Cref{fig:realistic-direct-arena35-sensitivity,fig:realistic-direct-search8-sensitivity,fig:realistic-direct-vision12-sensitivity} compare \(D\) with \(H\) on Arena-35, SearchArena-8, and VisionArena-12, respectively.

As in Appendix~\ref{sec:arena-replay}, each query draws an opponent uniformly from the other retained models, then a record uniformly with replacement from the selected cell. A win, a non-decisive outcome, and a loss return \(1\), \(1/2\), and \(0\). 
The proof of \Cref{thm:borda-upper} uses Hoeffding's inequality, which also applies to these \([0,1]\)-valued observations, so \Cref{rem:arena-replay-guarantee} extends to \(\mathcal P_H\) when its gaps are positive.

For batching, we average the three outcome probabilities over opponents and draw multinomial category counts, extending the batch-count implementation in Appendix~\ref{sec:arena-replay}. The resulting sums and counts have the same distribution as sequential queries; these are the only observation summaries used by \VBEGE.

\FloatBarrier

\subsubsection{Fitted Davidson construction}
\label{app:real-data-davidson-sensitivity}

\paragraph{Three-outcome likelihood.}
The fitted construction \(F\) uses the full count triple \((W_{ij,r},W_{ji,r},U_{ij,r})\) to fit, separately for each objective \(r\), a Davidson model \citep{davidson1970extending} with pair-specific tie weights, 
following the factored tie construction of \citet{ameli2025statistical}. 
Let \(\theta^{F}_{i,r}\) be the score of model \(i\) and, for \(i<j\), let \(\eta_{ij,r}\) be the symmetric log tie weight of the pair. 
The probabilities that \(i\) wins, that \(j\) wins, and that the comparison is non-decisive are parametrized as
\begin{equation}
    \left(
        \pi^{W}_{ij,r},\,
        \pi^{L}_{ij,r},\,
        \pi^{U}_{ij,r}
    \right)
    =
    \operatorname{softmax}\!\left(
        \frac{\theta^{F}_{i,r}-\theta^{F}_{j,r}}{2},\,
\frac{\theta^{F}_{j,r}-\theta^{F}_{i,r}}{2},\,
        \eta_{ij,r}
    \right),
    \qquad
    \sum_{i=1}^{K}\theta^{F}_{i,r}=0.
    \label{eq:davidson-three-outcome}
\end{equation}
Conditional on a decisive outcome, the win probability follows the Bradley--Terry model with score difference \(\theta^{F}_{i,r}-\theta^{F}_{j,r}\). The parameters are fitted jointly by minimizing the multinomial negative log-likelihood obtained by weighting the three log probabilities in \eqref{eq:davidson-three-outcome} by \((W_{ij,r},W_{ji,r},U_{ij,r})\) and summing over \(i<j\).

\paragraph{Factored tie weight.}
We use a parameter $k_{\mathrm{tie}}$ to control 
the factor size in the parameterization of $\eta_{ij,r}$.  
When \(k_{\rm tie}=0\), all pairs in objective \(r\) share one scalar tie logit, \(\eta_{ij,r}=\gamma^{\mathrm{tie}}_r\).  When \(k_{\rm tie}>0\), we set
\begin{equation}
    \eta_{ij,r}=(\Gamma_r)_{ij},
    \qquad
    \Gamma_r=V_r\Phi^\top+\Phi V_r^\top,
    \qquad
    (\Phi)_{a\ell}
    =
    \sqrt{\frac{2}{K}}
    \cos\!\left[
        \frac{\pi(2a-1)(2\ell-1)}{4K}
    \right],
    \label{eq:davidson-factored-tie}
\end{equation}
where \(V_r,\Phi\in\mathbb R^{K\times k_{\rm tie}}\) and \(\Phi\) consists of the first \(k_{\rm tie}\) columns of the fixed DCT-IV basis. This parameterization makes \(\Gamma_r\) symmetric while allowing the tie propensity to vary across pairs. We call \(k_{\rm tie}\), the number of factor columns, the \emph{tie-factor size}. \citet{ameli2025statistical} call it the rank of the factor matrices; we use a different name because \(\Gamma_r\) itself can have rank up to \(2k_{\rm tie}\). For \(k_{\rm tie}\geq1\), no separate scalar logit \(\gamma^{\mathrm{tie}}_r\) is included. The models with \(k_{\rm tie}\geq1\) are nested, because appending a zero column to \(V_r\) embeds size \(k_{\rm tie}\) in size \(k_{\rm tie}+1\), whereas the constant-logit model with \(k_{\rm tie}=0\) is generally not nested in them. We therefore compare all candidates by held-out likelihood.

\paragraph{Selection of \(k_{\rm tie}\) and refitting.}
Within each objective, we assign each observation independently to training with probability \(0.9\). Whenever a model pair has no training observations in that objective, we move one of its validation observations, taken from its most frequent outcome category, to training.

The candidate grids are prespecified and listed in \Cref{tab:davidson-rank-selection}; Arena-35 uses a broader grid than the two smaller model sets. Selection is restricted to these grids. We select the size with the smallest held-out mean negative log-likelihood and refit it to all observations. Fits use BFGS with analytic gradients; all 76 candidate fits and 22 full-data fits converged, with one candidate requiring an L-BFGS-B fallback.

\begin{table}[H]
\centering
\caption{Selection of the tie-factor size \(k_{\rm tie}\) on the fixed split described above. Selected sizes are listed in the objective order of Appendix~\ref{app:real-data-protocol}.}
\label{tab:davidson-rank-selection}
\footnotesize
\begin{tabular}{@{}p{0.18\textwidth}p{0.22\textwidth}p{0.52\textwidth}@{}}
\toprule
Dataset & Candidate \(k_{\rm tie}\) & Selected \(k_{\rm tie}\) by objective \\
\midrule
Arena-35 & \(\{0,1,2,3,5\}\) & \((2,1,0,0,0)\) \\
SearchArena-8 & \(\{0,1,2\}\) & \((2,0,2,0,0,1,2,1,0)\) \\
VisionArena-12 & \(\{0,1,2\}\) & \((2,2,1,2,1,0,1,2)\) \\
\bottomrule
\end{tabular}
\end{table}

\begin{figure}[H]
\centering
\includegraphics[width=0.9\textwidth,keepaspectratio]
{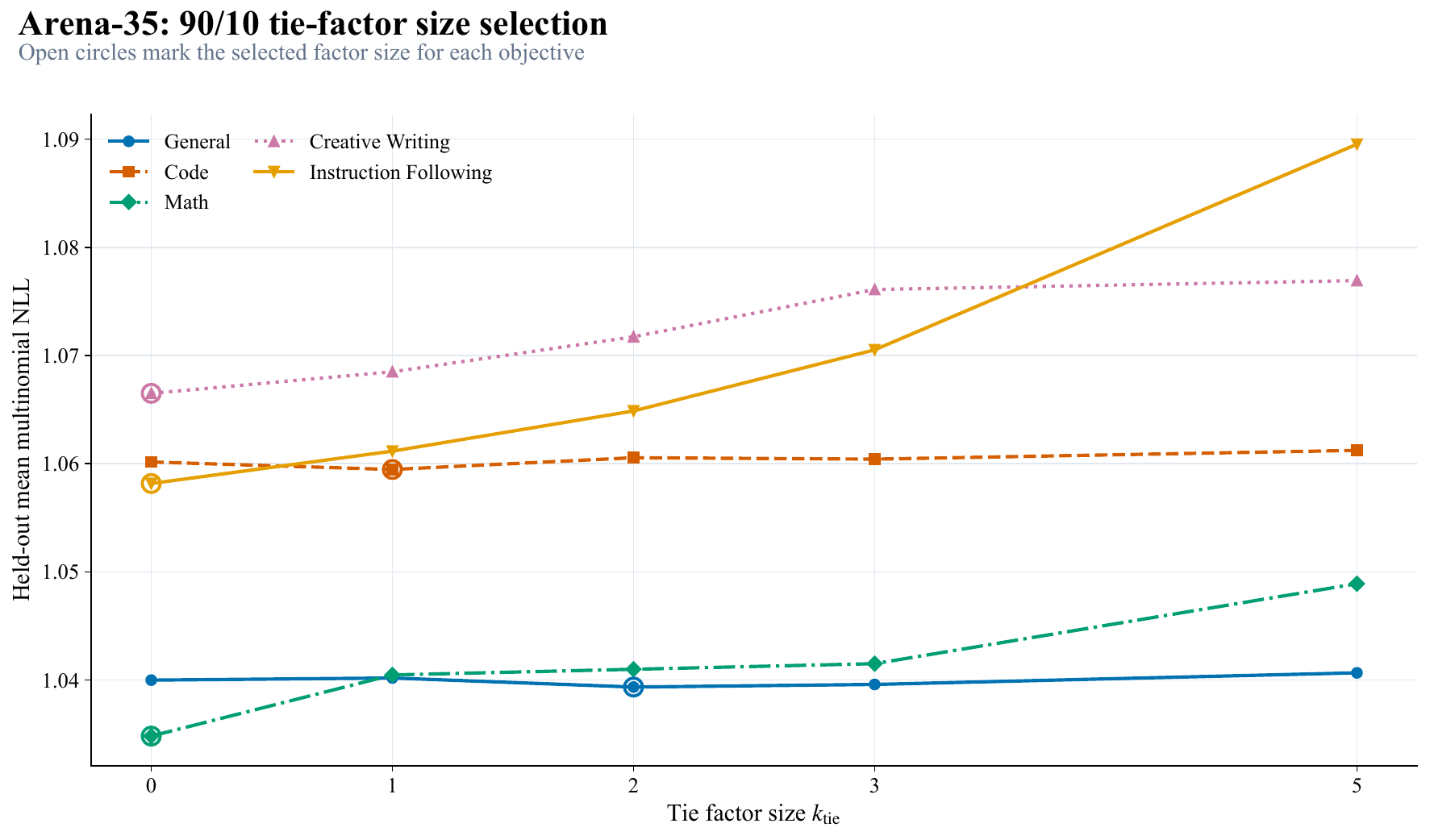}
\caption{Arena-35: held-out NLL for each candidate tie-factor size \(k_{\rm tie}\); open circles mark the selected size for each objective. The candidates \(0,1,2,3,5\) are plotted at their numerical positions.}
\label{fig:realistic-davidson-arena35-model-selection}
\end{figure}

\begin{figure}[H]
\centering
\includegraphics[width=0.9\textwidth,keepaspectratio]
{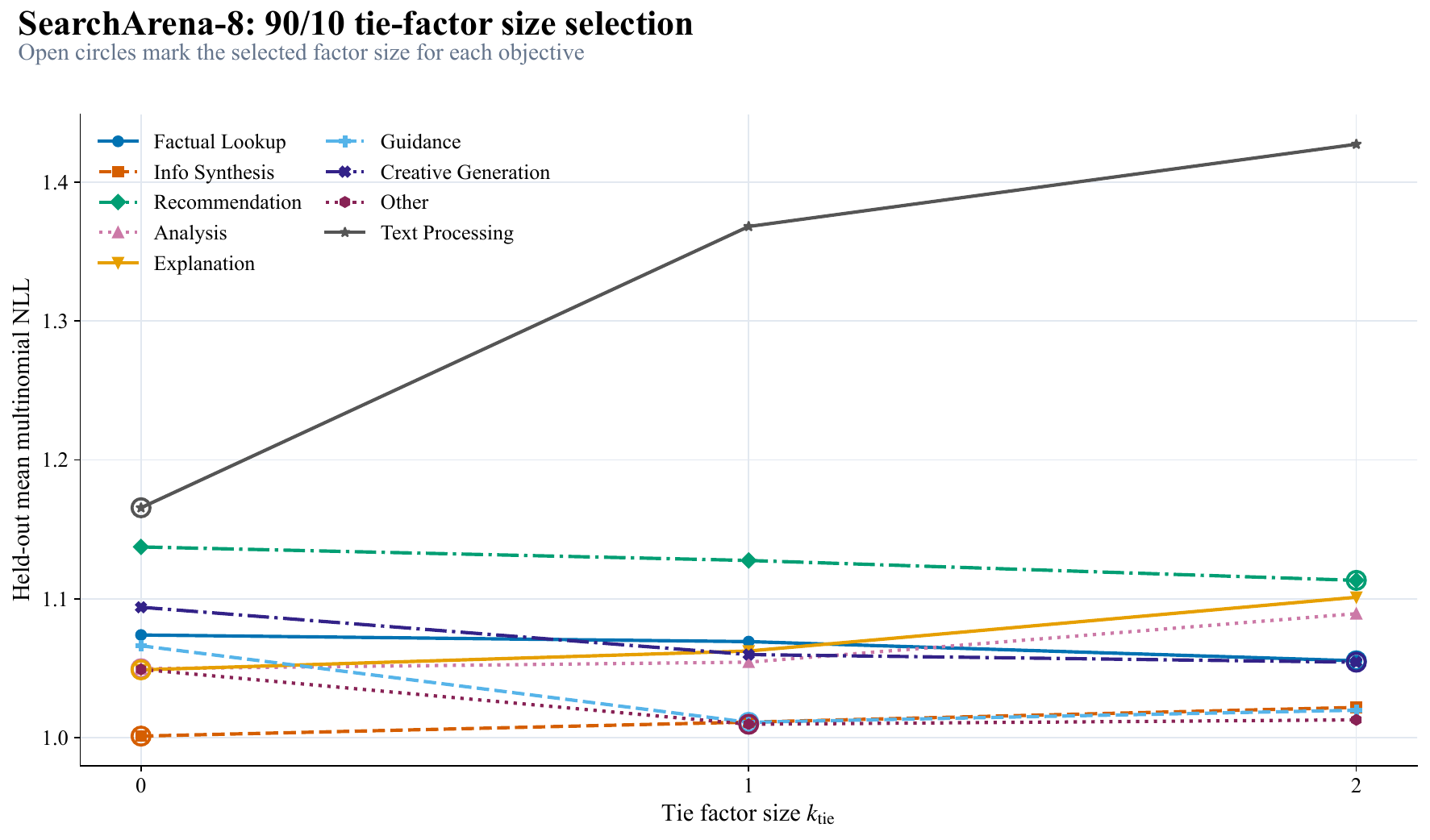}
\caption{SearchArena-8: held-out NLL for each candidate tie-factor size \(k_{\rm tie}\); open circles mark the selected size for each objective.}
\label{fig:realistic-davidson-search8-model-selection}
\end{figure}

\begin{figure}[H]
\centering
\includegraphics[width=0.9\textwidth,keepaspectratio]
{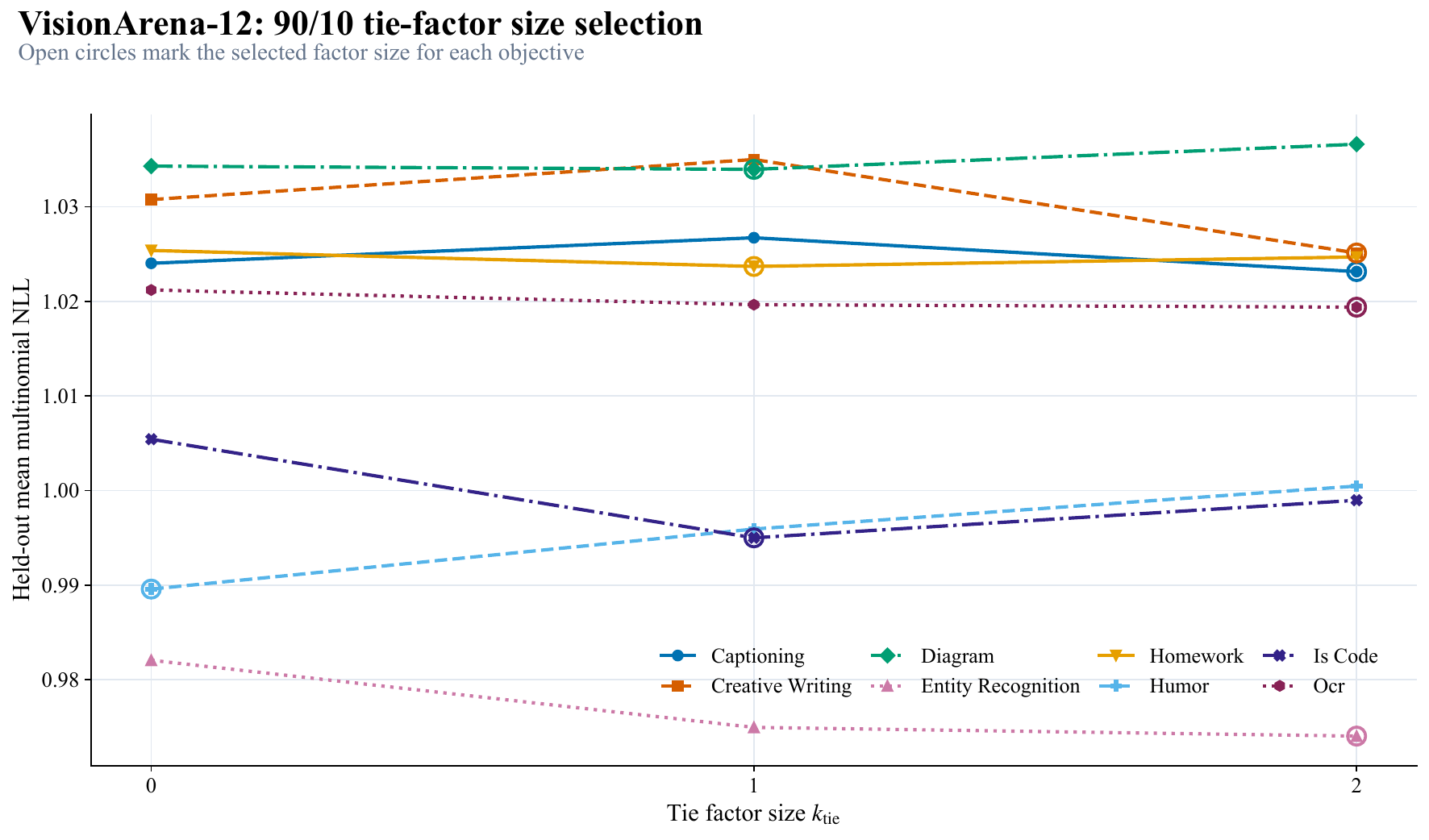}
\caption{VisionArena-12: held-out NLL for each candidate tie-factor size \(k_{\rm tie}\); open circles mark the selected size for each objective.}
\label{fig:realistic-davidson-vision12-model-selection}
\end{figure}

The held-out losses for all candidate tie-factor sizes are shown in \Cref{fig:realistic-davidson-arena35-model-selection,fig:realistic-davidson-search8-model-selection,fig:realistic-davidson-vision12-model-selection}.

\paragraph{Binary interface and fitted target.}
After refitting the selected model to all observations, we convert its fitted outcome probabilities to half-credit win probabilities
\begin{equation}
    \widehat p^{F}_{ij,r}
    =
    \pi^{W}_{ij,r}
    +
    \frac12\pi^{U}_{ij,r},
    \qquad
    \widehat p^{F}_{ji,r}
    =
    \pi^{L}_{ij,r}
    +
    \frac12\pi^{U}_{ij,r}
    =
    1-\widehat p^{F}_{ij,r}.
    \label{eq:davidson-half-credit-interface}
\end{equation}
As in Appendix~\ref{sec:arena-replay}, each oracle query first draws an opponent uniformly from the other retained models. Conditional on the selected pair \((i,j)\) and objective \(r\), the oracle draws an outcome from the fitted three-category distribution and resolves a non-decisive outcome with an independent fair coin, which is equivalent to a single Bernoulli draw with mean \(\widehat p^{F}_{ij,r}\). We then compute
\[
    b^{F}_{i,r}
    =
    \frac{1}{K-1}\sum_{j\ne i}\widehat p^{F}_{ij,r}
\]
and obtain the Pareto set \(\mathcal P_F\) of the fitted Borda scores.

For each model--objective pair, the marginal observation law is \(\operatorname{Bernoulli}(b^F_{i,r})\), so the implementation draws batch win counts from the corresponding binomial distribution, as under \(D\).

\FloatBarrier

\subsubsection{Target and stopping-time comparisons}
\label{app:real-data-davidson-fit}

\begin{figure}[H]
\centering
\includegraphics[width=0.98\textwidth,keepaspectratio]
{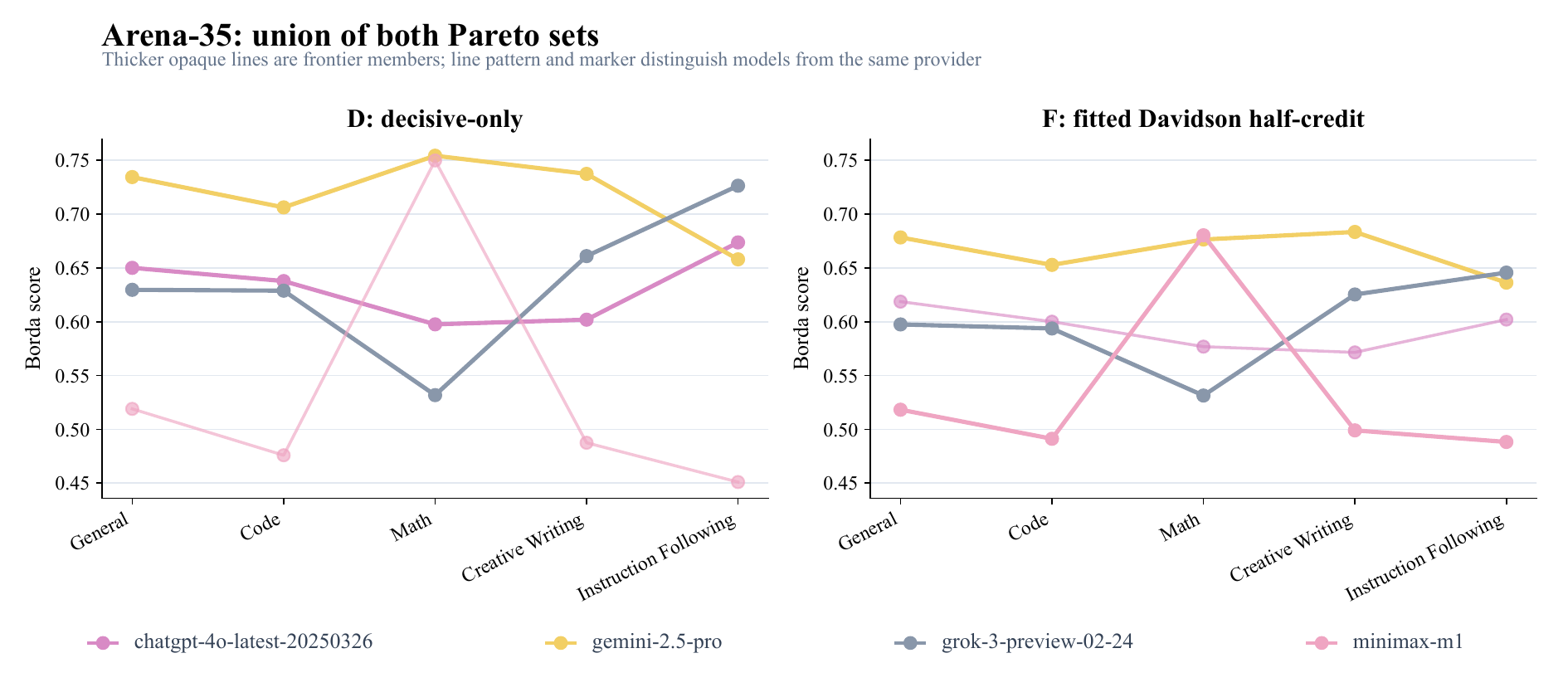}
\caption{Arena-35: Borda profiles of the models in \(\mathcal P_D\cup\mathcal P_F\) under each target; \texttt{minimax-m1} enters and \texttt{chatgpt-4o-\allowbreak latest-\allowbreak 20250326} leaves the strict Pareto frontier under \(F\). Left: decisive-only target \(D\); right: fitted Davidson target \(F\).}
\label{fig:realistic-davidson-arena35-profiles}
\end{figure}

\paragraph{Target sensitivity.}
\Cref{tab:realistic-tie-sensitivity} compares the two alternatives with \(D\). On Arena-35, \(H\) adds \texttt{minimax-m1} to the frontier, increasing its size from three to four. Under \(F\), \texttt{minimax-m1} enters and \texttt{chatgpt-4o-\allowbreak latest-\allowbreak 20250326} leaves, keeping the frontier size at three. The respective Jaccard overlaps with \(D\) are \(0.75\) and \(0.5\). SearchArena-8 and VisionArena-12 retain their decisive-only frontiers under both constructions. \Cref{fig:realistic-direct-arena35-sensitivity,fig:realistic-davidson-arena35-profiles} show the Arena-35 profiles where membership changes.

\begin{figure}[H]
\centering
\includegraphics[width=0.98\textwidth,keepaspectratio]
{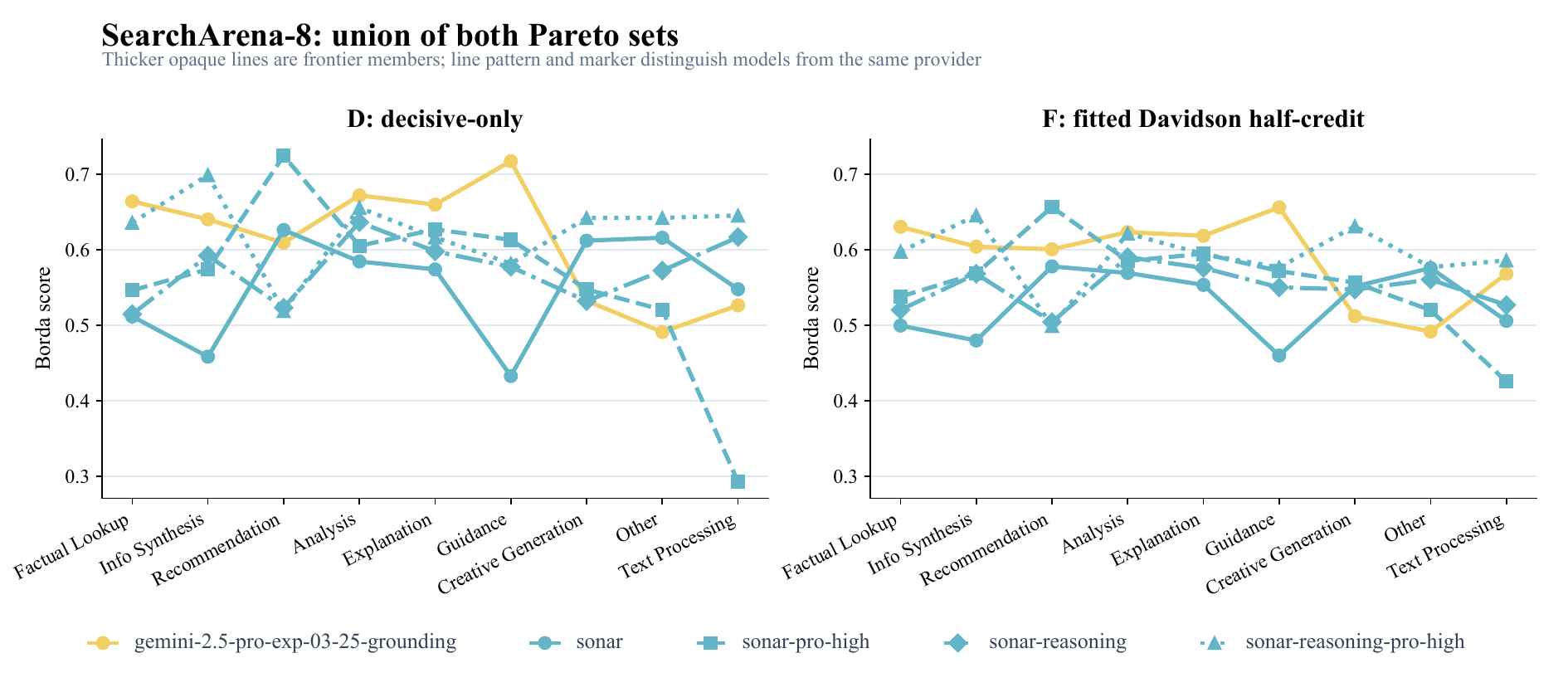}
\caption{SearchArena-8: Borda profiles of the models in \(\mathcal P_D\cup\mathcal P_F\) under each target; frontier membership is unchanged. Left: decisive-only target \(D\); right: fitted Davidson target \(F\).}
\label{fig:realistic-davidson-search8-profiles}
\end{figure}

\begin{figure}[H]
\centering
\includegraphics[width=0.98\textwidth,keepaspectratio]
{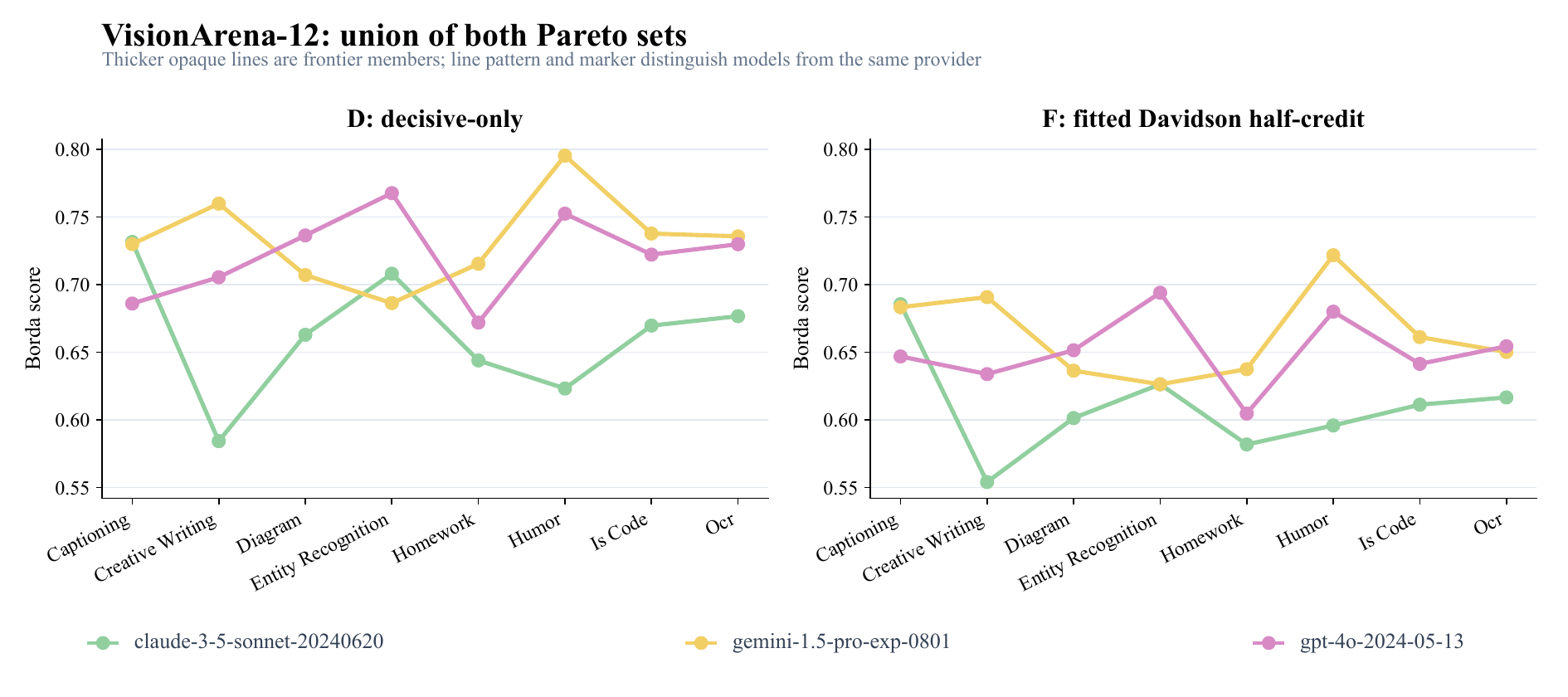}
\caption{VisionArena-12: Borda profiles of the models in \(\mathcal P_D\cup\mathcal P_F\) under each target; frontier membership is unchanged. Left: decisive-only target \(D\); right: fitted Davidson target \(F\).}
\label{fig:realistic-davidson-vision12-profiles}
\end{figure}

\begin{table}[H]
\centering
\caption{
    Target sensitivity on the fixed sets of retained models used in the main experiments. \(D/H/F\) gives the strict-frontier sizes under the decisive-only, direct half-credit, and fitted Davidson targets. \(\operatorname{Jac}(D,H)\) and \(\operatorname{Jac}(D,F)\) are Pareto-set Jaccard overlaps. Mean coordinate shifts are averaged over all model--objective coordinates, and \(\rho_{\min}\) is the minimum objective-wise Spearman correlation with the decisive-only target.
}
\label{tab:realistic-tie-sensitivity}
\footnotesize
\resizebox{\textwidth}{!}{%
\begin{tabular}{@{}lrrrrrrr@{}}
\toprule
Dataset& \(D/H/F\)& \(\operatorname{Jac}(D,H)\)& \(\operatorname{Jac}(D,F)\)& Mean \(\lvert b^H-b^D\rvert\)& Mean \(\lvert b^F-b^D\rvert\)& \(\rho_{\min}(D,H)\)& \(\rho_{\min}(D,F)\) \\
\midrule
Arena-35& \(3/4/3\)& \(0.750\)& \(0.500\)& \(0.0244\)& \(0.0267\)& \(0.993\)& \(0.973\) \\
SearchArena-8& \(5/5/5\)& \(1.000\)& \(1.000\)& \(0.0333\)& \(0.0352\)& \(0.929\)& \(0.810\) \\
VisionArena-12& \(3/3/3\)& \(1.000\)& \(1.000\)& \(0.0529\)& \(0.0524\)& \(0.958\)& \(0.951\) \\
\bottomrule
\end{tabular}%
}
\end{table}

\paragraph{Borda profiles under the fitted Davidson target.}
\Cref{fig:realistic-davidson-arena35-profiles,fig:realistic-davidson-search8-profiles,fig:realistic-davidson-vision12-profiles} compare \(D\) with \(F\) on Arena-35, SearchArena-8, and VisionArena-12, respectively.

\paragraph{Stopping times and gap summaries.}
Under \(D\) and \(F\), focal observations are Bernoulli, so their laws are determined by the corresponding Borda means. This comparison isolates the change in the mean array within the same observation family. The comparison of \(D\) with \(H\) also changes the observation family to \(\{0,1/2,1\}\). If \(u^H_{i,r}\) is the uniform-opponent probability of a non-decisive outcome, its variance is \(b^H_{i,r}(1-b^H_{i,r})-u^H_{i,r}/4\), smaller than the Bernoulli variance at the same mean; its mean also differs from that under \(D\).

\Cref{tab:realistic-davidson-algorithm-sensitivity} reports the minimum gaps, inverse-squared-gap sums, and stopping times. The gap summaries do not track stopping times closely: from \(D\) to \(F\), \(\mathcal H_X\) falls by \(16\%\) on SearchArena-8 while the mean stopping time rises by \(2\%\), and it rises by \(32\%\) on Arena-35 while the mean stopping time more than doubles.

\begin{table}[H]
\centering
\caption{Gap summaries and stopping times under \(D\), \(H\), and \(F\). Panel (a) gives the minimum gaps and \(\mathcal H_X=\sum_i\Delta_i(b^X)^{-2}\), with the latter in thousands. Panel (b) gives mean stopping times in millions of comparisons, with one standard error in parentheses, and ratios of means. In panel (c), each entry gives a final phase followed by its trial count in parentheses. Every trial recovers its construction's strict frontier.}
\label{tab:realistic-davidson-algorithm-sensitivity}
\footnotesize
\setlength{\tabcolsep}{4pt}
\textbf{(a) Gap summaries}\par\smallskip
\begin{tabular*}{\textwidth}{@{\extracolsep{\fill}}lrrrrrr@{}}
\toprule
Dataset & \(\Delta_{\min}^{D}\) & \(\Delta_{\min}^{H}\) & \(\Delta_{\min}^{F}\) & \(\mathcal H_D/10^3\) & \(\mathcal H_H/10^3\) & \(\mathcal H_F/10^3\) \\
\midrule
Arena-35 & 0.00441 & 0.00520 & 0.00391 & 111.72 & 101.03 & 147.38 \\
SearchArena-8 & 0.00460 & 0.00539 & 0.00506 & 94.97 & 69.93 & 79.71 \\
VisionArena-12 & 0.02171 & 0.00649 & 0.00225 & 5.38 & 49.25 & 397.08 \\
\bottomrule
\end{tabular*}
\par\medskip
\textbf{(b) Stopping times}\par\smallskip
\begin{tabular*}{\textwidth}{@{\extracolsep{\fill}}lrrrrr@{}}
\toprule
Dataset & \(\bar\tau_D\) & \(\bar\tau_H\) & \(\bar\tau_F\) & \(\bar\tau_H/\bar\tau_D\) & \(\bar\tau_F/\bar\tau_D\) \\
\midrule
Arena-35 & \(86.82\;(2.29)\) & \(116.01\;(0.39)\) & \(196.97\;(11.25)\) & \(1.34\times\) & \(2.27\times\) \\
SearchArena-8 & \(124.34\;(0.03)\) & \(124.79\;(0.03)\) & \(126.50\;(0.02)\) & \(1.00\times\) & \(1.02\times\) \\
VisionArena-12 & \(8.80\;(0.01)\) & \(114.52\;(0.04)\) & \(471.55\;(13.90)\) & \(13.01\times\) & \(53.59\times\) \\
\bottomrule
\end{tabular*}
\par\medskip
\textbf{(c) Final phases (numbers of trials)}\par\smallskip
\begin{tabular*}{\textwidth}{@{\extracolsep{\fill}}lccc@{}}
\toprule
Dataset & \(D\) & \(H\) & \(F\) \\
\midrule
Arena-35 & 9 (99), 10 (1) & 9 (100) & 9 (51), 10 (49) \\
SearchArena-8 & 9 (100) & 9 (100) & 9 (100) \\
VisionArena-12 & 7 (100) & 9 (100) & 10 (99), 11 (1) \\
\bottomrule
\end{tabular*}
\end{table}

In phase \(m\), \VBEGE\ tops up every active model--objective pair to \(n_m\) samples, and \(n_m\) grows roughly fourfold between phases. The extra phases on VisionArena-12 and under \(F\) on Arena-35 therefore help explain their large stopping-time increases. SearchArena-8 stops after nine phases under all three constructions.

Each phase's cost also depends on the number of active models. On Arena-35, all 100 trials under \(D\) enter phase nine with two active models. Under \(H\), 98 trials enter that phase with three active models and the remaining two with two. The mean comparison count in phase nine is therefore \(55.13\)M under \(D\) and \(82.14\)M under \(H\). This increase of \(27.01\)M accounts for approximately \(93\%\) of the overall increase of \(29.19\)M in mean stopping time.

\FloatBarrier

\section{Additional comparison with a Track-and-Stop extension}
\label{app:pareto-btglr-comparison}

The BT-MLE baseline in the main experiments samples pair--objective cells uniformly, so its gap to \VBEGE\ leaves open whether an adaptive likelihood-based method would perform better. 
This section empirically examines that question using a Pareto extension of the scalar top-\(k\) Track-and-Stop method of \citet{goldberger2026optimal}. 
Appendix~\ref{app:pareto-btglr-impl} describes the extension, and Appendix~\ref{app:pareto-btglr-results} compares it with \VBEGE\ on the synthetic tasks of Appendix~\ref{app:synthetic-experiments}. 

\subsection{Implementation used in the reported experiments}
\label{app:pareto-btglr-impl}

The scalar top-\(k\) method of \citet{goldberger2026optimal} combines Bradley--Terry MLE, adaptive allocation, and a generalized likelihood-ratio (GLR) stopping rule. We retain this template and replace its scalar alternatives with Pareto alternatives, calling the resulting procedure \emph{Pareto BT-GLR Track-and-Stop} (Pareto BT-GLR T\&S). Here, an \emph{alternative} is a set of score arrays defined by linear inequalities describing a change in the current Pareto frontier, up to boundary points. The implementation ranks these alternatives by approximate Fisher-information scores and refits the likelihood only under the selected alternative. We call this approximation \emph{screening}.

Let \(N_{\mathrm{cell}}=d\binom K2\) be the number of unordered pair--objective cells. The method takes \(n_0\) initial samples per cell and initializes sampling weights \(w_{ij,r}=1/N_{\mathrm{cell}}\), which are carried across checks. At check \(m\), let \(t\) be the total number of comparisons and \(N_{ij,r}(t)\) the count in cell \((i,j,r)\). Write \(\widehat\theta_{\cdot,r,t}\in\mathbb R^K\) for the fitted score vector on objective \(r\) and \(\widehat\theta_t=(\widehat\theta_{i,t})_{i=1}^K\), with \(\widehat\theta_{i,t}=(\widehat\theta_{i,r,t})_{r=1}^d\), for the array of model score vectors. Constants are collected in \Cref{tab:pareto-btglr-constants}.

\begin{table}[htbp]
\centering
\caption{Constants used by Pareto BT-GLR T\&S.}
\label{tab:pareto-btglr-constants}
\footnotesize
\begin{tabular*}{\textwidth}{@{\extracolsep{\fill}}llr@{}}
\toprule
Quantity & Notation & Value \\
\midrule
Threshold regularization & \(\lambda_{\mathrm{reg}}\) & \(0.1\) \\
Bernoulli variance bound & \(\bar\sigma^2\) & \(1/4\) \\
Score box bound & \(S_{\max}\) & \(2\) \\
Fitting tolerance; maximum iterations & & \(10^{-7};\ 1{,}000\) \\
Initial samples per cell & \(n_0\) & \(2\) \\
Check-growth factor & \(a_{\mathrm{batch}}\) & \(1.8\) \\
Comparison cap & \(T_{\max}\) & \(10^{18}\) \\
Multiplicative-weights coefficient & \(\alpha_{\mathrm{MW}}\) & \(0.8\) \\
Gain normalization floor & \(\varepsilon_\kappa\) & \(10^{-15}\) \\
Uniform-mixing exponent & \(\gamma_{\mathrm{mix}}\) & \(1/3\) \\
\bottomrule
\end{tabular*}
\end{table}

\begin{enumerate}
\item \textbf{Fit.} Separately for each objective \(r\), fit the Bradley--Terry model to all comparisons collected so far:
\[
    \widehat\theta_{\cdot,r,t}\in\arg\!\max_{\vartheta\in\Theta_{\mathrm{box}}}\ell_{r,t}(\vartheta),
    \qquad
    \Theta_{\mathrm{box}}:=\{\vartheta\in\mathbb R^K:\mathbf1^\top\vartheta=0,\ \|\vartheta\|_\infty\leq S_{\max}\},
\]
where \(\ell_{r,t}\) is the log-likelihood for objective \(r\). The numerical fits use the tolerance and iteration limit in \Cref{tab:pareto-btglr-constants}.

\item \textbf{Construct alternatives.} Let \(\widehat{\mathcal P}_t=\mathcal P(\widehat\theta_t)\), which is also the Pareto frontier of the induced Borda scores by \Cref{lem:borda-order-preservation}. There are two kinds of alternatives. A \emph{drop} alternative \((p,p')\), for distinct \(p,p'\in\widehat{\mathcal P}_t\), imposes \(\theta_{p',r}\geq\theta_{p,r}\) for every \(r\). These inequalities describe the closure of the strict-dominance region; boundary points need not remove \(p\) from the strict Pareto frontier. An \emph{add} alternative for \(c\notin\widehat{\mathcal P}_t\) assigns an objective \(g(p)\in[d]\) to each \(p\in\widehat{\mathcal P}_t\) and imposes \(\theta_{c,g(p)}\geq\theta_{p,g(p)}\). Thus no current Pareto model strictly dominates \(c\). Each add candidate has \(d^{|\widehat{\mathcal P}_t|}\) possible assignments.

\item \textbf{Screen alternatives and refit.} Set \(\widehat p_{ij,r,t}=\sigma(\widehat\theta_{i,r,t}-\widehat\theta_{j,r,t})\). The observed Fisher information is
\[
    \mathcal I_{r,t}=\sum_{i<j}N_{ij,r}(t)\widehat p_{ij,r,t}(1-\widehat p_{ij,r,t})(e_i-e_j)(e_i-e_j)^\top,
\]
where \(e_i\) is the \(i\)th standard basis vector. The local quadratic approximation to the likelihood loss from imposing \(\theta_{a,r}\geq\theta_{b,r}\) is
\[
    h_{ab,r,t}
    =\frac{\bigl((\widehat\theta_{b,r,t}-\widehat\theta_{a,r,t})^+\bigr)^2}
    {2(e_a-e_b)^\top\mathcal I_{r,t}^{\dagger}(e_a-e_b)},
\]
where \(\mathcal I_{r,t}^{\dagger}\) is the Moore--Penrose pseudoinverse. We score a drop alternative by \(\sum_r h_{p'p,r,t}\) and an add candidate \(c\) by \(\max_{p\in\widehat{\mathcal P}_t}\min_r h_{cp,r,t}\). The drop score sums losses on separate objectives. The add score keeps only the most expensive of the per-model cheapest constraints and can understate the joint loss, potentially favoring add candidates. These scores serve as a computational heuristic for selecting an alternative.

For an add candidate \(c\), we use the assignment \(g_c(p)\in\arg\min_r h_{cp,r,t}\) for every \(p\in\widehat{\mathcal P}_t\). This assignment minimizes \(\max_p h_{cp,g(p),t}\) over all assignments \(g\), and the minimum is the candidate's score above.

Let \(\widehat\xi_t\) be the drop or add alternative with the smallest screening score, and let \(h_t^{\mathrm{scr}}\) denote this score. 

For each objective \(r\), we refit the Bradley--Terry model subject to the ordering constraints specified by \(\widehat\xi_t\).
For a drop alternative \((p,p')\), the constraint is \(\vartheta_{p'}\geq\vartheta_p\).
For an add alternative \((c,g)\), the constraints are \(\vartheta_c\geq\vartheta_p\) for all \(p\in\widehat{\mathcal P}_t\) with \(g(p)=r\).
Let \(\Theta_{\xi,r}\) denote the subset of
\(\Theta_{\mathrm{box}}\) satisfying these constraints for alternative \(\xi\).
We numerically solve
\[
    \max_{\vartheta\in\Theta_{\widehat\xi_t,r}}
    \ell_{r,t}(\vartheta)
\]
and denote the resulting estimate by \(\widetilde\theta_{\cdot,r,t}\).
The optimization separates across objectives because the total log-likelihood is a sum over objectives and each constraint involves only one objective.

These fits separate across objectives. To account for numerical optimization error, write \(f_{r,t}=-\ell_{r,t}\) and compute
\[
    g_{r,t}=\left[\nabla f_{r,t}(\widetilde\theta_{\cdot,r,t})^\top\widetilde\theta_{\cdot,r,t}
    -\min_{\vartheta\in\Theta_{\widehat\xi_t,r}}
    \nabla f_{r,t}(\widetilde\theta_{\cdot,r,t})^\top\vartheta\right]^+.
\]
Convexity gives a lower bound \(f_{r,t}(\widetilde\theta_{\cdot,r,t})-g_{r,t}\) on the optimum over \(\Theta_{\widehat\xi_t,r}\). Since \(f_{r,t}(\widehat\theta_{\cdot,r,t})\) is an upper bound on the optimum over \(\Theta_{\mathrm{box}}\) without the alternative's constraints, a lower bound on the selected alternative's likelihood loss is
\[
    G_t^{\mathrm{alt}}=\sum_r\left[
    f_{r,t}(\widetilde\theta_{\cdot,r,t})-g_{r,t}
    -f_{r,t}(\widehat\theta_{\cdot,r,t})\right]^+.
\]
The implemented statistic is \(\Psi_t=\min\{h_t^{\mathrm{scr}},G_t^{\mathrm{alt}}\}\), so stopping requires both the screening score and this numerical lower bound to exceed the threshold.

\item \textbf{Stop.} Define the comparison-count matrix
\[
    \Omega_{r,t}=\sum_{i<j}N_{ij,r}(t)(e_i-e_j)(e_i-e_j)^\top
\]
and threshold
\[
    \beta_t(\delta)=\log\frac1\delta
    +\frac{\lambda_{\mathrm{reg}}}{2}\sum_r\|\widehat\theta_{\cdot,r,t}\|_2^2
    +\frac12\sum_r\log\det\!\left(I_K+\frac{\bar\sigma^2}{\lambda_{\mathrm{reg}}}\Omega_{r,t}\right).
\]
The method returns \(\widehat{\mathcal P}_t\) when the fits \(\widehat\theta_{\cdot,r,t}\) and \(\widetilde\theta_{\cdot,r,t}\) have converged for every \(r\) and \(\Psi_t\geq\beta_t(\delta)\). If it has not stopped at \(t=T_{\max}\), the trial is \emph{capped} and terminates. Otherwise, it proceeds to the next batch.

\item \textbf{Allocate the next batch.} For each pair--objective cell, compute
\[
    \kappa_{ij,r,t}=\kl\!\left(
    \sigma(\widehat\theta_{i,r,t}-\widehat\theta_{j,r,t}),
    \sigma(\widetilde\theta_{i,r,t}-\widetilde\theta_{j,r,t})\right),
    \qquad
    \kappa_{\max,t}:=\max\{\varepsilon_\kappa,\max_{a<b,s}\kappa_{ab,s,t}\}.
\]
Update the current weights by
\[
    w_{ij,r}^{\mathrm{new}}
    =\frac{w_{ij,r}^{\mathrm{old}}\exp\!\left(\alpha_{\mathrm{MW}}\kappa_{ij,r,t}/(\sqrt m\,\kappa_{\max,t})\right)}
    {\sum_{a<b}\sum_{s=1}^d w_{ab,s}^{\mathrm{old}}\exp\!\left(\alpha_{\mathrm{MW}}\kappa_{ab,s,t}/(\sqrt m\,\kappa_{\max,t})\right)}.
\]
This favors comparisons that distinguish the current fit from the selected alternative. Schedule the next check at
\[
    t_{\mathrm{next}}:=\min\{T_{\max},\max(t+N_{\mathrm{cell}},\lceil a_{\mathrm{batch}}t\rceil)\}.
\]
Form the mixed weights \(\widetilde w_{ij,r}=(1-\rho_t^{\mathrm{mix}})w_{ij,r}+\rho_t^{\mathrm{mix}}/N_{\mathrm{cell}}\) with \(\rho_t^{\mathrm{mix}}=t_{\mathrm{next}}^{-\gamma_{\mathrm{mix}}}\); the unmixed weights \(w_{ij,r}\) are the ones carried to the next check. The \(t_{\mathrm{next}}-t\) queries are allocated by a batched variant of C-tracking \citep{garivier2016optimal}: update the cumulative target counts, initialized at \(P_{ij,r}=n_0\), by \(P_{ij,r}\leftarrow P_{ij,r}+(t_{\mathrm{next}}-t)\,\widetilde w_{ij,r}\), then split the batch across cells in proportion to the positive deficits \((P_{ij,r}-N_{ij,r}(t))^+\), rounding to integers so that the allocations sum to \(t_{\mathrm{next}}-t\). Sample the batch and proceed to the next check.
\end{enumerate}


\paragraph{Relation to the scalar implementation.}
The threshold follows the form in \citet[Equation~(4)]{goldberger2026optimal}. Stacking the objective-specific score vectors gives a block-diagonal comparison-count matrix with blocks \(\Omega_{r,t}\), which yields the sums above. We use the mixed and cumulative targets of their C-tracking construction (Section~5.4) and the mixing exponent of their Section~7; unlike C-tracking, which assigns one query at a time to the cell with the largest deficit, we allocate each batch in proportion to the positive deficits. Our allocation uses one multiplicative-weights update per check against the selected alternative, with gains normalized to \([0,1]\); this provides a simple allocation rule compatible with the screened fits. It replaces the scalar method's entropic follow-the-regularized-leader updates, whose experiments use learning rate \(t^{-0.2}\). The screening and allocation choices are implementation choices for this exploratory extension.

\subsection{Comparison with VB-EGE}
\label{app:pareto-btglr-results}

We run this exploratory comparison at \(\delta=0.05\), using the instances and replication scheme of Appendix~\ref{app:synthetic-experiments}. This gives \(12{,}400\) trials per method over 28 settings: 12 scaling settings, 8 benchmarks, and 8 Pareto-set size settings. The \emph{capped mean stopping time} is the mean comparison count until stopping or capping, with capped trials recorded at \(T_{\max}\); standard errors use the same counts. For each setting, let \(R\) be the ratio of the capped mean stopping time of Pareto BT-GLR T\&S to that of \VBEGE. For a group of \(n\) settings, we report the geometric mean \(\exp(n^{-1}\sum_{k=1}^n\log R_k)\).

\begin{figure}[H]
\centering
\includegraphics[width=0.98\textwidth,keepaspectratio]
{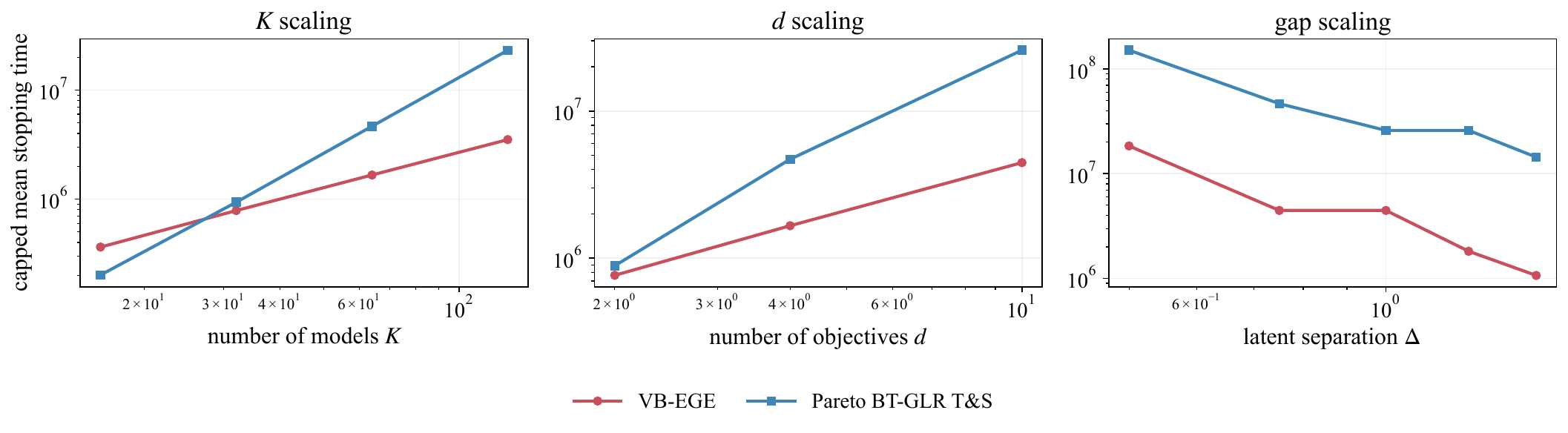}
\caption{Scaling comparison at \(\delta=0.05\). Pareto BT-GLR T\&S has a smaller capped mean stopping time than VB-EGE only at \(K=16\) and a larger one in the other 11 settings. Curves show capped mean stopping times \(\pm\) one standard error.}
\label{fig:pareto-btglr-scaling}
\end{figure}

\begin{figure}[H]
\centering
\includegraphics[width=0.98\textwidth,keepaspectratio]
{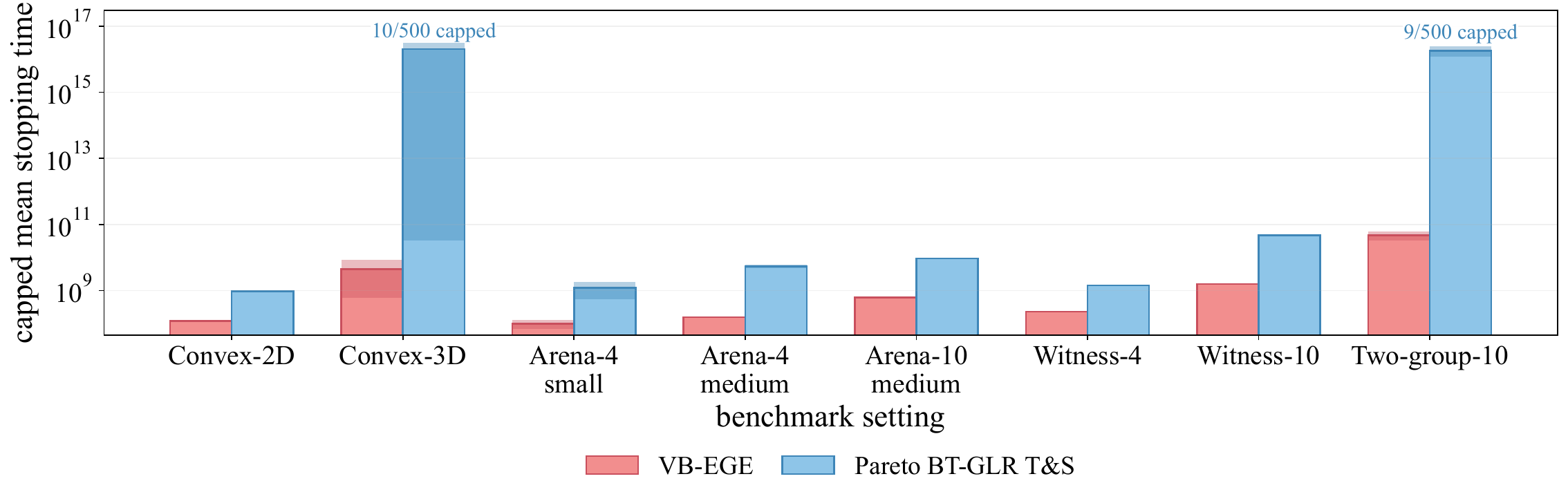}
\caption{Benchmark comparison at \(\delta=0.05\). Bars show capped mean stopping times \(\pm\) one standard error. Labels above the bars give the numbers of capped Pareto BT-GLR T\&S trials (10 of 500 on Convex-3D and 9 of 500 on Two-group-10). The two capped means are dominated by these trials under the capping convention above.}
\label{fig:pareto-btglr-benchmarks}
\end{figure}

\begin{figure}[H]
\centering
\includegraphics[width=0.98\textwidth,keepaspectratio]
{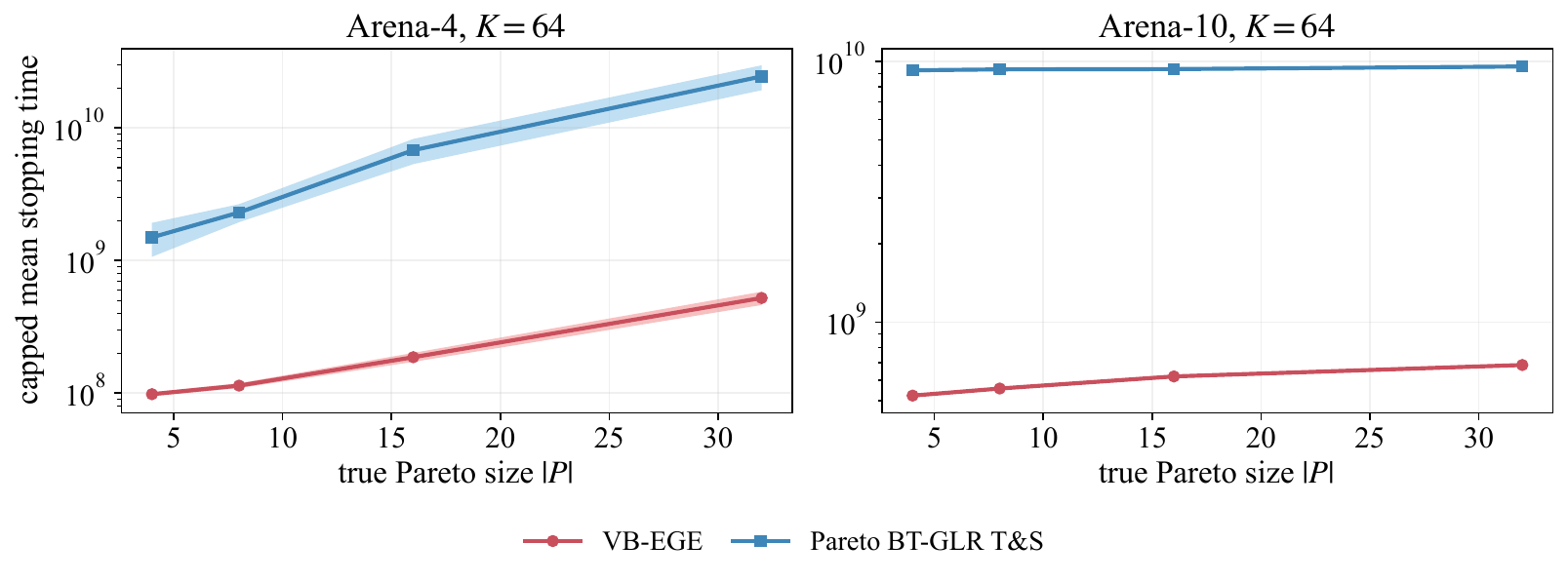}
\caption{Pareto-set size sensitivity at \(\delta=0.05\). Pareto BT-GLR T\&S has a larger capped mean stopping time than VB-EGE in all eight settings, with a geometric-mean ratio of \(20.6\). Curves show capped mean stopping times \(\pm\) one standard error.}
\label{fig:pareto-btglr-pareto-size}
\end{figure}

Pareto BT-GLR T\&S has the smaller capped mean stopping time in 1 of the 28 settings: 1 of the 12 scaling settings, and none of the 8 benchmarks or the 8 Pareto-set size settings (\Cref{fig:pareto-btglr-scaling,fig:pareto-btglr-benchmarks,fig:pareto-btglr-pareto-size}). The geometric-mean ratios are \(4.10\) over the scaling settings and \(20.6\) over the Pareto-set size settings.

On Convex-3D and Two-group-10, respectively, 10 and 9 of the 500 Pareto BT-GLR T\&S trials are capped (\(2.0\%\) and \(1.8\%\)). These capped trials contribute \(2.0\times10^{16}\) and \(1.8\times10^{16}\) to the corresponding means, accounting for more than \(99.8\%\) of each mean. Consequently, the two bars primarily reflect the chosen cap, and an aggregate benchmark ratio would be sensitive to that choice. We report a geometric-mean ratio of \(14.5\) over the six benchmarks with no capped trials; across all 26 settings with no capped trials, it is \(9.01\). Every \VBEGE\ trial stops. All \(12{,}381\) stopped Pareto BT-GLR T\&S trials and all \(12{,}400\) \VBEGE\ trials exactly recover the true Pareto frontier.

\end{document}